\documentclass[11pt, a4paper, copyright, gdm]{google}

\usepackage[authoryear, sort&compress, round]{natbib}
\usepackage{mathtools}
\usepackage{multirow}
\usepackage{longtable}
\usepackage{array}
\usepackage[above,below]{placeins}
\usepackage{algorithm}
\usepackage{algpseudocode}
\usepackage{cleveref}

\graphicspath{{./figures/}}
\theoremstyle{plain}

\theoremstyle{definition}

\theoremstyle{remark}

\Crefname{theorem}{Theorem}{Theorems}
\Crefname{proposition}{Proposition}{Propositions}
\Crefname{lemma}{Lemma}{Lemmas}
\Crefname{corollary}{Corollary}{Corollaries}

\newcommand{\R}{\mathbb{R}}

\newcommand{\sigmoid}{\sigma}

\newcommand{\games}{\mathcal{G}}
\newcommand{\modelset}{\mathcal{M}}

\newcommand{\bt}{Bradley--Terry}

\newcolumntype{Y}{>{\raggedright\arraybackslash}X}
\newcolumntype{C}{>{\centering\arraybackslash}p{0.09\textwidth}}


\uselogo{}
\title{Schedule, Rank, Profile: Continuous Evaluation of LLM Agents in Games}
\correspondingauthor{}
\reportnumber{}
\renewcommand{\today}{} 

\author[*,1,2]{Martyna Plomecka}
\author[*,1,2]{Yao Yan}
\author[1,2]{DJ Sterling}
\author[1,2]{Jon Lipovetz}
\author[1]{John Schultz}
\author[1,2]{Michael Aaron}
\author[1,2]{Chiamaka Chukwuka}
\author[1,2]{Will Cukierski}
\author[1,2]{Christopher D'Mello}
\author[1,2]{Bovard Doerschuk-Tiberi}
\author[1,2]{Robert Fraser}
\author[1,2]{Ian Gemp}
\author[1,2]{Yuting Han}
\author[1, 2]{Addison Howard}
\author[1]{Oran Kelly}
\author[1]{Siqi Liu}
\author[1,2]{Jeff Moser}
\author[1, 2]{Meghan O'Connell}
\author[1,2]{Jun Peng}
\author[1]{Laurel Prince}
\author[1,2]{Meg Risdal}
\author[1,2]{Jonathan Ward}
\author[1,2]{Domino Weir}
\author[1]{Marc Lanctot}

\affil[*]{Equal contribution}
\affil[1]{Google DeepMind}
\affil[2]{Kaggle}

\begin{abstract}
Games are a robust test for large language model agents: they require
sequential decisions, adversarial or cooperative interaction, state tracking,
reasoning under uncertainty, and long-horizon planning. However, a one-time game
benchmark can go stale as models, prompts, environment implementations,
and evaluation policies change, and refreshing it is costly: full round-robin
evaluation scales poorly in the number of model pairs, and the resulting
outcomes live on incompatible per-game scales. We present a framework for
continuous evaluation of LLM agents in game settings, organized around three
coupled components. First, an adaptive scheduler that maps pairwise \bt{} uncertainty to
reduced per-pair match targets, concentrating budget on comparisons that remain
unresolved. Second, a Unified Leaderboard (ULB) that reduces heterogeneous outcomes to pairwise
evidence and fits a game-balanced \bt{} model by minorization--maximization with
bootstrap intervals, so that high-throughput games cannot dominate the aggregate
ranking. Third, a game skill-profile methodology that scores each
game along diverse skill dimensions, making the capability coverage of the suite explicit and turning game selection into a design decision rather than an artifact of availability. In this paper, we specify the research design and outcome reduction of the current implementation and evaluate the framework on a ten-game suite spanning perfect-information board games,
imperfect-information games, and social/language settings, reporting LLM capability profiles, evaluation-cost savings across model-refresh cycles, and an empirical Pareto frontier relating model
performance to inference cost. As deployed, the scheduler saved 61.7\% of
episodes; it counted each episode in the two team games as four records,
overstating their savings, and the other eight games save 61.9\%.
\end{abstract}

\begin{document}
\maketitle

\section{Introduction}
\label{sec:intro}
Static benchmarks can lose diagnostic value as models improve: a fixed dataset
can saturate, become exposed through training data, or stop discriminating
between systems as capabilities change
\citep{kiela2021dynabench,white2025livebench}. Games offer a dynamic evaluation
opportunity. An agent must choose actions over time,
respond to opponents or partners, obey environment rules, manage partial
information, and recover from earlier mistakes. These demands exercise agentic
behavior rather than isolated task completion.

Game Arena (\url{https://www.kaggle.com/game-arena}) provides this substrate
\citep{risdal2025gamearena}. In this paper, we show how to operate it as a
long-standing scientific benchmark. Running games continuously raises operational
and statistical questions that a platform description can leave open: which
games to include, how to combine outcomes from games with incomparable reward
scales, when to schedule additional matches, how to absorb model refreshes, and
how stable the resulting leaderboard is over time. 

\subsubsection*{Contributions}

To address these questions, we make the following contributions:

\begin{enumerate}[leftmargin=*]
  \item We propose a continuous evaluation framework for LLM agents in
  competitive and cooperative game environments, built around three design
  requirements: repeatability, comparability, and refreshability
  (\Cref{sec:framework}).
  \item We formalize an adaptive scheduling rule: a Newton--Raphson
  Bradley--Terry fit yields pairwise confidence estimates, and a fixed cubic
  target-reduction rule maps those confidences to per-pair match targets
  (\Cref{sec:scheduler}).
  \item We define the Unified Leaderboard, which aggregates model performance
  across games via a game-balanced minorization--maximization Bradley--Terry
  model with bootstrap confidence intervals (\Cref{sec:ulb}).
  \item We introduce a game skill-profile methodology and use it to construct a
  ten-game suite covering six construct dimensions: search and planning,
  tactical and constraint reasoning, spatial and topological reasoning,
  uncertainty and belief reasoning, social and mental-state reasoning, and
  language and pragmatics (\Cref{sec:skills}).
\end{enumerate}

These contributions build on four related lines of work: pairwise LLM
evaluation, game-based agent benchmarks, persistent player-rating systems, and
adaptive experimental design. Chatbot Arena ranks models from crowdsourced
pairwise preferences using Bradley--Terry methods
\citep{chiang2024chatbot}. Subsequent analyses incorporate response-style
covariates \citep{arena2024style} and report leaderboards for
category-labeled subsets of battles \citep{arena2024categories}. Our Unified
Leaderboard instead balances evidence across structured game environments,
while retaining separate per-game results.

Game-based LLM evaluation frameworks include GameBench \citep{gamebench},
TextArena \citep{textarena}, GAMEBoT \citep{lin2025gamebot}, and LLMsPark
\citep{chen2025llmspark}; OpenSpiel provides a general framework for research
in games \citep{LanctotEtAl2019OpenSpiel}. This work is complementary: it
focuses on the scheduling, cross-game aggregation, and capability-coverage
requirements that arise when a game suite is operated and refreshed
continuously.

Persistent ratings in human-play platforms provide a further precedent.
Lichess uses Glicko-2 \citep{glickman2000glicko2,lichessRating}, Board Game
Arena uses a modified Elo system updated after each game
\citep{elo1978rating,bgaRating}, and Yucata maintains a separate TrueSkill
rating for each game \citep{herbrich2006trueskill,yucataRating}. These systems (along with Whole-History Rating \citep{coulom2008whr}) model time-varying human skill across self-selected arrivals; our setting periodically refits a batch Bradley--Terry model to deliberately scheduled agent evaluations, since pinned model checkpoints have fixed weights (a harness change enters as a new submission; \Cref{app:rq2-continuous-refresh}), pairings are actively selected from $C=H^{-1}$ under mirrored seeds, and outcomes pool across ten games.

Finally, recent work formalizes adaptive scheduling as \emph{active
evaluation} \citep{Lanctot26ActiveEval}, comparing methods by ranking error
relative to a ground-truth ranking. Elo is a reliable baseline in those
experiments, while Soft Condorcet Optimization performs better on the real
Atari data. Sequential optimal design \citep{wynn1970sequential} and Bayesian
preference elicitation \citep{viappiani2010optimal,houlsby2011bayesian}
similarly select observations using design or information-gain criteria. The
implemented scheduler instead uses a direct operational rule that maps local
pairwise confidence to the remaining evaluation target.

\section{Continuous Evaluation Setting}
\label{sec:framework}
Let $\games=\{1,\ldots,|\games|\}$ denote the included games and
$\modelset=\{1,\ldots,|\modelset|\}$ the evaluated model versions. A release at time $t$
consists of raw episodes, parsed outcomes, coverage diagnostics, leaderboard
estimates, confidence intervals, and a schedule of future matches. Releases are
produced repeatedly as models, game versions, prompts, parsers, and match
budgets change.

The framework is built around three design requirements.
\begin{description}[leftmargin=*,itemsep=2pt,topsep=2pt]
  \item[\textbf{Repeatability:}] Each release is reproducible from a pinned set of game definitions, model identifiers, prompts, seeds where available, parser versions, and scoring rules.
  \item[\textbf{Comparability:}] Models are compared on a shared scale even when games differ in episode count, native reward scale, role structure, and stochasticity.
  \item[\textbf{Refreshability:}] New models, their  versions, and new games can be incorporated, and existing rankings rechecked, without rerunning a full round-robin tournament.
\end{description}
\section{Adaptive Pairwise Scheduler}
\label{sec:scheduler}

The scheduler operates independently within each game and controls how many
episodes are allocated to each ordered model pair. Rather than selecting one
globally uncertain matchup after every episode, it begins with a base target
$T$ for every ordered pair $(i,j)$ and periodically reduces that pair's
effective target $T_{ij}^{\mathrm{eff}}$ when its fitted ordering becomes
confident. Hence $T_{ij}^{\mathrm{eff}}\leq T$.

Each update follows the same high-level cycle. Completed outcomes are used to
fit a game-specific Bradley--Terry model; the fit yields a confidence $q_{ij}$
for each observed pair; and a cubic target-reduction rule maps that confidence to
$T_{ij}^{\mathrm{eff}}$. The scheduler then compares the effective target
with the number of episodes already allocated and uses their difference to
form target-driven requests. Algorithm~\ref{alg:pairwise-scheduler}
summarizes this update.

Rather than refitting after every completed episode, this update is invoked
periodically in batches for each active game. Let $B$ denote the maximum number
of new episode requests that may be issued in one invocation. When $B>0$ and
early stopping is enabled, the scheduler rebuilds the completed-outcome data
and, when these data are nonempty, refits the confidence estimator and selects
at most $B$ requests from the resulting candidate set. Thus $B$ is an
operational, potentially time-varying throughput limit rather than a per-pair
target; the current deployment's invocation period, throughput limit $B$,
eligibility rules, and capacity filters are described in
\Cref{sec:schedule-op-cycle}.

\subsection{Pairwise Evaluation State}

Fix one game, since scheduling and model fitting are performed separately for
each game. The scheduler considers every ordered pair $(i,j)$ of models that
the platform currently permits to receive new episodes. The order identifies
model position or role, so $(i,j)$ and $(j,i)$ are distinct. Platform-level
eligibility details are given in \Cref{sec:schedule-op-cycle}.

For each ordered pair, let $T\in\mathbb Z_{\ge0}$ be the base target and let
$c_{ij}\in\mathbb Z_{\ge0}$ be the number of public episodes for that pair
that pass the operational filters and have already been allocated with status
\texttt{Created} or \texttt{Completed}. A
\texttt{Created} episode is allocated but does not yet have a final outcome;
a \texttt{Completed} episode has finished. Both statuses count toward
$c_{ij}$, preventing pending work from being requested again, whereas only
completed episodes with usable outcomes enter the confidence fit. Thus the
pairwise evaluation state comprises the target and accounting quantities $T$,
$c_{ij}$, $q_{ij}$, and $T_{ij}^{\mathrm{eff}}$; it is not an episode lifecycle
state. Consequently, an episode that takes longer than one invocation interval
remains pending across scheduler invocations: it counts toward $c_{ij}$ but
does not enter the confidence fit or generate a duplicate request.

For the configuration used here, and before any mirror repair, the
target-driven number of additional requests is
\begin{equation}
  r_{ij}^{\mathrm{target}}
  :=\max\{0,T_{ij}^{\mathrm{eff}}-c_{ij}\}.
  \label{eq:scheduler-deficit}
\end{equation}
Team-game and mirror-repair conventions are given in
\Cref{app:scheduler-team-mirror}.

When a new eligible model is introduced, the scheduler includes both of its
orderings with every existing eligible model. With no observed outcomes, these
pairs use confidence $q_{ij}=0.5$ (defined below), so
$T_{ij}^{\mathrm{eff}}=T$ and they begin with the full base target.

\subsection{Confidence Estimator}

Early stopping uses a Newton--Raphson \bt{} fit
\citep{bradley1952rank} on completed outcomes from the fixed game. Let
\[
  \mathcal D=\{(i_m,j_m,y_m)\}_{m=1}^{n}
\]
denote the $n$ pairwise outcome records derived from completed episodes,
where $i_m$ and $j_m$ are the two models in record $m$ and
$y_m\in\{0,0.5,1\}$ is the score of $i_m$ (loss, tie, or win, respectively).
Thus a tie enters the binary likelihood as half a win and half a loss; unlike
explicit tie extensions \citep{rao1967ties,ddm-with-ties}, the implemented
model does not fit a separate tie parameter.
Let $\modelset_{\mathcal D}$ denote the models that occur in $\mathcal D$, and
let $\theta=(\theta_i)_{i\in\modelset_{\mathcal D}}$ be their vector of
natural-log skill parameters. With $i\succ j$ denoting that $i$ defeats $j$,
the model is
\[
  p_{ij}:=P(i\succ j\mid\theta)
  =\sigmoid(\theta_i-\theta_j),
  \qquad
  \sigmoid(z):=\frac{1}{1+\exp(-z)},\qquad z\in\mathbb R.
\]

Let $\hat\theta$ denote the fitted skill vector. Let $H$ be the Hessian of the
penalized objective at $\hat\theta$, and let $C:=H^{-1}$ be the local
Hessian-based covariance approximation, with $C_{ab}$ denoting its entry for
models $a$ and $b$. The objective, regularization, and numerical
solver are specified in \Cref{app:scheduler-numerics}. For a directly observed
pair $(i,j)$, define the fitted skill difference
$d_{ij}:=\hat\theta_i-\hat\theta_j$ and its approximate
variance
\[
  V_{ij}:=C_{ii}+C_{jj}-2C_{ij}.
\]
The scheduler confidence is
\begin{equation}
  q_{ij}:=\Phi\left(\frac{|d_{ij}|}{\sqrt{V_{ij}}}\right),
  \label{eq:scheduler-confidence}
\end{equation}
where $\Phi$ is the standard normal cumulative distribution function. Thus
$q_{ij}\in[0.5,1]$ and $q_{ij}=q_{ji}$. It measures confidence in the sign of
the fitted difference under the local normal approximation. The implementation
stores this quantity only for pairs represented in $\mathcal D$; for any
other pair, the scheduler uses $q_{ij}=0.5$.

This is a Wald-style normal approximation \citep{Wasserman10All} used as a
scheduling signal rather than a calibrated posterior probability.

The implemented cubic target-reduction rule defines $T_{ij}^{\mathrm{eff}} = T$ if $q_{ij} \le 0.95$; otherwise $T_{ij}^{\mathrm{eff}}$ is reduced monotonically to 0 as $q_{ij}$ approaches 1. 
The precise definition of $T_{ij}^{\mathrm{eff}}$ and a numerical example are given in
\Cref{app:scheduler-cubic-rule}. Together with
\Cref{eq:scheduler-deficit}, this is how greater confidence reduces the
number of additional episodes requested.
Reducing the target does not cancel episodes that have already been created; it
only prevents additional requests. If confidence later decreases, the
effective target can increase and scheduling can resume.

For games in which first-player position, color, seat, or role can matter, the
scheduler can also mirror evaluations by swapping model order under the same
randomly generated environment seed
(\Cref{app:scheduler-team-mirror}). Candidate ordering, batching, and
capacity filters are detailed in \Cref{sec:schedule-op-cycle}.

\begin{algorithm}[H]
\caption{Adaptive pairwise scheduling update}
\label{alg:pairwise-scheduler}
\begin{algorithmic}[1]
\State \textbf{input:} eligible ordered model pairs, base target $T$, per-invocation request limit $B$
\State Count valid \texttt{Created} and \texttt{Completed} episodes as $c_{ij}$
\State Form $\mathcal D$ from valid completed outcomes
\If{$\mathcal D\ne\varnothing$}
  \State Fit the Newton--Raphson \bt{} model and compute available $q_{ij}$
\EndIf
\State Initialize the candidate list $\mathcal Q\gets\varnothing$
\For{each unordered model pair $\{i,j\}$}
  \State Set $q_{ij}\gets0.5$ if no confidence estimate is available
  \State Compute $T_{ij}^{\mathrm{eff}}$ using the cubic target-reduction rule
  \State If mirroring is enabled, add any required mirror-repair requests to $\mathcal Q$
  \State Add the remaining target-driven requests for both orderings,
    accounting for queued repairs
\EndFor
\State Order $\mathcal Q$ and select at most $B$ requests
\State Remove requests blocked by capacity or health filters
\State \Return the remaining requests
\end{algorithmic}
\end{algorithm}

\section{Unified Leaderboard}
\label{sec:ulb}
The Unified Leaderboard (ULB) estimates a single cross-game strength for each
model by fitting a game-balanced \bt{} model~\citep{bradley1952rank} across
all ten games simultaneously.
\subsection{Game-Balanced \bt{} Ratings}
Each completed episode collapses to a single whole-episode pairwise record $(i,j,s_i,g)$ where $s_i\in\{0,0.5,1\}$
is model~$i$'s score (loss, draw, or win, respectively) and $g$ is the game
(see \Cref{sec:ulb-details} for the full record construction).  To prevent
high-volume games from dominating, every record in game~$g$ receives weight
$w_g = 1/N_g$, where $N_g$ is the number of records for that game, so each
game contributes equal total weight to the likelihood.
The \bt{} model assigns each model a latent strength $\gamma_i > 0$ with
$P(i \succ j) = \gamma_i/(\gamma_i + \gamma_j)$.  Strengths are estimated via
the minorization--maximization (MM) algorithm~\citep{hunter2004} on the
game-weighted sufficient statistics.  Fitted strengths are converted to
the Elo scale~\citep{elo1978rating} via $R_i = 400\log_{10}\gamma_i$ and
shifted so the weakest model anchors at $R = 0$ within a single snapshot, or anchored to the frozen mean of unchanged baseline models across multi-wave refreshes so that new entrants do not shift the reference level.  Asymmetric
95\% confidence intervals are obtained from 500 pair-stratified bootstrap resamples~\citep{efron1979,Wasserman10All} (auditing mirror-seed cluster resampling, statistical units, and aggregation rules in \Cref{sec:ulb-details,app:rq3-robustness}).
Per-game leaderboards use the same MM fit without game-balancing weights
(\Cref{sec:per_game_bootstrap}).
\subsection{Empirical Pareto Frontier}
\label{sec:cost-normalization}
To compare models on performance versus cost, we need a cost metric where each
game contributes equally.  Raw per-turn costs vary by orders of magnitude
across games: environments with verbose game states or multi-hand structures
incur far higher costs than compact grid games, so a simple average would be
dominated by the most expensive game.
We normalize costs within each game and aggregate via geometric
mean~\citep{fleming1986geometric}.  For each model~$m$ and game~$g$, let
$\rho_{m,g} = C_{m,g}/\bar{C}_g$ be the cost ratio relative to the benchmark
average $\bar{C}_g = \frac{1}{|\modelset|}\sum_m C_{m,g}$ for that game.  The
model's \emph{Relative Cost Index} is the geometric mean of these ratios:
\[
  I_m = \exp\!\Bigl(\tfrac{1}{|\games|}
        \textstyle\sum_{g\in\games}\ln \rho_{m,g}\Bigr).
\]
An index of $I_m = 1$ means a model is average-cost across all games; $I_m = 2$
means it is consistently twice as expensive.  For display we rescale to
standardized cents via $C_m^{\mathrm{norm}} = I_m \times \bar{C}_{\mathrm{macro}}$,
where $\bar{C}_{\mathrm{macro}} = 7.94$\textcent{} is the macro-average
benchmark turn cost.  This rescaling is a constant applied to all models and
does not affect the Pareto ranking.
The normalization is scale-invariant: scaling costs in any single game (e.g.,
doubling the number of poker hands) scales both $C_{m,g}$ and $\bar{C}_g$
identically, leaving $\rho_{m,g}$ and thus $I_m$ unchanged (freezing initial-release normalizers $\bar{C}_g$ across refreshes for longitudinal comparability).  Models on the Pareto
frontier achieve the highest Elo rating for their cost tier; no other model is
simultaneously cheaper and stronger. The resulting frontier is reported in
\Cref{sec:results-ulb,fig:pareto-frontier}. Per-model normalized costs, which
are proportional to $I_m$ by the constant $\bar C_{\mathrm{macro}}$, are
listed in \Cref{tab:pareto-frontier}.

\section{Game Selection and Skill Profiles}
\label{sec:skills}

A useful evaluation suite must cover meaningfully different capabilities.
We therefore ground game selection in two layers: (i)~formal rule-level
properties: observability, stochasticity, payoff structure, communication
channel, state topology, and planning horizon, and (ii)~a small set of
\emph{construct-demand dimensions} that map those properties to the cognitive
skills required for competent play.
This two-layer approach prevents the taxonomy from becoming an informal
capability list: a game's skill demands follow from its rules, not from
ad-hoc labeling.

We define six constructs that span the space of cognitive demands relevant
to the arena games:  Search and planning, Tactical/constraint reasoning, Spatial/topological reasoning, Uncertainty and belief reasoning, Social/mental-state reasoning, Language and pragmatics. Each is rated on a 0–5 ordinal scale (0 = absent,
5 = defining). Appendix Table~\ref{tab:constructs} lists the full definitions and
scale anchors.

\newcommand{\anchors}[2]{%
  \newline\smallskip{\footnotesize\textit{0:} #1\enspace\textbar\enspace\textit{5:} #2}}

Table~\ref{tab:game-properties} summarizes the rule-level properties of
the ten selected games, spanning perfect and imperfect information,
stochasticity, language, cooperation, and long-horizon planning.

\begin{table*}[t]
\centering
\caption{Rule-level properties of the games in the evaluation suite.
``Solved'' refers to the game-theoretic status of the variant configured in
Game Arena, not to model performance.}
\label{tab:game-properties}
\small
\setlength{\tabcolsep}{6.5pt}
\renewcommand{\arraystretch}{1.15}
\begin{tabular}{@{}llllll@{}}
\toprule
\textbf{Game} & \textbf{Perf.\ info} & \textbf{Solved} & \textbf{Players}
  & \textbf{First-mover adv.} & \textbf{Zero-sum} \\
\midrule
\addlinespace[1pt]
\multicolumn{6}{@{}l}{\itshape Perfect information, two-player} \\
\addlinespace[1pt]
Chess           & Yes & No\textsuperscript{a}          & 2 & Small, White\textsuperscript{b} & Yes \\
Checkers        & Yes & Yes (draw)\textsuperscript{c}  & 2 & None                            & Yes \\
Four in a Row   & Yes & Yes (first player)\textsuperscript{d} & 2 & Decisive, first player   & Yes \\
Dots and Boxes  & Yes & Partial\textsuperscript{e}     & 2 & Parity-dependent\textsuperscript{f} & Yes\textsuperscript{g} \\
Lines of Action & Yes & No\textsuperscript{h}          & 2 & Slight, first player            & Yes \\
\addlinespace[3pt]
\multicolumn{6}{@{}l}{\itshape Imperfect / Incomplete information, two-player} \\
\addlinespace[1pt]
Poker (HUNL)    & No  & No\textsuperscript{i}          & 2 & None\textsuperscript{j}         & Yes \\
Dark Hex        & No\textsuperscript{k} & No\textsuperscript{l} & 2 & Empirical, first player\textsuperscript{v} & Yes \\
Bargaining      & No\textsuperscript{m} & No            & 2 & Small, first proposer           & No\textsuperscript{n} \\
\addlinespace[3pt]
\multicolumn{6}{@{}l}{\itshape Team, intra-model coordination} \\
\addlinespace[1pt]
Word Association & No\textsuperscript{o} & ---          & $2\times2$\textsuperscript{p}
  & Controlled\textsuperscript{q} & Match level\textsuperscript{r} \\
Coin Game  & No\textsuperscript{s} & ---          & $2\times2$
  & None\textsuperscript{t}       & Outcome only\textsuperscript{u} \\
\bottomrule
\end{tabular}

\vspace{4pt}
{\footnotesize
\begin{minipage}{\textwidth}
\raggedright
\textsuperscript{a}\,Seven-piece Syzygy endgame tables are available, but they cover endgames rather than general play \citep{haworth2019syzygy}.
\textsuperscript{b}\,High-level match data show a small positive empirical advantage for White; its magnitude depends on the population and analysis measure \citep{ribeiro2013chess}.
\textsuperscript{c}\,Weakly solved by \citet{schaeffer2007checkers}; optimal play leads to a forced draw.
\textsuperscript{d}\,Weakly solved by \citet{allis}; the first player forces a win on the standard $7\times 6$ board.
\textsuperscript{e}\,Specific small boards have been solved by search; generalized Dots and Boxes is PSPACE-complete \citep{buchin2021dots}.
\textsuperscript{f}\,Advantage depends on board parity and the long-chain rule.
\textsuperscript{g}\,Scored by box margin; draws possible on even total box counts.
\textsuperscript{h}\,Boards through $6\times 6$ have been weakly solved; the configured standard $8\times 8$ variant has no cited exact solution \citep{winands2008loa}.
\textsuperscript{i}\,Heads-Up \emph{Limit} Hold'em is essentially weakly solved \citep{bowling}; agents for HUNL achieve superhuman play without solving the full game \citep{libratus}.
\textsuperscript{j}\,Button acts last post-flop; positional advantage is neutralized by duplicated hands.
\textsuperscript{k}\,Non-abrupt variant: opponent stones revealed only upon collision during a placement attempt.
\textsuperscript{l}\,\citet{darkhex} reports approximate CFR-based strategies and bounds for $4\times 3$ Dark Hex and treats $4\times 4$ as future work; no exact solution is claimed here.
\textsuperscript{m}\,Private item valuation utility functions are hidden; only proposals and chat messages are observable.
\textsuperscript{n}\,General-sum negotiation: positive payoffs upon agreement; mutual zero payoff under disagreement.
\textsuperscript{o}\,Cluemaster observes all secret word-role assignments; Guesser does not.
\textsuperscript{p}\,Four seats, two per team; each team is fielded by instances of a single model.
\textsuperscript{q}\,Starting team has 9 words vs 8; first-mover advantage is balanced by alternating starts across matches.
\textsuperscript{r}\,Zero-sum competition between teams, fully cooperative within a team.
\textsuperscript{s}\,Own board grid is visible; opposing team's board is hidden.
\textsuperscript{t}\,The two team boards are independent parallel rollouts.
\textsuperscript{u}\,Only the cross-team score comparison is zero-sum; underlying coin collection is mixed-motive.
\textsuperscript{v}\,In the benchmark data reported by Kaggle, the first mover won 61.6\% of 2,424 games \citep{kaggle2026darkhex}.
\end{minipage}}
\end{table*}

While the unified Elo leaderboard (Table~\ref{tab:ulb-results}) collapses
performance across games that tax different abilities, construct-demand skill
profiles give a multidimensional view of each model's strengths. To construct
these profiles, $N{=}15$ AI research scientists (specializing in game research or game enthusiasts), none of whom are paper authors, independently rated
the games on the six constructs (Appendix Table~\ref{tab:constructs}; 0--5 scale).
Raters skipped games they did not know, leaving 3--15 raters per game (median~8.5; \Cref{tab:appF-survey}), with squared-difference $\alpha{=}0.766$ and ordinal Krippendorff's $\alpha_{\mathrm{ord}}{=}0.745$ ($95\%$ CI $[0.725, 0.824]$) \citep{krippendorff2011}.
The resulting consensus scores are shown in Appendix Table~\ref{tab:construct-demand}.
For each game~$g$, its six consensus means form a construct-demand vector
$\mathbf e_g\in\mathbb R^6$.

We then combine these game-level demand vectors with per-game
Bradley--Terry ratings from a separate regularized fit
(\Cref{app:skill-profile-method}). Pairwise Elo differences are converted to
expected win probabilities and centered at $0.5$; each resulting margin is
weighted by the corresponding game's construct-demand vector and averaged over
games and opponents. A larger coordinate therefore indicates stronger
relative performance in games rated as more demanding on that construct. For
visualization, scores are rescaled separately within each dimension to span
$[0,1]$ across the evaluated model pool. The complete construction is given in
\Cref{app:skill-profile-method}.

Figure~\ref{fig:skill-profiles-combined} shows the profiles for the ten
highest-ranked models.

\section{Results}
\label{sec:results}
We report results for the three components of the Game Arena pipeline: the
Adaptive Pairwise Scheduler (\S\ref{sec:scheduler}), the Unified Leaderboard
(\S\ref{sec:ulb}), and the game and model skill profiles
(\S\ref{sec:skills}), along with multi-wave continuous refresh dynamics. Following an initial 23-model release ($S_0$: $64{,}251$ episodes across Waves~1--2), 16 additional models entered across 6 post-$S_0$ release waves (Waves~3--8; 8 chronological waves in total); results cover all 39 active models ($S_1$: $147{,}015$ episodes) from four providers (OpenAI, Anthropic, Google, and xAI) across 10 games.

\subsection{Scheduler Efficiency and Ground-Truth Accuracy}
\label{sec:results-scheduler}

The adaptive scheduler uses Hessian-based pairwise confidence to decide which
matchups still need data and which can be skipped because the relative ordering
is already well established (\S\ref{sec:scheduler}). Across the ten games and 39 active models, the
evaluation contains $147{,}015$ played episodes instead of the $384{,}200$
required by the uniform per-game baselines, a reduction of $237{,}185$ episodes
(61.7\%; 60.3\% in the initial 23-model $S_0$ release). Per-game savings range from 50.5\% to 72.7\%
(\Cref{tab:scheduler-savings}); at token-level API pricing, actual evaluation spend was \$1.438M versus \$3.807M under uniform caps, a 62.2\% reduction (\Cref{tab:appG-dollars}).
These counts are for the deployed scheduler, which treated each $2\times 2$ team episode as four records in the two team games, overstating their savings (\Cref{tab:scheduler-savings}, note~f); the eight two-player games, unaffected, save 61.9\% of episodes and 62.3\% of cost.
The baseline counts a pair in a game only if the two models were ever scheduled against each other there: six models were retired in early October 2026 as newer models arrived, so 293 of the $6{,}972$ pair--game combinations were never scheduled and are not counted as savings (\Cref{app:rq2-continuous-refresh}).
Appendix Figure~\ref{fig:scheduler-heatmaps} visualizes the corresponding matchup-level allocations, where closely contested pairs ($|\Delta\text{Elo}|\le 40$) use 95.6\% of their configured cap compared with 15.5\% for $|\Delta\text{Elo}|\ge 500$.

\begin{table}[t]
\centering
\caption{Deployed-scheduler savings against a uniform allocation baseline across the 39-model suite ($147{,}015$ episodes; $S_0$ 23-model snapshot saved $60.3\%$). Row order matches Table~\ref{tab:game-properties}.}
\label{tab:scheduler-savings}
\footnotesize
\setlength{\tabcolsep}{6pt}
\renewcommand{\arraystretch}{1.02}
\begin{tabular}{@{}lrrrrrrr@{}}
\toprule
\textbf{Game} & \textbf{Models\textsuperscript{a}} & \textbf{Pairs\textsuperscript{b}} & \textbf{Cap\textsuperscript{c}} & \textbf{Played} & \textbf{Baseline\textsuperscript{d}} & \textbf{Saved} & \textbf{\% saved\textsuperscript{e}} \\
\midrule
\multicolumn{8}{@{}l}{\itshape Perfect information, two-player} \\
Chess              & 39 & 711 &  40 &  8838 &  28440 & 19602 & 68.9 \\
Checkers           & 39 & 711 &  40 &  9470 &  28440 & 18970 & 66.7 \\
Four in a Row      & 39 & 699 &  80 & 15256 &  55920 & 40664 & 72.7 \\
Dots and Boxes     & 39 & 711 &  40 & 12020 &  28440 & 16420 & 57.7 \\
Lines of Action    & 39 & 711 &  40 &  9928 &  28440 & 18512 & 65.1 \\
\multicolumn{8}{@{}l}{\itshape Imperfect information, two-player} \\
Poker (HUNL)       & 28 & 378 & 200 & 29932 &  75600 & 45668 & 60.4 \\
Dark Hex           & 39 & 711 &  40 & 12314 &  28440 & 16126 & 56.7 \\
Bargaining         & 39 & 715 &  80 & 28326 &  57200 & 28874 & 50.5 \\
\multicolumn{8}{@{}l}{\itshape Team, intra-model coordination\textsuperscript{f}} \\
Word Association   & 37 & 624 &  40 &  9612 &  24960 & 15348 & 61.5 \\
Coin Game          & 39 & 708 &  40 & 11319 &  28320 & 17001 & 60.0 \\
\midrule
\textbf{Total} & 39 & \textbf{6679} & & \textbf{147015} & \textbf{384200} & \textbf{237185} & \textbf{61.7} \\
\bottomrule
\end{tabular}

\vspace{2pt}
{\footnotesize
\begin{minipage}{\textwidth}
\raggedright
\textsuperscript{a}\,Active coverage is $39$ models in $8$ games, $37$ in Word Assoc., and $28$ in Poker, whose public dataset was last updated on Oct~8 (Nov~11 for the other nine games); \textsuperscript{b}\,pairs scheduled at least once, out of $\binom{n_g}{2}$; the other 293 pair--game combinations involve a model retired before the pair could be scheduled (\Cref{app:rq2-continuous-refresh}); \textsuperscript{c}\,mirrored cap $2T$; \textsuperscript{d}\,pairs $\times$ cap; \textsuperscript{e}\,Poker saves 60.4\% (\$1.642M at \$35.51/ep) while using 79.2 eps/pair vs.\ 12.4 in Chess and 16.0 in Coin Game; \textsuperscript{f}\,the deployed scheduler counted each $2\times 2$ team episode as four records, overstating these savings; the ten-game floors with one record are 53.3\% (episodes) and 61.0\% (cost) (\Cref{app:rq2-continuous-refresh}).
\end{minipage}}
\end{table}

The per-pair budget itself is game-dependent: most games use a target of 20
episodes per ordered pair, while Four in a Row and Bargaining use 40 and Poker (HUNL)
uses 100. These are operational settings rather than statistically derived
sample sizes. Poker alone accounts for 20.4\% of scheduled episodes ($29{,}932$) and 72.5\% of API spend due to its 100-hand horizon ($257.6$ calls/side).

Ideally, we would compare the adaptive scheduler against a complete round-robin on real gameplay across every game; due to API budget constraints (full round-robins across the ten-game suite would cost an additional \$2.369M; \Cref{tab:appG-dollars}), we use a complete real-game round-robin from an earlier Chess tournament ($18$ models, $6{,}120$ episodes; \Cref{app:chess-replay}). Replaying the scheduler saves $76.1\%$ of episodes ($1{,}462$ vs.\ $6{,}120$) while remaining highly correlated with the complete round-robin ($\rho = 0.9965$, $\tau_K = 0.9780$, vs.\ $\tau_K = 0.9585$ for uniform allocation at the same budget): unlike uniform round-robin, which provides no stopping criterion and only matches the $23.9\%$ budget here in hindsight, the scheduler halts automatically once pairwise orderings reach confidence ($q_{ij}>0.95$), nearly halves remaining pairwise ordering error ($1.10\%$ vs.\ $2.08\%$), recovers the same top-five models in $100.0\%$ vs.\ $54.5\%$ of runs, and reaches a $\tau_K$ that uniform sampling matches only at $2.5\times$ the budget ($60\%$ of the tournament). We then run controlled ground-truth simulations across $N \in \{10, 20, 30, 40, 50\}$ models ($\mathrm{SD}=220$ Elo, $2T=40$; \Cref{tab:ground-truth-scheduler-sim}, \Cref{app:rq1-scheduler-validation}), confirming that episode savings grow from 61.6\% to 76.3\% via shared transitive comparisons in $C=H^{-1}$ (\Cref{fig:scaling-laws-main}) while maintaining rank accuracy ($\rho = 0.9744\text{--}0.9946$ against true skill; $\rho = 0.9681\text{--}0.9935$ directly against Full Round-Robin; \Cref{tab:gt-direct-rr,fig:appG-gt-scatter}). Beyond initial warm-up rounds ($\ge 15\%$ of cap), the scheduler outperforms smaller fixed-target Round-Robin at matched budgets across all $N$ (\Cref{fig:appG-gt-frontier}), retains 69.6\%--77.2\% savings across five skill, draw, seat-bias, and high-noise regimes (\Cref{tab:appG-gt-regimes}), and achieves 99.87\% true-sign accuracy among pairs with $q_{ij}>0.95$ (\Cref{fig:appG-gt-calib}).

\begin{table}[htbp]
\centering
\footnotesize
\caption{Ground-truth tournament simulation comparing Full Round-Robin ($2T=40$) and the Adaptive Pairwise Scheduler ($\tau_0=0.95$) across $N \in \{10, 20, 30, 40, 50\}$ models (\Cref{app:rq1-scheduler-validation}).}
\label{tab:ground-truth-scheduler-sim}
\resizebox{\textwidth}{!}{%
\begin{tabular}{@{}lrrrrrrrr@{}}
\toprule
\textbf{$N$ Models} & \textbf{RR Eps} & \textbf{Scheduler Eps} & \textbf{\% Saved} & \textbf{Sched $\rho$ (True)} & \textbf{RR $\rho$ (True)} & \textbf{Sched $\tau_K$ (True)} & \textbf{RR $\tau_K$ (True)} & \textbf{Direct $\rho$ (Sched vs.\ RR)} \\
\midrule
$N=10$ & 1{,}800 & \textbf{691} & \textbf{61.6\%} & \textbf{0.9744} & 0.9811 & \textbf{0.9238} & 0.9425 & \textbf{0.9681} \\
$N=20$ & 7{,}600 & \textbf{2{,}391} & \textbf{68.5\%} & \textbf{0.9877} & 0.9927 & \textbf{0.9395} & 0.9589 & \textbf{0.9835} \\
$N=30$ & 17{,}400 & \textbf{4{,}969} & \textbf{71.4\%} & \textbf{0.9917} & 0.9947 & \textbf{0.9470} & 0.9631 & \textbf{0.9890} \\
$N=40$ & 31{,}200 & \textbf{7{,}820} & \textbf{74.9\%} & \textbf{0.9941} & 0.9969 & \textbf{0.9543} & 0.9714 & \textbf{0.9928} \\
$N=50$ & 49{,}000 & \textbf{11{,}625} & \textbf{76.3\%} & \textbf{0.9946} & 0.9974 & \textbf{0.9531} & 0.9729 & \textbf{0.9935} \\
\bottomrule
\end{tabular}}
\end{table}

\subsection{Continuous Refresh Across Model Releases ($23 \to 39$ Models)}
\label{sec:results-refresh}
Across the live expansion from $S_0$ (23 models, Waves~1--2) to $S_1$ (39 models after 6 post-$S_0$ waves, Waves~3--8; \Cref{app:rq2-continuous-refresh}, \Cref{tab:appH-decomp,tab:appH-stability,tab:appH-waves,tab:appH-full39} and \Cref{fig:refresh-main}), the 15 unchanged $S_0$ submissions retained $28{,}573$ historical episodes (\$367.2k) and required $2{,}081$ ($1.76\%$) confidence re-checks between existing pairs (\$11.7k), while 8 models were re-evaluated under an updated unified harness ($34{,}725$ episodes). Onboarding the 16 new models used $81{,}636$ episodes (\$639.1k) compared with $223{,}160$ (\$1.806M) under uniform incremental caps, saving 63.4\% of onboarding episodes and 64.6\% of API cost (\$1.167M saved; 53.1\%--65.6\% per wave), while preserving Kendall's $\tau_K = 0.9619$ ($\rho = 0.9929$, maximum rank shift 1) among unchanged $S_0$ models. Simulated single-model hold-out insertions across all 39 models (\Cref{fig:appH-waves-holdout}) save 74.4\% of episodes on average and end closer on average to each model's full-data rank than equal-budget uniform allocation (1.31 vs.\ 2.74 ranks).

\subsection{Unified Leaderboard}
\label{sec:results-ulb}

Table~\ref{tab:ulb-results} presents the full 39-model Unified Leaderboard, ranked by
game-balanced Elo (anchored to the frozen $S_0$ cohort mean) with 95\% bootstrap confidence intervals computed from 500
resamples.

\begin{table*}[t]
\centering
\begin{minipage}[t]{0.52\textwidth}
  \vspace{0pt}
  \caption{Unified Game-Balanced Bradley--Terry Elo ratings for all 39 active models with 95\% bootstrap CIs (500 resamples; anchored to frozen $S_0$ cohort mean).}
  \label{tab:ulb-results}
  \fontsize{6.3pt}{7.05pt}\selectfont
  \setlength{\tabcolsep}{2.5pt}
  \begin{tabular}{@{}lrrrr@{}}
  \toprule
  \textbf{Model} & \textbf{Elo} & $\text{CI}_{\text{lo}}$ & $\text{CI}_{\text{hi}}$ & \textbf{Episodes} \\
  \midrule
  Gemini 4 Carbon        & 424.1 & 416.2 & 431.6 & 3,739 \\
  Gemini 4 Argon         & 361.1 & 354.3 & 368.1 & 5,488 \\
  Claude Fable 5.2       & 356.1 & 348.6 & 363.5 & 4,759 \\
  GPT-6 Astra            & 313.3 & 306.1 & 320.8 & 6,125 \\
  GPT-5.5                & 283.6 & 276.5 & 290.1 & 7,410 \\
  GPT-6.1 Sol            & 278.4 & 271.6 & 285.2 & 5,823 \\
  Gemini 3.8 Flash       & 266.1 & 259.3 & 272.8 & 8,722 \\
  Claude Opus 5          & 248.2 & 242.2 & 254.1 & 8,474 \\
  Claude Fable 5.1       & 231.7 & 225.0 & 237.9 & 7,880 \\
  Gemini 3.1 Pro Preview & 227.5 & 221.5 & 234.0 & 8,771 \\
  Claude Opus 5.5        & 226.3 & 218.6 & 234.6 & 5,481 \\
  Gemini 3.7 Flash       & 222.2 & 215.2 & 229.4 & 7,678 \\
  GPT-5.6 Sol            & 221.8 & 215.9 & 228.4 & 9,028 \\
  GPT-6.1 Terra          & 210.6 & 203.2 & 217.5 & 6,195 \\
  GPT-5.4                & 208.1 & 201.4 & 213.8 & 10,636 \\
  GPT-6 Sol              & 201.7 & 194.3 & 208.1 & 6,669 \\
  Gemini 3.5 Flash       & 201.4 & 192.8 & 210.2 & 5,323 \\
  Gemini 3.6 Flash       & 201.4 & 195.0 & 206.5 & 9,037 \\
  Grok 4.7               & 195.3 & 189.0 & 201.9 & 7,501 \\
  Grok 4.6               & 191.9 & 185.8 & 198.5 & 10,178 \\
  Claude Sonnet 5.5      & 183.1 & 175.5 & 190.1 & 5,796 \\
  Claude Opus 4.8        & 178.2 & 172.0 & 184.8 & 9,695 \\
  Gemini 3 Flash Preview & 177.6 & 171.4 & 183.9 & 9,134 \\
  GPT-5.2                & 176.8 & 169.9 & 183.2 & 9,840 \\
  o3                     & 164.5 & 157.9 & 171.0 & 9,210 \\
  Grok 4.5               & 147.5 & 141.6 & 152.8 & 10,774 \\
  Claude Opus 4.6        & 146.9 & 140.0 & 153.5 & 9,026 \\
  Claude Sonnet 5        & 134.1 & 127.4 & 140.5 & 8,901 \\
  Claude Opus 4.7        & 122.4 & 115.8 & 128.7 & 8,041 \\
  Claude Haiku 5.5       & 119.2 & 110.9 & 127.1 & 5,872 \\
  GPT-5.6 Terra          & 114.3 & 107.8 & 120.5 & 8,501 \\
  Grok 4.20 Reasoning    & 106.4 & 100.3 & 114.2 & 6,738 \\
  GPT-5.6 Luna           & 104.1 &  96.4 & 111.4 & 9,019 \\
  Claude Sonnet 4.6      & 100.3 &  93.3 & 107.2 & 7,070 \\
  Claude Sonnet 4.5      &  94.9 &  87.6 & 102.4 & 7,458 \\
  GPT-5 mini             &  86.5 &  79.1 &  94.4 & 5,838 \\
  GPT-5.4 mini           &  81.4 &  74.2 &  88.6 & 6,999 \\
  Grok 4.3               &  80.9 &  73.3 &  88.2 & 5,883 \\
  GPT-6 Luna             &  77.8 &  69.4 &  86.0 & 5,318 \\
  \bottomrule
  \end{tabular}
\end{minipage}
\hfill
\begin{minipage}[t]{0.45\textwidth}
  \vspace{0pt}
  \centering
  \vspace{0.7cm}
  \includegraphics[width=\linewidth,keepaspectratio]{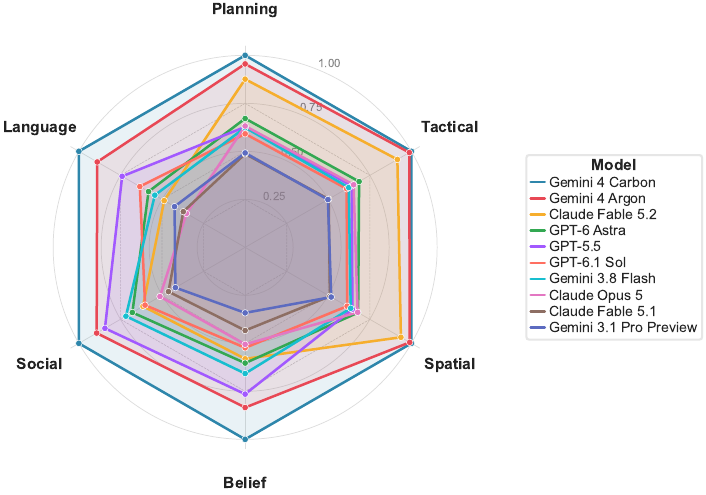}
  \captionof{figure}{Probability-weighted skill profiles for the ten highest-ranked models, normalized to $[0,1]$ within each dimension (Eq.~\ref{eq:norm}).}
  \label{fig:skill-profiles-combined}
\end{minipage}
\end{table*}

Gemini 4 Carbon leads the ranking with an Elo of 424.1, establishing a clear separation
of ${\sim}63.0$ points above Gemini 4 Argon (361.1) and Claude Fable~5.2 (356.1), followed by GPT-6 Astra (313.3) and former $S_0$ leader GPT-5.5 (283.6; 310.6 in $S_0$).
A dense mid-tier spans from GPT-6.1 Terra (210.6) through GPT-5.2 (176.8),
with successive models separated by at most 9 Elo points where
bootstrap confidence intervals frequently overlap, reflecting the fact that
small strength differences require substantially more evidence to statistically resolve relative rank. At the bottom of the ranking, GPT-6 Luna sits at Elo~77.8 ($[69.4, 86.0]$). Sensitivity checks on resampling units and cross-game weighting (\Cref{sec:ulb-details,app:rq3-robustness}) show that mirroring-aware episode cluster bootstrap yields a mean 95\% CI width of 13.82 Elo ($693/741$ separated pairs), and rankings remain consistent across six alternative aggregation rules ($\tau_K \in [0.9163, 0.9460]$, with at least four of the top five models retained).

\begin{figure}[t]
\centering
\includegraphics[width=0.95\linewidth,height=0.27\textheight,keepaspectratio]{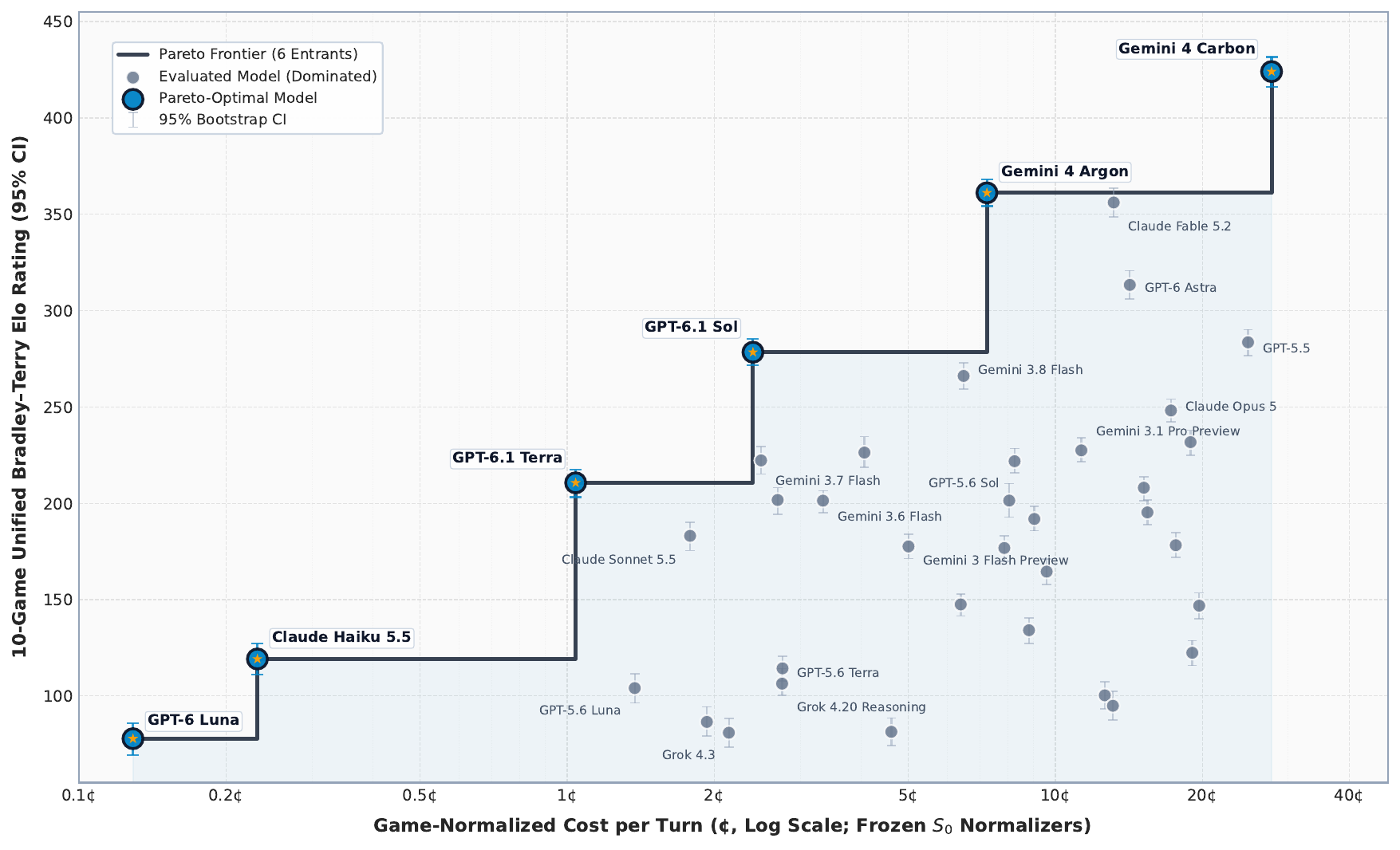}
\caption{Cross-game performance vs.\ execution cost Pareto frontier across all 39 foundation models ($147{,}015$ episodes). Performance is ten-game Unified Bradley--Terry Elo with 95\% bootstrap CIs; cost is game-normalized cost per turn (\textcent, log scale; frozen $S_0$ normalizers). Non-dominated Pareto-optimal models form the stepped frontier (blue markers with gold stars); dominated models are depicted in gray.}
\label{fig:pareto-frontier}
\end{figure}

As shown in Figure~\ref{fig:pareto-frontier} and Appendix Table~\ref{tab:pareto-frontier}, 6 post-$S_0$ models define the empirical Pareto frontier: Gemini 4 Carbon ($424.1$ Elo, $27.74\text{\textcent}$/turn), Gemini 4 Argon ($361.1$ Elo, $7.24\text{\textcent}$), GPT-6.1 Sol ($278.4$ Elo, $2.40\text{\textcent}$), GPT-6.1 Terra ($210.6$ Elo, $1.04\text{\textcent}$), Claude Haiku~5.5 ($119.2$ Elo, $0.23\text{\textcent}$), and GPT-6 Luna ($77.8$ Elo, $0.13\text{\textcent}$). Game normalization places GPT-5.6 Terra ahead of Grok 4.20 Reasoning ($114.3$ vs.\ $106.4$ Elo at $2.76\text{\textcent}$/turn): although Grok 4.20 has lower unnormalized costs in four games (Checkers, Chess, Lines of Action, and Poker), GPT-5.6 Terra is more cost-effective in 6 of the 10 environments and achieves higher unified Elo.

\subsection{Game and Model Skill Profiles}
\label{sec:results-skill-profiles}

Figure~\ref{fig:skill-profiles-combined} shows the derived model skill profiles.
The suite illustrates the complementarity of the benchmarks: strategic board
games emphasize planning, tactical, and spatial demands, whereas Bargaining
and Poker emphasize belief and social reasoning, and Word Association emphasizes
language and social reasoning.

In the normalized model profiles, Gemini 4 Carbon reaches the upper end of all six
dimensions, while the remaining top-ranked models exhibit more differentiated
profiles. These profiles complement the scalar Unified Leaderboard by showing
where relative performance differs: for instance, Claude Opus~5 and GPT-5.6 Sol had nearly identical unified ratings in $S_0$ ($247.0$ vs.\ $243.3$ Elo), yet Opus~5 scores higher on planning, tactical, and spatial dimensions, winning every combinatorial board matchup except Chess ($54\text{--}1\text{--}13$) while expending $1.7\times\text{--}2.9\times$ more reasoning tokens per turn, whereas Sol scores higher on belief, social, and language dimensions, winning Poker ($118\text{--}82$) and Bargaining ($8\text{--}12\text{--}2$) while expending $2.11\times\text{--}2.15\times$ more reasoning tokens per turn (\Cref{app:rq4-skill-specialization}, \Cref{tab:appJ-opussol}). Because normalization is performed
separately within each dimension, the model coordinates are relative to the
evaluated pool and should not be interpreted as absolute capability scores.

\section{Conclusion and Limitations}
\label{sec:conclusion}
This paper contributes a methodology for operating Game Arena as a long-term
evaluation instrument, rather than the initial platform itself. The framework
combines adaptive scheduling for cost control, game-balanced aggregation for
cross-game comparability, and construct-demand profiling for interpretation
and suite design.

Across 39 models and ten games ($147{,}015$ episodes), the deployed scheduler used 61.7\% fewer
episodes (62.2\% lower API cost; 61.9\% and 62.3\% in two-player games) than uniform allocation at the configured per-game caps, with
per-game savings of 50.5--72.7\% (Table~\ref{tab:scheduler-savings}) and 63.4\% onboarding savings across the $23\to 39$ model refresh. This is
not a uniform reduction: closely matched pairs retain more evaluation, whereas
targets for confidently ordered pairs are reduced. Because targets are
recomputed as new outcomes arrive, evaluation can resume if confidence later
decreases. Introducing a model requires new comparisons with the existing pool
rather than rerunning the full tournament, which makes repeated releases
practical as the model population grows.

The Unified Leaderboard converts heterogeneous game outcomes to pairwise
evidence and gives every game equal total weight in a single Bradley--Terry
fit. This avoids directly averaging incompatible native metrics and prevents
high-volume games from dominating the aggregate. Bootstrap intervals quantify
sampling uncertainty, while the per-game
leaderboards retain game-specific performance information. Combining unified
ratings with game-normalized inference costs identified a six-model empirical
Pareto frontier whose costs spanned two orders of magnitude ($0.13$\textcent--$27.74$\textcent/turn), with mid-tier
models achieving strong performance at a fraction of flagship cost. The construct-demand profiles provide a complementary
multidimensional summary based on six expert-rated dimensions, helping identify
both model-specific performance patterns and gaps in the capability coverage
of the game suite. Because the profiles are derived from existing match data,
they can be recomputed when the construct ratings or game set change without
running additional episodes.

Several limitations qualify these findings. The cubic reduction rule and
game-specific base targets are operational heuristics rather than outputs of a
formal optimal-design or power calculation, and the scheduler allocates effort
within games rather than jointly across the suite. Agreement with a full
round-robin on recorded games was checked only for Chess, in an earlier
tournament. The Unified Leaderboard
assumes that one transitive latent strength can summarize performance across
games and weights games equally regardless of redundancy or discriminative
power. An alternative based on normalized per-game ratings relaxes this
single-strength assumption but introduces additional distributional choices
(e.g., the choice of robust $z$-score transformation);
geometry-aware weighting may offer another way to address redundant inputs
\citep{Kaisers25Spectral}. All reported ratings remain conditional on the
model pool, game set, filtering policy, and observed episodes. The normalized
cost measure is likewise pool-dependent, so the Pareto frontier can change
across releases. Finally, the skill profiles are relative summaries derived
from expert judgments ($N=15$, 3--15 responses per game,
ordinal Krippendorff's $\alpha_{\mathrm{ord}}=0.745$, squared-difference $\alpha=0.766$) and dimension-wise normalization, not absolute
measurements of model capability.

Future work should therefore examine
principled target selection, joint allocation across games, aggregation models
that relax cross-game transitivity, repeated expert elicitation, and broader
coverage of imperfect-information, multi-party, and mixed-motive games. These
extensions would further support repeatable, comparable, and refreshable
evaluation as model and game populations evolve.

Disclaimer. The metrics and visual results presented in this paper reflect specific game environment configurations tested under controlled conditions. They do not constitute general commercial product claims.
\clearpage

\clearpage
\appendix
\raggedbottom
\counterwithin{table}{section}
\counterwithin{figure}{section}
\setcounter{topnumber}{5}
\setcounter{bottomnumber}{5}
\setcounter{totalnumber}{8}
\renewcommand{\topfraction}{0.95}
\renewcommand{\bottomfraction}{0.95}
\renewcommand{\textfraction}{0.05}
\renewcommand{\floatpagefraction}{0.65}
\captionsetup{font=small}
\captionsetup[table]{skip=4pt}

\section{Scheduler Implementation Details}
\label{app:scheduler-implementation}

\subsection{Operational Cycle}
\label{sec:schedule-op-cycle}

In the current deployment, the scheduling handler runs separately for each
active competition approximately every four minutes. One invocation is a batch
update (Algorithm~\ref{alg:pairwise-scheduler}); the interval is not a model
parameter and does not imply one update per episode. At the implementation
level, scheduling operates on \emph{submissions}: runnable platform agent
entries associated with model versions. Before each invocation, the
per-invocation request limit $B$ is computed from the total number of
submissions, configured daily run rate, four-minute period, and number of
agents per episode, then capped by the remaining concurrent-episode capacity
and a platform safety limit.

For a fixed
competition, the platform's scheduling-eligibility filter determines the
submission IDs whose ordered Cartesian product defines the candidate pairs.
Episode eligibility is then applied to existing episode records. An episode
contributes to $c_{ij}$ when it is public, has status \texttt{Created} or
\texttt{Completed}, contains the eligible submissions in the representative
positions, and is not associated with a configured invalid harness session.
Only valid public completed episodes with eligible participating submissions
and non-null rewards are converted to outcome records in $\mathcal D$.

Candidate requests are ordered by model-proxy health, request round, seed, and
a random key before the batch limit $B$ is applied. Capacity and health filters
can reduce the returned batch but do not alter
\Cref{eq:scheduler-confidence,eq:cubic-target}. The implementation also
supports deterministic seeded scheduling, but that option was not used for
the experiments reported here.

\subsection{Cubic Target-Reduction Rule}
\label{app:scheduler-cubic-rule}

Let $T$ be the base target per ordered pair and let $q_{ij}$ be the confidence
from \Cref{eq:scheduler-confidence}. The effective total target is
\begin{equation}
  T_{ij}^{\mathrm{eff}}
  =\left\lfloor \left(1-x(q_{ij})^3\right)T\right\rfloor,
  \qquad
  x(q)=\mathrm{clip}\left(\frac{q-0.95}{0.05},0,1\right)
      =\mathrm{clip}\left(20(q-0.95),0,1\right).
  \label{eq:cubic-target}
\end{equation}
Here
$\mathrm{clip}(u,0,1):=\min\{1,\max\{0,u\}\}$ restricts its argument to
$[0,1]$. The target is unchanged for $q_{ij}\le0.95$ and decreases
monotonically above this value. The target-driven request deficit then follows
\Cref{eq:scheduler-deficit}; mirror-repair requests are handled separately as
described in \Cref{app:scheduler-team-mirror}.

The calculation in \Cref{eq:cubic-target} has three steps. First, $x(q)$
rescales the confidence range from 0.95 to 1.00 onto $[0,1]$: $q\le0.95$
gives $x=0$ and $q=1$ gives $x=1$. Second, $1-x(q)^3$ is the fraction of the
base target retained, not the fraction saved. Third, the scheduler multiplies
this fraction by $T$ and rounds down.

\begin{table}[!htbp]
\centering
\caption{Implemented cubic target-reduction heuristic for base target $T=100$
episodes per ordered pair.}
\label{tab:cubic-ramp}
\small
\begin{tabular}{@{}lrrr@{}}
\toprule
$q_{ij}$ & Fraction retained & $T_{ij}^{\mathrm{eff}}$ & Target reduction \\
\midrule
$0.50\le q_{ij}\le0.95$ & 1.000 & 100 & 0 \\
0.96 & 0.992 & 99 & 1 \\
0.97 & 0.936 & 93 & 7 \\
0.98 & 0.784 & 78 & 22 \\
0.99 & 0.488 & 48 & 52 \\
1.00 & 0.000 & 0 & 100 \\
\bottomrule
\end{tabular}
\end{table}

\Cref{tab:cubic-ramp} evaluates the rule at selected confidence values. The
second column gives the fraction of the base target retained, the third gives
the integer effective target after rounding down, and the final column gives
the reduction relative to $T$. The reduction in the target need not equal the
number of episodes saved because episodes may already have been allocated when
confidence is recomputed.

For example, $q_{ij}=0.98$ gives $x=0.60$ and retained fraction
$1-0.60^3=0.784$, hence
\[
  T_{ij}^{\mathrm{eff}}=\lfloor0.784\times100\rfloor=78.
\]
Before any mirror repair, if 60 episodes have already been allocated, the
scheduler requests 18 more; if at least 78 have been allocated, it requests
none.

\subsection{Scheduler Newton--Raphson Solver}
\label{app:scheduler-numerics}

For the data and Bradley--Terry probabilities defined in the main text, the
implementation minimizes
\begin{equation}
  \mathcal J(\theta)
  =-\sum_{m=1}^{n}
    \left[y_m\log p_{i_m,j_m}+(1-y_m)\log(1-p_{i_m,j_m})\right]
    +\frac{1}{2\sigma_{\mathrm{prior}}}
      \sum_{i\in\modelset_{\mathcal D}}\theta_i^2,
  \label{eq:scheduler-bt-objective}
\end{equation}
where $\sigma_{\mathrm{prior}}>0$ is named \texttt{sigmaPrior} in the code.
Because the objective uses the coefficient $1/(2\sigma_{\mathrm{prior}})$, it
acts as a precision $1/\sigma_{\mathrm{prior}}$; under a Gaussian-prior
interpretation, $\sigma_{\mathrm{prior}}$ is a variance rather than a standard
deviation. The default $\sigma_{\mathrm{prior}}=1$ corresponds to independent
standard-normal anchors for the components of $\theta$. The quadratic term
makes the Hessian positive definite, including for a disconnected comparison
graph or completely separated outcomes.

Let $t\in\{0,1,\ldots\}$ index Newton iterations for the objective
$\mathcal J(\theta)$ (\Cref{eq:scheduler-bt-objective}), with
$\theta^{(0)}=0$. At iteration $t$,
define
\[
  g^{(t)}:=\nabla\mathcal J(\theta^{(t)}),\qquad
  H^{(t)}:=\nabla^2\mathcal J(\theta^{(t)}),\qquad
  \Delta\theta^{(t)}:=-(H^{(t)})^{-1}g^{(t)}.
\]
The fit is declared converged when
\begin{equation}
  \lVert\Delta\theta^{(t)}\rVert_\infty
  :=\max_{i\in\modelset_{\mathcal D}}|\Delta\theta_i^{(t)}|
  <\varepsilon_{\mathrm{NR}},
  \qquad \varepsilon_{\mathrm{NR}}=10^{-6}.
  \label{eq:scheduler-newton-convergence}
\end{equation}
Otherwise, $\theta^{(t+1)}=\theta^{(t)}+\Delta\theta^{(t)}$. The implementation
allows at most $I_{\max}=20$ iterations and raises an error if
\Cref{eq:scheduler-newton-convergence} has not been satisfied. Here $I_{\max}$
is a computational cap, whereas $\varepsilon_{\mathrm{NR}}$ is a numerical
tolerance in natural-log skill units. At convergence, every proposed per-model
update is smaller than $10^{-6}$, or approximately $1.74\times10^{-4}$ Elo
under the conventional scaling $400/\log(10)$. These values are implementation
defaults, not statistically estimated quantities or confidence thresholds.

\subsection{Team and Mirror Conventions}
\label{app:scheduler-team-mirror}

For an implemented homogeneous team game, let $k\in\{1,2,\ldots\}$ denote the
number of agent slots per team. Model $i$ fills all $k$ slots on the first
team and model $j$ fills all $k$ slots on the second team. The
pairwise state remains keyed by $(i,j)$; the notation does not cover
heterogeneous teams composed of different models.

The outcome conversion operates on agent positions. Consequently, a completed
homogeneous $k$-versus-$k$ team episode produces $k^2$ identical cross-team
Bradley--Terry records after same-model and within-team comparisons are
discarded. The confidence fit therefore weights such an episode by $k^2$
relative to a 1-versus-1 episode. This implementation convention is one reason
to interpret $q_{ij}$ as a scheduling signal rather than as a calibrated
probability across game formats.

For games where first-player position, color, seat, or role can matter, the
scheduler can use mirrored orderings. A mirrored pair schedules both $(i,j)$
and $(j,i)$ with the same randomly generated environment seed, so paired
outcomes share the same random or initial-state condition while swapping model
position. This holds the environment condition fixed across the two
orientations.

The four-minute period is not an episode timeout. An allocated orientation that
is still \texttt{Created} at the next invocation counts toward its ordered pair
and is not requested again, but it enters the confidence fit only after becoming
\texttt{Completed}. If both orientations are pending, neither is duplicated.
If only one orientation of a randomly generated environment seed has been
allocated, the missing orientation remains in the candidate set regardless of
the current confidence-reduced target.

The final selection sorts healthy before unhealthy candidates, followed by
request round, seed, and a random key. The two orientations of a mirrored pair
share the random key and therefore remain adjacent within the same sorting
bucket, reducing the chance that a batch boundary separates them. If the
boundary does separate them, the missing orientation remains eligible for
mirror repair in a later invocation. Mirror repair is thus preserved as
eligible work but does not have unconditional priority over every fresh request.

\begin{figure}[p]
\centering
\includegraphics[width=\linewidth,height=0.85\textheight,keepaspectratio]{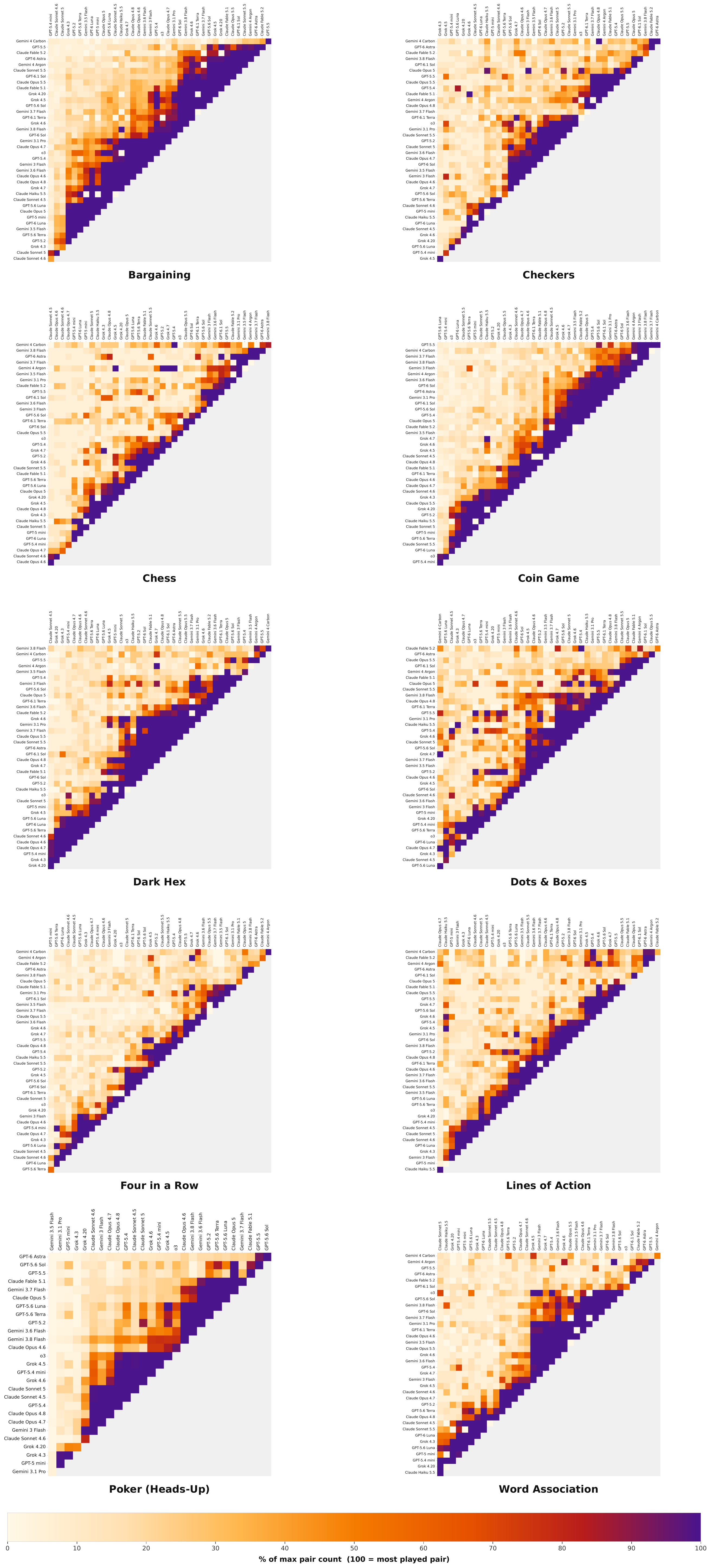}
\caption{Adaptive scheduler matchup allocation per game. Each cell shows the
fraction of the per-pair episode budget consumed (0\%--100\%). The non-uniform allocation pattern that concentrated along
the diagonal where models are closely matched demonstrates that the scheduler
focuses evaluation resources on the most informative matchups. Empty cells are pairs that were never
scheduled: each involves one of the six models retired in early October 2026 (o3, GPT-5 mini, Grok 4.20
Reasoning, Claude Sonnet~4.5, Gemini 3.5 Flash, and Gemini 3.7 Flash). These pairs are excluded from the uniform
baseline of \Cref{tab:scheduler-savings} (\Cref{app:rq2-continuous-refresh}).}
\label{fig:scheduler-heatmaps}
\end{figure}

\section{Unified Leaderboard: Technical Details}
\label{sec:ulb-details}
This appendix provides the full specification of the pairwise record
construction, MM fitting procedure, and bootstrap protocol summarized in
\S\ref{sec:ulb}.
\subsection{Pairwise Record Construction and Statistical Unit Across Pipeline Stages}
In the Unified Leaderboard and per-game MM estimators, each completed valid episode $e$ in game $g$ collapses to \emph{one} whole-episode pairwise outcome record $m = (i, j, s_i, e, g)$, where $i, j \in \modelset$, $s_i \in \{0, 0.5, 1\}$ is model~$i$'s episode score against model~$j$ (loss, draw, or win), and $s_j = 1 - s_i$; self-play episodes ($i = j$) are excluded. Across the three pipeline stages: (1)~the production within-game scheduler (\Cref{app:scheduler-team-mirror}) iterates over agent slots, expanding a homogeneous $2\times 2$ team episode into $k^2 = 4$ cross-team records (in simulated scheduler runs on the fitted Word Association and Coin Game ratings, one record per episode lowers savings from $79.6\%$ to $69.2\%$ and from $75.6\%$ to $62.8\%$ and lowers the false-sign rate among pairs with $q_{ij}\ge 0.95$ from $1.68\%$ to $0.26\%$ and from $0.43\%$ to $0.03\%$; \Cref{app:rq3-robustness}); (2)~ULB point estimation uses $1\times$ collapsed episode record (and $w_g = 1/N_g$ makes point ratings invariant to any constant per-game multiplicity); and (3)~ULB bootstrap intervals resample $1\times$ episodes (with mirror-seed cluster resampling audited in \Cref{tab:appI-boot}).
\subsection{Weighted Sufficient Statistics}
Let $N_g$ be the number of retained pairwise records for game~$g$.  Each
record receives weight $w_g = 1/N_g$, so every game contributes total
weight~one.  The weighted sufficient statistics are:
\begin{align*}
  W_i &\leftarrow W_i + w_g\, s_i, &
  W_j &\leftarrow W_j + w_g\,(1 - s_i),\\
  N_{ij} &\leftarrow N_{ij} + w_g, &
  N_{ji} &\leftarrow N_{ji} + w_g,
\end{align*}
yielding weighted log-likelihood
\[
  \ell(\gamma) = \sum_i W_i \log\gamma_i
  - \sum_{i < j} N_{ij}\log(\gamma_i + \gamma_j).
\]
\subsection{MM Algorithm and Convergence}
Strengths are estimated via a minorization--maximization (MM)
algorithm.\footnote{The implementation follows the Bradley--Terry MM algorithm
of \citet{hunter2004}.}  All $\gamma_i$ are initialized to~$1$ and renormalized to
sum to one after each update.  Iteration stops when
$\lVert\gamma^{(t+1)} - \gamma^{(t)}\rVert_2 < 10^{-6}$ or after 2\,000
iterations.
A smoothing factor $\varepsilon = 10^{-8}$ seeds the statistics: $W_i$ is
initialized to $\varepsilon(K{-}1)$ and each $N_{ij}$ to $2\varepsilon$, where
$K = |\modelset|$.  This prevents zero denominators in sparse comparison graphs
without materially affecting fitted strengths.
\subsection{Elo Conversion}
Fitted strengths are converted to the Elo
scale\footnote{See \citet{elo1978rating}.} via
$R_i^{\mathrm{raw}} = 400\log_{10}\gamma_i$ and shifted so the weakest model
anchors at zero:
\[
  R_i = R_i^{\mathrm{raw}} - \min_k R_k^{\mathrm{raw}}.
\]
Pairwise win probabilities depend only on Elo differences:
$P(i \succ j) = 1/(1 + 10^{-(R_i - R_j)/400})$. (Per-game tables in \Cref{sec:per_game_bootstrap} use this $\min_k R_k^{\mathrm{raw}} = 0$ convention; across multi-wave refreshes, the 39-model Unified Leaderboard in \Cref{tab:ulb-results} shifts $R^{\mathrm{raw}}$ so the mean of the 15 unchanged $S_0$ submissions remains fixed at its initial $S_0$ value of $165.49$ Elo.)
\subsection{Bootstrap Confidence Intervals}
\label{sec:ulb-bootstrap}
The ULB reports asymmetric 95\% bootstrap confidence
intervals\footnote{For bootstrap inference, see \citet{efron1979} and
\citet{Wasserman10All}.} from $N = 500$ simulations with a
fixed random seed.  Records are grouped by \emph{ordered} pair $(i, j)$ within
each game, as $(i, j)$ and $(j, i)$ form distinct buckets because player position
can encode seat, color, or role.  Each simulation resamples with replacement
within these buckets, rebuilds the game-balanced statistics with the same
$1/N_g$ weights, and refits the MM model.
The Elo shift computed from the point estimate is applied to all bootstrap
samples (samples are \emph{not} individually re-anchored), preserving bootstrap
variability.  Non-converging samples are discarded.  The reported margins are:
\[
  \Delta^{-}_i = R_i - \hat R_i^{(0.025)}, \qquad
  \Delta^{+}_i = \hat R_i^{(0.975)} - R_i,
\]
where $\hat R_i^{(q)}$ is the $q$-th percentile of converged bootstrap Elo
values for model~$i$.  These bootstrap intervals are separate from the
Hessian-based pairwise confidence estimates used by the adaptive scheduler
(\S\ref{sec:scheduler}).
\subsection{Per-Game Leaderboards}
Each per-game leaderboard uses the same MM \bt{} fit with identical smoothing
and convergence criteria, but without game-balancing weights (i.e., all records
have weight~1).

\section{Benchmark Game Environment Descriptions}
\label{app:game-descriptions}
Below, we provide short descriptions of the ten benchmark games included in the evaluation suite (\Cref{tab:game-properties}).
\subsection{Bargaining}
Two players engage in bilateral negotiation over a shared pool of three item types (books, hats, and basketballs), where each agent holds private integer valuations that are hidden from the opponent. Players alternate making proposals specifying the allocation they wish to keep, with either player free to accept the other's latest offer on their turn. If agreement is reached before the turn limit, each player scores their utility based on items received; otherwise, both players receive zero payoff.
\subsection{Checkers}
Played on an $8\times 8$ draughts board, two players alternate moving pieces diagonally forward onto dark squares, capturing opposing pieces via compulsory jumping sequences. Pieces that advance to the opponent's back rank are crowned kings, gaining the ability to move and capture backward along diagonals. A player wins by capturing all opposing pieces or leaving the opponent with no legal moves remaining.
\subsection{Chess}
A classical two-player adversarial board game played on an $8\times 8$ grid where each player commands sixteen pieces possessing unique movement geometries and tactical roles. The objective is to deliver checkmate to the opponent's king while preventing one's own king from falling under an unavoidable attack. The environment tests multi-step tactical calculation, strategic positional evaluation, and long-horizon look-ahead under complete information.
\subsection{Coin Game}
A multi-agent coordination benchmark played by two teams of two players on independent, private grid worlds where agents move to collect colored coins. Each player possesses a preferred coin color, and the team scoring function rewards collecting either team member's preferred coins while heavily penalizing the collection of opposing coins. Because teammates are separate model instances operating without direct communication, agents must establish cooperative conventions purely through observed board state transitions.
\subsection{Dark Hex}
An imperfect-information variant of Hex played on a rhombus grid of hexagonal cells, where two players strive to construct an unbroken connected path of stones between their designated board sides. The board is subject to fog-of-war: opponent moves remain hidden until an agent attempts an illegal placement on an already occupied cell, revealing that cell's state. Success demands topological reasoning, Bayesian belief updating, and risk-sensitive path exploration under partial observability.
\subsection{Dots and Boxes}
A combinatorial game played on a lattice of dots where two players alternate drawing unoriented line segments between adjacent coordinates. Completing the fourth boundary edge of a $1\times 1$ unit square captures that box for points and immediately grants the active player a compulsory bonus turn. The game requires precise tactical counting, sacrifice strategies (such as the double-cross), and cycle control to capture maximal scoring chains.
\subsection{Four in a Row}
A two-player connection game inspired by Connect Four, played on a vertical $7\times 6$ grid where players alternate dropping colored tokens into one of seven columns. Tokens drop to the lowest unoccupied slot within the selected column under simulated gravity. The first player to align an unbroken sequence of four tokens horizontally, vertically, or diagonally wins the match.
\subsection{Lines of Action}
An abstract strategy game played on an $8\times 8$ board where each player begins with twelve pieces positioned along opposite perimeter edges. Pieces move along straight lines (orthogonal or diagonal) by an exact distance equal to the total piece count currently occupying the line of movement, jumping over friendly pieces and capturing enemy pieces. The objective is to be the first player to unite all of one's remaining pieces into a single connected group.
\subsection{Poker (Repeated Heads-Up No-Limit Texas Hold'em)}
A two-player imperfect-information card game played across repeated hands where each player receives two private hole cards and shares five public community cards dealt across three betting streets. Agents select among discrete or continuous actions (fold, check, call, bet, or raise up to their full stack) to maximize long-term chip expected value. Strong play requires joint mastery of hidden-information state inference, pot-odds evaluation, bluffing, and opponent modeling across repeated encounters.
\subsection{Word Association}
A cooperative, team-based semantic association benchmark inspired by Codenames, played by two competing pairs of Cluemasters and Guessers on a $5\times 5$ grid of words. The Cluemaster emits a single clue word paired with an integer count to guide their teammate toward uncovering team-assigned words while avoiding neutral cards, opponent words, and the instant-loss assassin card. The Guesser must infer the semantic and pragmatic intent behind the clue to reveal target words under strict communication constraints.

\section{Unified Leaderboard Bootstrap Distributions}
\label{app:ulb-bootstrap-dist}

\begin{figure}[!htbp]
\centering
\includegraphics[width=0.88\linewidth,height=0.68\textheight,keepaspectratio]{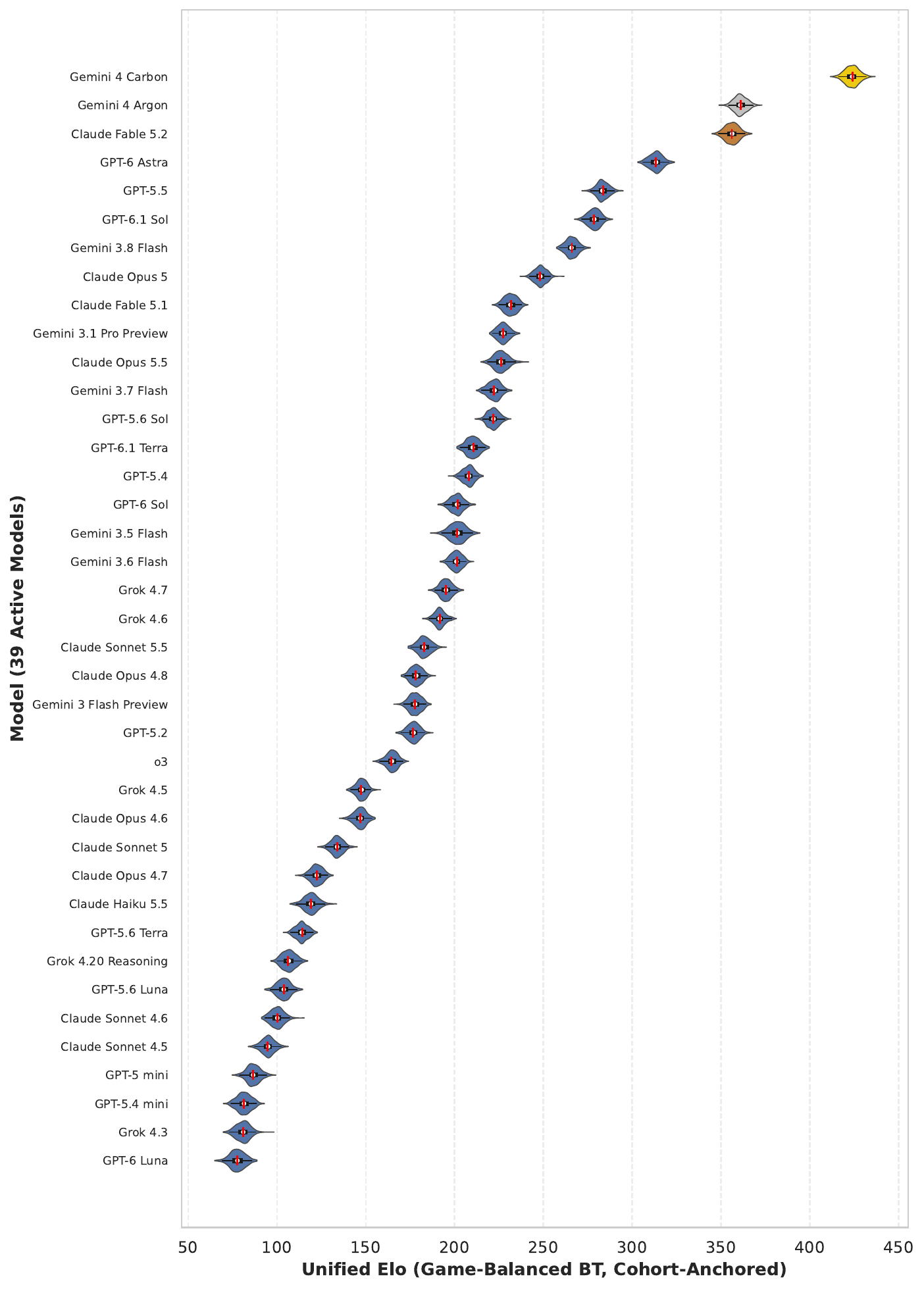}
\caption{Bootstrap Elo distribution across all 39 active models (500 resamples, anchored to the frozen $S_0$ untouched-cohort mean). Each violin shows the density of resampled Elo values; the white dot marks the median, the thick bar spans the interquartile range, and the thin whiskers extend to the 95\% confidence interval. The red marker indicates the point estimate.}
\label{fig:ulb-bootstrap}
\end{figure}

Figure~\ref{fig:ulb-bootstrap} shows the full bootstrap Elo distribution for each of the 39 models (mean 95\% CI width 13.8 Elo; \Cref{tab:appI-boot}).
There is an important distinction between \emph{parameter variance} (the width of the marginal Elo confidence interval) and \emph{rank distinguishability}.
In a Bradley--Terry model, Fisher information per pair is $I_{ij} = N_{ij} p_{ij}(1 - p_{ij})$, so a model's interval narrows with the number of closely contested episodes it has played. The 15 models with the most episodes ($8{,}501$--$10{,}774$ in \Cref{tab:ulb-results}) are all ranked between 7th and 33rd and have interval widths of $11.2$--$15.0$ Elo (e.g., GPT-5.4 at $[201.4, 213.8]$), whereas the 11 models with the fewest episodes ($3{,}739$--$5{,}883$), which include the top three and the two lowest-ranked models, have interval widths of $13.6$--$17.4$ Elo (e.g., $[416.2, 431.6]$ for Gemini 4 Carbon and $[69.4, 86.0]$ for GPT-6 Luna).
However, because adjacent mid-pack models are separated by point differences smaller than their individual interval widths, their ranks remain indistinguishable, creating overlap in the mid-pack violin densities despite low parameter variance.

\FloatBarrier

\section{Per-Game Bradley--Terry Elo Ratings Across Ten Game Arena Benchmarks}
\label{sec:per_game_bootstrap}

Tables~\ref{tab:elo-chess}--\ref{tab:elo-poker-heads-up} report per-game Bradley--Terry Elo ratings with 95\% bootstrap confidence intervals ($500$ resamples), mean tokens per turn (Tok/T), and mean execution cost per turn (\textcent/T) across all ten Game Arena benchmarks. Within each game, active models are ranked by point Elo with the lowest-rated model anchored at $0$.

\begin{table}[!htbp]
\captionsetup{font=footnotesize,skip=3pt}
\centering
\begin{minipage}[t]{0.486\textwidth}
\vspace{0pt}
\centering
\caption{Chess ($39$ models): Elo, 95\% CI, tokens, and cost.}
\label{tab:elo-chess}
\fontsize{6.6pt}{9.0pt}\selectfont
\setlength{\tabcolsep}{2.1pt}
\begin{tabular}{@{}lrrrrr@{}}
\toprule
Model & Elo & $\text{CI}_{\text{lo}}$ & $\text{CI}_{\text{hi}}$ & Tok/T & \textcent/T \\
\midrule
Gemini 4 Carbon & 1647.7 & 1608.2 & 1687.0 & 16{,}255 & 15.96 \\
Gemini 3.8 Flash & 1459.4 & 1421.6 & 1495.9 & 11{,}455 & 4.19 \\
GPT-6 Astra & 1436.6 & 1400.4 & 1466.8 & 6{,}684 & 30.67 \\
Gemini 3.7 Flash & 1398.6 & 1351.9 & 1437.7 & 3{,}436 & 1.19 \\
Gemini 4 Argon & 1372.2 & 1333.7 & 1406.1 & 2{,}993 & 2.70 \\
Gemini 3.5 Flash & 1343.0 & 1296.5 & 1380.0 & 4{,}032 & 3.37 \\
Gemini 3.1 Pro Preview & 1314.8 & 1276.2 & 1346.6 & 9{,}176 & 10.67 \\
Claude Fable 5.2 & 1314.7 & 1275.1 & 1351.7 & 2{,}231 & 7.30 \\
GPT-5.5 & 1245.4 & 1206.4 & 1280.7 & 13{,}786 & 40.47 \\
GPT-6.1 Sol & 1234.6 & 1193.0 & 1268.3 & 5{,}494 & 4.94 \\
Gemini 3.6 Flash & 1225.9 & 1182.5 & 1266.9 & 5{,}821 & 2.08 \\
Gemini 3 Flash Preview & 1214.8 & 1171.7 & 1250.6 & 8{,}727 & 2.53 \\
GPT-5.6 Sol & 1134.2 & 1086.9 & 1173.1 & 6{,}128 & 13.75 \\
GPT-6.1 Terra & 1070.5 & 1025.4 & 1111.0 & 3{,}744 & 1.59 \\
GPT-6 Sol & 1010.2 & 960.5 & 1054.4 & 4{,}978 & 4.40 \\
Claude Opus 5.5 & 976.2 & 925.6 & 1019.7 & 1{,}531 & 2.28 \\
o3 & 931.1 & 880.3 & 970.4 & 19{,}572 & 15.44 \\
GPT-5.4 & 893.4 & 848.7 & 928.5 & 10{,}182 & 14.83 \\
Grok 4.7 & 844.1 & 792.0 & 882.4 & 29{,}989 & 17.19 \\
GPT-5.2 & 830.9 & 784.0 & 865.3 & 11{,}624 & 15.83 \\
Grok 4.6 & 809.5 & 760.5 & 852.1 & 14{,}534 & 8.48 \\
Claude Sonnet 5.5 & 784.7 & 735.3 & 824.8 & 1{,}878 & 1.49 \\
Claude Fable 5.1 & 763.0 & 714.8 & 801.1 & 3{,}071 & 13.48 \\
GPT-5.6 Terra & 726.5 & 674.3 & 764.6 & 3{,}742 & 4.45 \\
GPT-5.6 Luna & 701.2 & 653.9 & 739.1 & 7{,}100 & 2.92 \\
Claude Opus 5 & 692.6 & 641.2 & 726.9 & 6{,}409 & 15.09 \\
Grok 4.20 Reasoning & 662.9 & 608.8 & 706.3 & 9{,}265 & 3.22 \\
Grok 4.5 & 645.5 & 593.0 & 685.5 & 14{,}974 & 8.76 \\
Claude Opus 4.8 & 621.6 & 573.7 & 663.4 & 7{,}384 & 17.56 \\
Grok 4.3 & 595.8 & 546.3 & 638.2 & 11{,}233 & 2.72 \\
Claude Haiku 5.5 & 594.8 & 539.9 & 638.9 & 4{,}585 & 0.21 \\
Claude Sonnet 5 & 588.2 & 531.9 & 630.3 & 10{,}448 & 10.30 \\
GPT-5 mini & 542.2 & 488.9 & 582.6 & 13{,}909 & 2.72 \\
GPT-6 Luna & 450.2 & 388.0 & 496.1 & 5{,}153 & 0.23 \\
GPT-5.4 mini & 435.2 & 375.9 & 483.9 & 15{,}946 & 7.05 \\
Claude Opus 4.7 & 284.1 & 201.1 & 350.8 & 7{,}212 & 17.21 \\
Claude Sonnet 4.6 & 97.8 & -7.3 & 162.9 & 10{,}049 & 14.55 \\
Claude Opus 4.6 & 50.3 & -40.9 & 113.8 & 6{,}062 & 14.35 \\
Claude Sonnet 4.5 & 0.0 & -110.8 & 77.7 & 7{,}244 & 10.34 \\
\bottomrule
\end{tabular}
\end{minipage}\hfill
\begin{minipage}[t]{0.486\textwidth}
\vspace{0pt}
\centering
\caption{Checkers ($39$ models): Elo, 95\% CI, tokens, and cost.}
\label{tab:elo-checkers}
\fontsize{6.6pt}{9.0pt}\selectfont
\setlength{\tabcolsep}{2.1pt}
\begin{tabular}{@{}lrrrrr@{}}
\toprule
Model & Elo & $\text{CI}_{\text{lo}}$ & $\text{CI}_{\text{hi}}$ & Tok/T & \textcent/T \\
\midrule
Gemini 4 Carbon & 1375.0 & 1342.3 & 1406.1 & 19{,}233 & 18.39 \\
GPT-6 Astra & 1328.0 & 1295.2 & 1360.9 & 2{,}974 & 9.54 \\
Claude Fable 5.2 & 1319.4 & 1293.2 & 1350.6 & 4{,}329 & 13.06 \\
Gemini 3.8 Flash & 1231.0 & 1201.3 & 1259.8 & 16{,}313 & 5.81 \\
GPT-6.1 Sol & 1198.8 & 1167.9 & 1229.4 & 2{,}778 & 1.71 \\
Claude Opus 5 & 1162.2 & 1136.9 & 1184.1 & 7{,}486 & 16.07 \\
GPT-5.5 & 1119.0 & 1092.1 & 1141.6 & 6{,}403 & 16.76 \\
Claude Opus 5.5 & 1109.0 & 1082.0 & 1132.2 & 2{,}729 & 3.33 \\
GPT-5.4 & 1088.0 & 1060.9 & 1112.6 & 7{,}965 & 10.72 \\
Claude Fable 5.1 & 1079.3 & 1051.4 & 1103.3 & 4{,}085 & 15.12 \\
Gemini 4 Argon & 1075.8 & 1047.2 & 1101.3 & 4{,}950 & 4.11 \\
Claude Opus 4.8 & 1074.6 & 1046.4 & 1096.8 & 8{,}416 & 18.38 \\
Gemini 3.7 Flash & 1067.8 & 1033.3 & 1097.8 & 7{,}383 & 2.46 \\
GPT-6.1 Terra & 1017.5 & 991.8 & 1042.1 & 2{,}906 & 0.91 \\
o3 & 986.6 & 957.1 & 1010.9 & 12{,}026 & 9.02 \\
Gemini 3.1 Pro Preview & 974.2 & 949.1 & 996.5 & 9{,}252 & 10.08 \\
Claude Sonnet 5.5 & 972.9 & 948.7 & 994.9 & 2{,}555 & 1.49 \\
GPT-5.2 & 965.2 & 938.0 & 986.4 & 5{,}386 & 6.33 \\
Claude Sonnet 5 & 946.8 & 919.0 & 970.9 & 9{,}375 & 8.80 \\
Gemini 3.6 Flash & 946.7 & 918.5 & 970.0 & 10{,}826 & 3.75 \\
Claude Opus 4.7 & 946.1 & 920.1 & 970.5 & 9{,}922 & 22.14 \\
GPT-6 Sol & 933.1 & 906.2 & 958.6 & 3{,}958 & 2.89 \\
Gemini 3.5 Flash & 918.5 & 887.5 & 945.8 & 9{,}903 & 8.15 \\
Gemini 3 Flash Preview & 889.4 & 862.1 & 914.2 & 19{,}258 & 5.52 \\
Claude Opus 4.6 & 885.4 & 856.3 & 912.7 & 7{,}906 & 17.57 \\
Grok 4.7 & 880.8 & 850.3 & 910.0 & 9{,}484 & 4.63 \\
GPT-5.6 Sol & 877.8 & 848.0 & 903.3 & 3{,}383 & 5.61 \\
GPT-5.6 Terra & 739.5 & 702.8 & 772.6 & 2{,}302 & 1.65 \\
Claude Sonnet 4.6 & 683.1 & 637.7 & 719.9 & 8{,}043 & 10.76 \\
GPT-5 mini & 674.5 & 629.4 & 711.1 & 6{,}350 & 1.10 \\
Claude Haiku 5.5 & 636.2 & 584.1 & 679.1 & 4{,}051 & 0.15 \\
GPT-6 Luna & 633.0 & 589.5 & 671.4 & 2{,}597 & 0.08 \\
Claude Sonnet 4.5 & 561.2 & 507.1 & 600.9 & 9{,}188 & 12.47 \\
Grok 4.6 & 547.6 & 490.7 & 589.7 & 5{,}136 & 2.56 \\
Grok 4.20 Reasoning & 418.2 & 352.9 & 470.8 & 4{,}183 & 1.24 \\
GPT-5.6 Luna & 323.1 & 244.2 & 382.9 & 2{,}086 & 0.46 \\
GPT-5.4 mini & 283.6 & 211.5 & 350.8 & 4{,}094 & 1.49 \\
Grok 4.5 & 83.2 & -19.4 & 161.6 & 2{,}653 & 1.12 \\
Grok 4.3 & 0.0 & -149.0 & 99.0 & 2{,}800 & 0.51 \\
\bottomrule
\end{tabular}
\end{minipage}
\end{table}

\begin{table}[!htbp]
\captionsetup{font=footnotesize,skip=3pt}
\centering
\begin{minipage}[t]{0.486\textwidth}
\vspace{0pt}
\centering
\caption{Four in a Row ($39$ models): Elo, 95\% CI, tokens, and cost.}
\label{tab:elo-four-in-a-row}
\fontsize{6.6pt}{9.0pt}\selectfont
\setlength{\tabcolsep}{2.1pt}
\begin{tabular}{@{}lrrrrr@{}}
\toprule
Model & Elo & $\text{CI}_{\text{lo}}$ & $\text{CI}_{\text{hi}}$ & Tok/T & \textcent/T \\
\midrule
Gemini 4 Carbon & 994.5 & 965.0 & 1024.8 & 28{,}764 & 28.46 \\
Gemini 4 Argon & 907.8 & 880.4 & 942.6 & 9{,}078 & 8.81 \\
Claude Fable 5.2 & 810.8 & 771.3 & 847.5 & 7{,}727 & 29.53 \\
GPT-6 Astra & 693.0 & 652.2 & 722.1 & 6{,}470 & 29.82 \\
Gemini 3.8 Flash & 690.2 & 657.2 & 723.0 & 21{,}655 & 8.03 \\
Claude Opus 5 & 677.9 & 645.5 & 705.1 & 17{,}263 & 42.29 \\
Claude Fable 5.1 & 599.5 & 569.8 & 626.3 & 8{,}559 & 41.07 \\
Gemini 3.1 Pro Preview & 596.6 & 565.9 & 622.3 & 18{,}767 & 22.22 \\
GPT-6.1 Sol & 575.4 & 543.8 & 604.7 & 5{,}233 & 4.73 \\
Gemini 3.5 Flash & 550.8 & 518.7 & 581.1 & 15{,}016 & 13.29 \\
Gemini 3.7 Flash & 544.0 & 512.3 & 574.0 & 9{,}111 & 3.33 \\
Claude Opus 5.5 & 533.3 & 500.3 & 563.8 & 3{,}800 & 6.91 \\
Gemini 3.6 Flash & 495.4 & 465.7 & 521.4 & 12{,}946 & 4.76 \\
Grok 4.6 & 493.6 & 464.0 & 517.0 & 26{,}865 & 15.88 \\
Grok 4.7 & 465.8 & 436.8 & 490.8 & 56{,}796 & 33.21 \\
GPT-5.5 & 457.1 & 428.9 & 484.4 & 12{,}133 & 35.67 \\
Claude Opus 4.8 & 427.9 & 394.5 & 451.7 & 13{,}061 & 31.78 \\
GPT-5.4 & 423.7 & 392.6 & 448.9 & 14{,}402 & 21.24 \\
Claude Haiku 5.5 & 410.5 & 379.0 & 439.0 & 5{,}292 & 0.25 \\
Claude Sonnet 5.5 & 403.4 & 367.9 & 431.9 & 2{,}999 & 2.66 \\
GPT-5.2 & 388.1 & 355.7 & 412.7 & 14{,}029 & 19.29 \\
Grok 4.5 & 360.3 & 329.3 & 388.6 & 25{,}435 & 15.04 \\
GPT-5.6 Sol & 344.8 & 312.4 & 371.3 & 5{,}856 & 12.93 \\
GPT-6 Sol & 332.4 & 300.1 & 358.4 & 4{,}019 & 3.51 \\
GPT-6.1 Terra & 325.3 & 292.2 & 353.1 & 3{,}344 & 1.42 \\
Claude Sonnet 5 & 314.2 & 282.6 & 340.2 & 12{,}432 & 13.00 \\
o3 & 291.8 & 256.9 & 319.5 & 17{,}892 & 14.15 \\
Grok 4.20 Reasoning & 278.2 & 244.7 & 309.8 & 11{,}749 & 4.65 \\
Gemini 3 Flash Preview & 252.2 & 219.4 & 280.1 & 22{,}450 & 6.66 \\
Claude Opus 4.6 & 203.6 & 167.2 & 234.4 & 10{,}508 & 25.56 \\
GPT-5.4 mini & 197.8 & 164.9 & 225.4 & 15{,}610 & 6.92 \\
Claude Opus 4.7 & 189.7 & 152.0 & 218.1 & 15{,}083 & 36.80 \\
Grok 4.3 & 188.4 & 152.2 & 222.1 & 13{,}616 & 3.33 \\
GPT-5.6 Luna & 180.7 & 143.8 & 212.6 & 5{,}232 & 1.74 \\
Claude Sonnet 4.5 & 150.0 & 114.2 & 180.6 & 10{,}833 & 15.85 \\
Claude Sonnet 4.6 & 126.7 & 92.5 & 156.8 & 13{,}063 & 19.17 \\
GPT-6 Luna & 120.9 & 81.5 & 152.9 & 3{,}544 & 0.15 \\
GPT-5.6 Terra & 86.0 & 45.7 & 118.9 & 2{,}850 & 3.22 \\
GPT-5 mini & 0.0 & -52.6 & 45.8 & 6{,}957 & 1.34 \\
\bottomrule
\end{tabular}
\end{minipage}\hfill
\begin{minipage}[t]{0.486\textwidth}
\vspace{0pt}
\centering
\caption{Dots and Boxes ($39$ models): Elo, 95\% CI, tokens, and cost.}
\label{tab:elo-dots-and-boxes}
\fontsize{6.6pt}{9.0pt}\selectfont
\setlength{\tabcolsep}{2.1pt}
\begin{tabular}{@{}lrrrrr@{}}
\toprule
Model & Elo & $\text{CI}_{\text{lo}}$ & $\text{CI}_{\text{hi}}$ & Tok/T & \textcent/T \\
\midrule
Claude Fable 5.2 & 906.6 & 869.5 & 947.1 & 7{,}385 & 26.59 \\
GPT-6 Astra & 716.0 & 673.7 & 756.6 & 3{,}157 & 11.71 \\
Claude Opus 5.5 & 667.9 & 631.5 & 700.8 & 3{,}840 & 6.23 \\
GPT-6.1 Sol & 633.0 & 597.0 & 664.1 & 2{,}953 & 2.14 \\
Gemini 4 Argon & 612.8 & 579.3 & 641.7 & 13{,}880 & 13.29 \\
Claude Fable 5.1 & 609.1 & 572.8 & 643.4 & 7{,}444 & 33.59 \\
Claude Opus 5 & 579.2 & 549.0 & 606.8 & 14{,}502 & 34.45 \\
Claude Sonnet 5.5 & 568.6 & 534.2 & 599.7 & 3{,}731 & 3.01 \\
Gemini 3.8 Flash & 531.5 & 499.0 & 559.3 & 19{,}745 & 7.20 \\
Claude Opus 4.8 & 518.4 & 483.0 & 544.1 & 12{,}038 & 28.29 \\
GPT-6.1 Terra & 496.5 & 464.2 & 524.7 & 2{,}182 & 0.69 \\
GPT-5.5 & 487.8 & 455.7 & 513.0 & 6{,}893 & 18.99 \\
Gemini 3.1 Pro Preview & 479.7 & 450.3 & 506.6 & 9{,}414 & 10.61 \\
Claude Haiku 5.5 & 471.4 & 437.0 & 498.5 & 11{,}009 & 0.51 \\
GPT-5.4 & 455.4 & 427.3 & 478.4 & 11{,}929 & 17.05 \\
Grok 4.6 & 448.2 & 419.4 & 475.0 & 13{,}130 & 7.49 \\
Claude Sonnet 5 & 444.8 & 414.0 & 471.4 & 15{,}609 & 15.69 \\
GPT-5.6 Sol & 436.3 & 405.4 & 464.5 & 4{,}411 & 8.85 \\
Grok 4.7 & 430.2 & 398.7 & 455.4 & 17{,}028 & 9.30 \\
Gemini 3.7 Flash & 414.1 & 382.0 & 444.0 & 7{,}536 & 2.62 \\
Gemini 3.5 Flash & 410.5 & 373.6 & 440.4 & 10{,}940 & 9.33 \\
GPT-5.2 & 391.2 & 360.9 & 416.2 & 9{,}552 & 12.55 \\
Claude Opus 4.6 & 378.6 & 346.7 & 403.1 & 12{,}084 & 28.69 \\
Grok 4.5 & 370.7 & 341.5 & 399.8 & 8{,}703 & 4.85 \\
GPT-6 Sol & 337.3 & 300.4 & 372.3 & 3{,}120 & 2.31 \\
Claude Sonnet 4.6 & 314.4 & 273.9 & 349.6 & 13{,}971 & 20.05 \\
Gemini 3.6 Flash & 302.2 & 266.0 & 331.3 & 9{,}357 & 3.30 \\
Gemini 3 Flash Preview & 268.0 & 229.2 & 294.9 & 20{,}616 & 6.01 \\
GPT-5 mini & 221.3 & 186.9 & 249.9 & 11{,}705 & 2.22 \\
Grok 4.20 Reasoning & 142.4 & 100.1 & 179.1 & 9{,}510 & 3.74 \\
GPT-5.4 mini & 120.2 & 79.2 & 147.8 & 11{,}601 & 4.97 \\
GPT-5.6 Terra & 116.0 & 73.1 & 150.6 & 3{,}492 & 3.65 \\
o3 & 111.9 & 68.2 & 146.3 & 13{,}749 & 10.60 \\
GPT-6 Luna & 95.8 & 53.3 & 132.3 & 3{,}935 & 0.16 \\
Claude Opus 4.7 & 83.6 & 34.3 & 122.6 & 14{,}641 & 34.80 \\
Grok 4.3 & 72.0 & 27.1 & 104.3 & 8{,}029 & 1.87 \\
Claude Sonnet 4.5 & 57.7 & 12.9 & 96.3 & 10{,}398 & 14.70 \\
GPT-5.6 Luna & 15.0 & -24.7 & 48.3 & 4{,}636 & 1.34 \\
Gemini 4 Carbon & 0.0 & -64.3 & 46.7 & 46{,}326 & 45.73 \\
\bottomrule
\end{tabular}
\end{minipage}
\end{table}

\begin{table}[!htbp]
\captionsetup{font=footnotesize,skip=3pt}
\centering
\begin{minipage}[t]{0.486\textwidth}
\vspace{0pt}
\centering
\caption{Bargaining ($39$ models): Elo, 95\% CI, tokens, and cost.}
\label{tab:elo-bargaining}
\fontsize{6.6pt}{9.0pt}\selectfont
\setlength{\tabcolsep}{2.1pt}
\begin{tabular}{@{}lrrrrr@{}}
\toprule
Model & Elo & $\text{CI}_{\text{lo}}$ & $\text{CI}_{\text{hi}}$ & Tok/T & \textcent/T \\
\midrule
Gemini 4 Carbon & 316.9 & 302.7 & 330.6 & 32{,}190 & 31.50 \\
GPT-5.5 & 274.2 & 265.9 & 281.5 & 8{,}827 & 24.48 \\
Claude Fable 5.2 & 252.3 & 242.8 & 263.0 & 2{,}432 & 5.64 \\
GPT-6 Astra & 245.5 & 235.1 & 255.9 & 2{,}213 & 6.37 \\
Gemini 4 Argon & 243.5 & 234.3 & 254.2 & 5{,}895 & 5.21 \\
Claude Sonnet 5.5 & 242.9 & 234.0 & 251.7 & 2{,}237 & 1.25 \\
GPT-6.1 Sol & 239.1 & 229.8 & 248.0 & 2{,}191 & 1.25 \\
Claude Opus 5.5 & 234.2 & 224.0 & 245.0 & 2{,}444 & 2.91 \\
Claude Fable 5.1 & 223.9 & 214.3 & 233.3 & 2{,}601 & 8.09 \\
Grok 4.20 Reasoning & 220.0 & 211.3 & 229.2 & 13{,}150 & 3.08 \\
Grok 4.5 & 218.0 & 209.0 & 227.8 & 20{,}903 & 12.11 \\
GPT-5.6 Sol & 217.7 & 208.0 & 226.0 & 3{,}466 & 5.75 \\
Gemini 3.7 Flash & 214.4 & 203.7 & 224.5 & 5{,}980 & 1.99 \\
GPT-6.1 Terra & 214.0 & 202.1 & 225.0 & 2{,}600 & 0.83 \\
Grok 4.6 & 213.5 & 203.9 & 222.3 & 22{,}864 & 13.27 \\
Gemini 3.8 Flash & 207.8 & 198.2 & 217.7 & 22{,}202 & 8.08 \\
GPT-6 Sol & 205.6 & 197.0 & 214.0 & 2{,}584 & 1.65 \\
Gemini 3.1 Pro Preview & 188.4 & 178.6 & 198.5 & 6{,}999 & 7.58 \\
Claude Opus 4.7 & 188.3 & 178.1 & 197.7 & 4{,}545 & 8.99 \\
o3 & 186.8 & 175.6 & 197.7 & 12{,}139 & 9.23 \\
GPT-5.4 & 185.6 & 175.3 & 196.0 & 8{,}473 & 11.72 \\
Gemini 3 Flash Preview & 180.3 & 169.3 & 190.8 & 19{,}336 & 5.59 \\
Gemini 3.6 Flash & 172.8 & 162.0 & 184.1 & 8{,}256 & 2.85 \\
Claude Opus 4.6 & 172.1 & 160.8 & 183.5 & 5{,}908 & 13.02 \\
Claude Opus 4.8 & 171.9 & 161.6 & 182.3 & 3{,}958 & 7.50 \\
Grok 4.7 & 163.1 & 149.5 & 174.8 & 66{,}254 & 38.74 \\
Claude Haiku 5.5 & 162.6 & 149.3 & 174.7 & 4{,}593 & 0.18 \\
Claude Sonnet 4.5 & 161.3 & 148.8 & 173.1 & 8{,}872 & 12.25 \\
GPT-5.6 Luna & 156.1 & 143.6 & 168.2 & 5{,}463 & 1.71 \\
Claude Opus 5 & 152.3 & 140.9 & 163.8 & 2{,}302 & 3.38 \\
GPT-5 mini & 142.5 & 128.7 & 157.6 & 11{,}363 & 2.14 \\
GPT-6 Luna & 141.9 & 129.1 & 154.9 & 3{,}572 & 0.13 \\
Gemini 3.5 Flash & 141.7 & 127.3 & 155.8 & 9{,}802 & 8.21 \\
GPT-5.6 Terra & 131.5 & 116.4 & 147.4 & 2{,}560 & 2.21 \\
GPT-5.2 & 126.2 & 111.3 & 141.9 & 5{,}262 & 6.42 \\
Grok 4.3 & 124.6 & 108.1 & 139.1 & 12{,}586 & 3.01 \\
Claude Sonnet 5 & 81.9 & 56.5 & 105.0 & 3{,}411 & 2.54 \\
Claude Sonnet 4.6 & 74.9 & 49.3 & 96.9 & 5{,}277 & 6.89 \\
GPT-5.4 mini & 0.0 & -38.2 & 38.9 & 11{,}497 & 4.90 \\
\bottomrule
\end{tabular}
\end{minipage}\hfill
\begin{minipage}[t]{0.486\textwidth}
\vspace{0pt}
\centering
\caption{Lines of Action ($39$ models): Elo, 95\% CI, tokens, and cost.}
\label{tab:elo-lines-of-action}
\fontsize{6.6pt}{9.0pt}\selectfont
\setlength{\tabcolsep}{2.1pt}
\begin{tabular}{@{}lrrrrr@{}}
\toprule
Model & Elo & $\text{CI}_{\text{lo}}$ & $\text{CI}_{\text{hi}}$ & Tok/T & \textcent/T \\
\midrule
Gemini 4 Carbon & 1273.8 & 1232.8 & 1313.4 & 34{,}545 & 33.65 \\
Claude Fable 5.2 & 1116.9 & 1062.5 & 1168.2 & 11{,}826 & 42.85 \\
Gemini 4 Argon & 1066.7 & 1010.8 & 1110.0 & 14{,}475 & 13.59 \\
GPT-6 Astra & 1014.2 & 956.5 & 1064.2 & 12{,}849 & 58.58 \\
GPT-6.1 Sol & 865.7 & 792.4 & 916.9 & 11{,}520 & 10.39 \\
Claude Opus 5 & 821.5 & 756.5 & 873.0 & 23{,}050 & 54.90 \\
Claude Fable 5.1 & 654.6 & 581.2 & 713.0 & 9{,}731 & 43.12 \\
Claude Opus 5.5 & 607.5 & 540.0 & 660.9 & 4{,}996 & 7.82 \\
GPT-5.5 & 600.2 & 534.1 & 647.8 & 22{,}620 & 65.18 \\
Grok 4.7 & 570.3 & 507.6 & 621.3 & 46{,}246 & 26.66 \\
GPT-5.6 Sol & 522.1 & 457.6 & 571.5 & 12{,}843 & 28.26 \\
Grok 4.6 & 521.2 & 457.7 & 571.2 & 43{,}621 & 25.59 \\
GPT-5.4 & 488.8 & 428.4 & 537.6 & 27{,}434 & 39.83 \\
Grok 4.5 & 469.3 & 407.7 & 514.0 & 37{,}898 & 22.18 \\
Gemini 3.1 Pro Preview & 443.1 & 373.3 & 490.2 & 23{,}457 & 27.06 \\
GPT-6 Sol & 435.0 & 372.6 & 485.1 & 11{,}108 & 9.97 \\
Gemini 3.8 Flash & 431.1 & 363.6 & 479.6 & 28{,}058 & 10.19 \\
GPT-5.2 & 426.7 & 364.6 & 473.7 & 23{,}923 & 32.20 \\
Claude Opus 4.8 & 404.8 & 341.4 & 448.9 & 16{,}612 & 38.81 \\
GPT-6.1 Terra & 393.4 & 331.5 & 439.4 & 7{,}048 & 2.94 \\
Claude Opus 4.6 & 380.0 & 318.1 & 431.1 & 19{,}121 & 45.55 \\
Gemini 3.7 Flash & 351.2 & 285.3 & 400.9 & 14{,}943 & 5.28 \\
Gemini 3.6 Flash & 350.3 & 283.8 & 393.8 & 17{,}408 & 6.21 \\
Claude Sonnet 5.5 & 337.5 & 270.9 & 387.4 & 3{,}563 & 2.47 \\
Gemini 3.5 Flash & 336.2 & 270.3 & 384.2 & 21{,}511 & 18.55 \\
GPT-5.6 Luna & 320.9 & 256.5 & 366.4 & 15{,}449 & 5.36 \\
GPT-5.6 Terra & 310.1 & 248.2 & 353.7 & 9{,}244 & 10.99 \\
o3 & 292.6 & 223.7 & 343.8 & 28{,}330 & 22.02 \\
Grok 4.20 Reasoning & 284.0 & 220.5 & 334.0 & 19{,}981 & 4.75 \\
GPT-5.4 mini & 274.4 & 207.5 & 319.0 & 34{,}588 & 15.17 \\
Claude Sonnet 4.5 & 226.0 & 158.5 & 274.3 & 17{,}822 & 25.34 \\
Claude Sonnet 5 & 221.8 & 159.3 & 272.8 & 29{,}248 & 29.05 \\
Claude Sonnet 4.6 & 209.1 & 147.9 & 255.9 & 17{,}922 & 25.51 \\
GPT-6 Luna & 196.5 & 132.3 & 248.8 & 9{,}153 & 0.40 \\
Grok 4.3 & 183.7 & 116.8 & 233.6 & 16{,}254 & 3.87 \\
Gemini 3 Flash Preview & 159.4 & 88.5 & 210.5 & 28{,}455 & 8.27 \\
GPT-5 mini & 137.5 & 70.1 & 192.3 & 23{,}934 & 4.60 \\
Claude Haiku 5.5 & 44.9 & -33.9 & 108.3 & 12{,}661 & 0.58 \\
Claude Opus 4.7 & 0.0 & -80.9 & 71.6 & 30{,}646 & 73.84 \\
\bottomrule
\end{tabular}
\end{minipage}
\end{table}

\begin{table}[!htbp]
\captionsetup{font=footnotesize,skip=3pt}
\centering
\begin{minipage}[t]{0.486\textwidth}
\vspace{0pt}
\centering
\caption{Dark Hex ($39$ models): Elo, 95\% CI, tokens, and cost.}
\label{tab:elo-dark-hex}
\fontsize{6.6pt}{9.0pt}\selectfont
\setlength{\tabcolsep}{2.1pt}
\begin{tabular}{@{}lrrrrr@{}}
\toprule
Model & Elo & $\text{CI}_{\text{lo}}$ & $\text{CI}_{\text{hi}}$ & Tok/T & \textcent/T \\
\midrule
Gemini 3.8 Flash & 543.7 & 509.2 & 580.3 & 16{,}214 & 5.88 \\
Gemini 4 Carbon & 515.0 & 481.4 & 552.1 & 26{,}669 & 26.09 \\
GPT-5.5 & 457.9 & 428.4 & 486.9 & 4{,}332 & 11.40 \\
Gemini 4 Argon & 439.9 & 408.3 & 472.3 & 4{,}872 & 4.32 \\
Gemini 3.5 Flash & 402.4 & 363.3 & 438.2 & 6{,}043 & 4.94 \\
GPT-5.4 & 350.3 & 321.1 & 377.3 & 6{,}305 & 8.66 \\
Gemini 3 Flash Preview & 323.9 & 296.5 & 349.0 & 13{,}743 & 3.96 \\
GPT-5.6 Sol & 316.3 & 288.3 & 337.2 & 2{,}223 & 3.38 \\
Claude Opus 5 & 314.5 & 287.9 & 338.2 & 4{,}384 & 9.04 \\
GPT-6.1 Terra & 310.7 & 280.3 & 342.1 & 1{,}588 & 0.40 \\
Gemini 3.6 Flash & 303.4 & 276.6 & 326.8 & 6{,}747 & 2.33 \\
Claude Fable 5.2 & 281.6 & 253.5 & 305.1 & 2{,}292 & 5.79 \\
Grok 4.6 & 257.4 & 233.8 & 278.0 & 12{,}355 & 7.02 \\
Gemini 3.1 Pro Preview & 256.7 & 233.6 & 277.1 & 5{,}583 & 6.04 \\
Gemini 3.7 Flash & 254.3 & 227.5 & 279.4 & 4{,}746 & 1.58 \\
Claude Opus 5.5 & 245.1 & 221.7 & 268.2 & 1{,}955 & 2.30 \\
Claude Sonnet 5.5 & 242.4 & 218.0 & 265.4 & 1{,}762 & 0.96 \\
GPT-6 Astra & 241.2 & 212.4 & 264.3 & 1{,}622 & 4.19 \\
GPT-6.1 Sol & 232.7 & 207.8 & 255.8 & 1{,}612 & 0.83 \\
Claude Opus 4.8 & 226.3 & 201.7 & 249.4 & 4{,}765 & 10.05 \\
Grok 4.7 & 224.0 & 196.7 & 245.8 & 14{,}703 & 7.90 \\
Claude Fable 5.1 & 213.4 & 189.0 & 236.7 & 2{,}379 & 7.88 \\
GPT-6 Sol & 211.9 & 186.0 & 235.8 & 1{,}832 & 1.05 \\
GPT-5.2 & 210.9 & 187.7 & 234.1 & 1{,}700 & 1.60 \\
Claude Haiku 5.5 & 202.4 & 175.6 & 229.4 & 3{,}398 & 0.13 \\
o3 & 182.4 & 155.9 & 208.7 & 4{,}692 & 3.37 \\
Claude Sonnet 5 & 167.7 & 140.5 & 190.5 & 6{,}659 & 6.35 \\
GPT-5 mini & 149.1 & 124.2 & 175.1 & 4{,}417 & 0.77 \\
Grok 4.5 & 134.0 & 109.6 & 156.1 & 4{,}385 & 2.26 \\
GPT-5.6 Luna & 104.4 & 79.8 & 127.4 & 1{,}614 & 0.33 \\
GPT-6 Luna & 97.7 & 70.0 & 123.5 & 1{,}558 & 0.04 \\
GPT-5.6 Terra & 90.1 & 61.8 & 114.9 & 1{,}262 & 0.61 \\
Claude Sonnet 4.6 & 75.2 & 47.3 & 97.6 & 3{,}460 & 4.34 \\
Claude Opus 4.6 & 68.9 & 37.4 & 97.7 & 6{,}119 & 13.87 \\
Claude Opus 4.7 & 53.7 & 22.3 & 80.1 & 4{,}524 & 9.44 \\
GPT-5.4 mini & 53.4 & 23.5 & 80.1 & 4{,}178 & 1.64 \\
Grok 4.3 & 27.7 & -3.4 & 56.3 & 5{,}046 & 1.14 \\
Grok 4.20 Reasoning & 0.5 & -34.5 & 33.5 & 4{,}428 & 1.38 \\
Claude Sonnet 4.5 & 0.0 & -38.3 & 38.5 & 4{,}929 & 6.54 \\
\bottomrule
\end{tabular}
\end{minipage}\hfill
\begin{minipage}[t]{0.486\textwidth}
\vspace{0pt}
\centering
\caption{Coin Game ($39$ models): Elo, 95\% CI, tokens, and cost.}
\label{tab:elo-coin-game}
\fontsize{6.6pt}{9.0pt}\selectfont
\setlength{\tabcolsep}{2.1pt}
\begin{tabular}{@{}lrrrrr@{}}
\toprule
Model & Elo & $\text{CI}_{\text{lo}}$ & $\text{CI}_{\text{hi}}$ & Tok/T & \textcent/T \\
\midrule
GPT-5.5 & 513.7 & 497.7 & 530.2 & 4{,}732 & 11.34 \\
Gemini 4 Carbon & 506.4 & 489.0 & 523.5 & 15{,}692 & 14.64 \\
Gemini 3.7 Flash & 484.5 & 468.6 & 501.2 & 5{,}866 & 1.81 \\
Gemini 3.8 Flash & 484.3 & 469.4 & 500.2 & 12{,}908 & 4.45 \\
Gemini 3 Flash Preview & 471.7 & 457.2 & 485.3 & 19{,}197 & 5.44 \\
Gemini 4 Argon & 457.6 & 443.2 & 472.3 & 5{,}249 & 4.20 \\
Gemini 3.6 Flash & 440.2 & 425.0 & 454.6 & 8{,}785 & 2.91 \\
GPT-6 Sol & 438.0 & 422.5 & 452.4 & 2{,}316 & 1.13 \\
GPT-6 Astra & 424.1 & 409.5 & 437.9 & 2{,}142 & 4.76 \\
Gemini 3.1 Pro Preview & 420.1 & 405.9 & 434.6 & 8{,}560 & 8.98 \\
GPT-6.1 Sol & 418.9 & 405.9 & 432.8 & 2{,}160 & 0.97 \\
GPT-5.6 Sol & 416.1 & 401.4 & 430.0 & 2{,}390 & 2.57 \\
GPT-5.4 & 416.0 & 401.2 & 431.4 & 6{,}962 & 9.01 \\
Claude Opus 5 & 405.7 & 389.1 & 422.3 & 4{,}696 & 8.49 \\
Claude Fable 5.2 & 394.8 & 380.0 & 411.0 & 2{,}694 & 5.60 \\
Gemini 3.5 Flash & 384.1 & 365.1 & 400.6 & 9{,}504 & 7.59 \\
Grok 4.7 & 382.4 & 366.7 & 397.0 & 27{,}351 & 15.28 \\
Grok 4.6 & 378.5 & 362.5 & 393.0 & 14{,}180 & 7.89 \\
Grok 4.5 & 362.8 & 347.4 & 376.8 & 12{,}190 & 6.71 \\
Claude Sonnet 4.5 & 358.3 & 338.0 & 376.2 & 8{,}846 & 11.65 \\
Claude Opus 4.8 & 348.8 & 332.5 & 364.6 & 6{,}739 & 13.61 \\
Claude Fable 5.1 & 329.8 & 311.2 & 349.7 & 2{,}754 & 7.18 \\
GPT-6.1 Terra & 329.7 & 308.0 & 347.8 & 2{,}538 & 0.67 \\
Claude Opus 4.6 & 315.5 & 296.1 & 334.5 & 5{,}512 & 11.08 \\
Claude Opus 4.7 & 312.8 & 296.2 & 331.1 & 5{,}657 & 10.90 \\
Claude Sonnet 4.6 & 292.5 & 274.1 & 309.6 & 5{,}772 & 7.04 \\
Grok 4.3 & 282.8 & 263.0 & 303.0 & 6{,}689 & 1.49 \\
Claude Opus 5.5 & 250.3 & 231.5 & 268.8 & 2{,}297 & 2.00 \\
Grok 4.20 Reasoning & 248.9 & 225.0 & 270.4 & 5{,}421 & 1.08 \\
GPT-5.2 & 246.4 & 224.7 & 268.2 & 2{,}794 & 2.50 \\
Claude Haiku 5.5 & 238.8 & 217.6 & 260.5 & 3{,}578 & 0.11 \\
Claude Sonnet 5 & 237.4 & 214.9 & 258.6 & 7{,}512 & 6.68 \\
GPT-5 mini & 227.8 & 199.5 & 250.6 & 7{,}530 & 1.30 \\
GPT-5.6 Terra & 206.4 & 180.5 & 229.1 & 2{,}001 & 0.95 \\
Claude Sonnet 5.5 & 199.8 & 179.9 & 219.0 & 2{,}234 & 0.94 \\
GPT-6 Luna & 177.3 & 153.7 & 200.4 & 2{,}306 & 0.06 \\
o3 & 154.2 & 127.1 & 180.8 & 5{,}339 & 3.58 \\
GPT-5.4 mini & 88.6 & 47.3 & 124.1 & 7{,}777 & 3.07 \\
GPT-5.6 Luna & 0.0 & -50.2 & 45.5 & 2{,}845 & 0.67 \\
\bottomrule
\end{tabular}
\end{minipage}
\end{table}

\begin{table}[!htbp]
\captionsetup{font=footnotesize,skip=3pt}
\centering
\begin{minipage}[t]{0.486\textwidth}
\vspace{0pt}
\centering
\caption{Word Association ($37$ models): Elo, 95\% CI, tokens, and cost.}
\label{tab:elo-word-association}
\fontsize{6.6pt}{9.0pt}\selectfont
\setlength{\tabcolsep}{2.1pt}
\begin{tabular}{@{}lrrrrr@{}}
\toprule
Model & Elo & $\text{CI}_{\text{lo}}$ & $\text{CI}_{\text{hi}}$ & Tok/T & \textcent/T \\
\midrule
Gemini 4 Carbon & 706.3 & 653.5 & 763.8 & 13{,}297 & 12.44 \\
Gemini 4 Argon & 596.7 & 553.9 & 641.5 & 7{,}371 & 6.51 \\
GPT-5.5 & 533.2 & 489.3 & 575.4 & 3{,}943 & 9.40 \\
GPT-6 Astra & 475.1 & 429.7 & 516.1 & 1{,}673 & 3.07 \\
Claude Fable 5.2 & 416.7 & 374.1 & 456.3 & 2{,}274 & 4.03 \\
GPT-6.1 Sol & 413.4 & 376.0 & 448.0 & 1{,}662 & 0.59 \\
o3 & 327.7 & 290.1 & 361.0 & 7{,}756 & 5.61 \\
GPT-5.6 Sol & 322.3 & 285.4 & 355.0 & 1{,}791 & 1.58 \\
Gemini 3.8 Flash & 316.9 & 281.5 & 346.4 & 8{,}464 & 2.84 \\
GPT-6 Sol & 312.9 & 277.8 & 340.7 & 1{,}826 & 0.75 \\
Gemini 3.7 Flash & 301.3 & 262.7 & 330.8 & 4{,}415 & 1.33 \\
Gemini 3.1 Pro Preview & 288.9 & 258.1 & 314.2 & 4{,}349 & 4.12 \\
GPT-6.1 Terra & 282.4 & 241.9 & 310.9 & 1{,}689 & 0.30 \\
Claude Opus 4.6 & 275.9 & 242.2 & 304.5 & 3{,}912 & 7.42 \\
Gemini 3.5 Flash & 267.4 & 227.5 & 297.2 & 4{,}229 & 2.98 \\
Claude Opus 5.5 & 262.2 & 227.1 & 293.0 & 2{,}168 & 1.75 \\
Grok 4.6 & 261.2 & 224.4 & 291.4 & 5{,}825 & 2.98 \\
Gemini 3.6 Flash & 260.3 & 224.9 & 290.7 & 4{,}816 & 1.48 \\
GPT-5.4 & 253.5 & 221.5 & 282.8 & 4{,}839 & 6.01 \\
Grok 4.7 & 251.8 & 215.5 & 277.7 & 7{,}101 & 3.23 \\
Gemini 3 Flash Preview & 233.8 & 201.2 & 259.0 & 7{,}892 & 2.10 \\
Grok 4.5 & 226.3 & 190.6 & 254.3 & 4{,}121 & 1.97 \\
Claude Sonnet 4.6 & 189.1 & 149.4 & 221.4 & 4{,}895 & 5.93 \\
Claude Opus 4.7 & 182.6 & 142.3 & 214.4 & 2{,}469 & 2.94 \\
GPT-5.2 & 162.6 & 123.0 & 194.0 & 2{,}141 & 1.78 \\
GPT-5.6 Terra & 150.2 & 113.7 & 184.2 & 1{,}689 & 0.75 \\
Claude Opus 4.8 & 134.1 & 97.6 & 166.8 & 2{,}538 & 3.16 \\
Claude Sonnet 4.5 & 101.5 & 59.7 & 133.8 & 3{,}389 & 3.67 \\
Claude Sonnet 5.5 & 101.2 & 59.3 & 139.7 & 2{,}006 & 0.74 \\
GPT-6 Luna & 92.2 & 50.9 & 123.7 & 1{,}880 & 0.04 \\
Grok 4.3 & 83.3 & 42.1 & 116.1 & 5{,}196 & 1.14 \\
GPT-5.6 Luna & 57.0 & 16.7 & 86.1 & 2{,}045 & 0.29 \\
GPT-5 mini & 54.3 & 9.7 & 95.3 & 5{,}019 & 0.83 \\
GPT-5.4 mini & 45.5 & -1.4 & 80.7 & 4{,}375 & 1.60 \\
Grok 4.20 Reasoning & 33.6 & -8.5 & 72.4 & 4{,}822 & 1.58 \\
Claude Haiku 5.5 & 0.7 & -52.8 & 46.7 & 2{,}523 & 0.06 \\
Claude Sonnet 5 & 0.0 & -42.7 & 37.6 & 2{,}247 & 1.01 \\
\bottomrule
\end{tabular}
\end{minipage}\hfill
\begin{minipage}[t]{0.486\textwidth}
\vspace{0pt}
\centering
\caption{Poker HUNL ($28$ models): Elo, 95\% CI, tokens, and cost.}
\label{tab:elo-poker-heads-up}
\fontsize{6.6pt}{9.0pt}\selectfont
\setlength{\tabcolsep}{2.1pt}
\begin{tabular}{@{}lrrrrr@{}}
\toprule
Model & Elo & $\text{CI}_{\text{lo}}$ & $\text{CI}_{\text{hi}}$ & Tok/T & \textcent/T \\
\midrule
GPT-6 Astra & 439.2 & 417.5 & 457.9 & 13{,}272 & 21.65 \\
GPT-5.6 Sol & 417.3 & 403.2 & 433.1 & 13{,}669 & 11.19 \\
GPT-5.5 & 412.4 & 395.9 & 430.8 & 15{,}630 & 14.98 \\
Claude Fable 5.1 & 379.1 & 367.1 & 391.9 & 18{,}742 & 26.20 \\
Gemini 3.7 Flash & 376.7 & 361.8 & 390.9 & 13{,}068 & 1.39 \\
Claude Opus 5 & 375.4 & 362.1 & 389.4 & 18{,}506 & 12.59 \\
GPT-5.6 Luna & 366.3 & 355.1 & 378.2 & 14{,}241 & 2.01 \\
GPT-5.6 Terra & 364.0 & 351.8 & 376.1 & 12{,}646 & 4.26 \\
GPT-5.2 & 363.1 & 351.3 & 375.2 & 13{,}918 & 3.80 \\
Gemini 3.6 Flash & 358.3 & 344.7 & 371.1 & 13{,}344 & 1.20 \\
Gemini 3.8 Flash & 353.2 & 342.0 & 366.2 & 15{,}468 & 2.31 \\
Claude Opus 4.6 & 351.7 & 340.0 & 364.3 & 14{,}751 & 10.67 \\
o3 & 333.1 & 320.2 & 344.9 & 14{,}372 & 3.49 \\
Grok 4.5 & 330.8 & 319.2 & 343.5 & 12{,}311 & 3.02 \\
GPT-5.4 mini & 327.9 & 314.8 & 340.4 & 17{,}426 & 3.05 \\
Grok 4.6 & 324.2 & 310.6 & 337.6 & 13{,}803 & 3.87 \\
Claude Sonnet 5 & 314.6 & 302.8 & 327.7 & 20{,}111 & 7.42 \\
Claude Sonnet 4.5 & 313.2 & 300.5 & 326.1 & 16{,}178 & 8.73 \\
GPT-5.4 & 312.2 & 300.0 & 325.6 & 15{,}906 & 7.78 \\
Claude Opus 4.8 & 310.9 & 298.8 & 322.5 & 18{,}408 & 12.12 \\
Claude Opus 4.7 & 299.0 & 283.3 & 313.8 & 17{,}281 & 9.81 \\
Gemini 3 Flash Preview & 297.0 & 281.9 & 312.1 & 14{,}051 & 1.27 \\
Claude Sonnet 4.6 & 289.7 & 276.0 & 304.2 & 15{,}687 & 8.12 \\
Grok 4.20 Reasoning & 264.1 & 246.4 & 279.8 & 12{,}904 & 1.59 \\
Grok 4.3 & 214.5 & 188.6 & 239.9 & 12{,}513 & 1.57 \\
GPT-5 mini & 209.2 & 183.5 & 233.1 & 18{,}070 & 1.49 \\
Gemini 3.1 Pro Preview & 187.0 & 157.4 & 213.2 & 12{,}508 & 4.43 \\
Gemini 3.5 Flash & 0.0 & -76.0 & 65.9 & 13{,}404 & 2.95 \\
\bottomrule
\end{tabular}
\end{minipage}
\end{table}
\FloatBarrier

\section{Cost-Performance Trade-Offs}
\label{sec:cost-performance-tradeoffs}

Table~\ref{tab:pareto-frontier} lists the ten-game Unified Leaderboard Elo ratings, 95\% bootstrap confidence intervals, active game coverage, game-normalized turn costs (\textcent; frozen $S_0$ normalizers, \Cref{sec:cost-normalization}), and Pareto-optimality status across all 39 active models corresponding to Figure~\ref{fig:pareto-frontier}.

\begin{table}[!htbp]
\centering
\caption{Game Arena: ten-game Unified Leaderboard Elo ratings (anchored to the frozen $S_0$ cohort mean), 95\% bootstrap confidence intervals, active game coverage, game-normalized cost per turn (\textcent, frozen $S_0$ normalizers), and Pareto optimality status across all 39 active models. Non-dominated Pareto frontier models are indicated in \textbf{bold}.}
\label{tab:pareto-frontier}
\fontsize{7.4pt}{8.4pt}\selectfont
\setlength{\tabcolsep}{4.5pt}
\begin{tabular}{lcrrrc}
\toprule
Model & Games & Unified Elo & 95\% CI & Norm.\ Cost (\textcent) & Pareto \\
\midrule
\textbf{Gemini 4 Carbon} & 9/10 & 424.1 & [416.2, 431.6] & 27.74 & \checkmark \\
\textbf{Gemini 4 Argon} & 9/10 & 361.1 & [354.3, 368.1] & 7.24 & \checkmark \\
Claude Fable 5.2 & 9/10 & 356.1 & [348.6, 363.5] & 13.17 & --- \\
GPT-6 Astra & 10/10 & 313.3 & [306.1, 320.8] & 14.21 & --- \\
GPT-5.5 & 10/10 & 283.6 & [276.5, 290.1] & 24.82 & --- \\
\textbf{GPT-6.1 Sol} & 9/10 & 278.4 & [271.6, 285.2] & 2.40 & \checkmark \\
Gemini 3.8 Flash & 10/10 & 266.1 & [259.3, 272.8] & 6.49 & --- \\
Claude Opus 5 & 9/10 & 248.2 & [242.2, 254.1] & 17.26 & --- \\
Claude Fable 5.1 & 9/10 & 231.7 & [225.0, 237.9] & 18.92 & --- \\
Gemini 3.1 Pro Preview & 10/10 & 227.5 & [221.5, 234.0] & 11.31 & --- \\
Claude Opus 5.5 & 9/10 & 226.3 & [218.6, 234.6] & 4.06 & --- \\
Gemini 3.7 Flash & 10/10 & 222.2 & [215.2, 229.4] & 2.50 & --- \\
GPT-5.6 Sol & 10/10 & 221.8 & [215.9, 228.4] & 8.25 & --- \\
\textbf{GPT-6.1 Terra} & 9/10 & 210.6 & [203.2, 217.5] & 1.04 & \checkmark \\
GPT-5.4 & 10/10 & 208.1 & [201.4, 213.8] & 15.18 & --- \\
GPT-6 Sol & 9/10 & 201.7 & [194.3, 208.1] & 2.70 & --- \\
Gemini 3.5 Flash & 10/10 & 201.4 & [192.8, 210.2] & 8.05 & --- \\
Gemini 3.6 Flash & 10/10 & 201.4 & [195.0, 206.5] & 3.34 & --- \\
Grok 4.7 & 9/10 & 195.3 & [189.0, 201.9] & 15.44 & --- \\
Grok 4.6 & 10/10 & 191.9 & [185.8, 198.5] & 9.06 & --- \\
Claude Sonnet 5.5 & 9/10 & 183.1 & [175.5, 190.1] & 1.79 & --- \\
Claude Opus 4.8 & 10/10 & 178.2 & [172.0, 184.8] & 17.65 & --- \\
Gemini 3 Flash Preview & 10/10 & 177.6 & [171.4, 183.9] & 5.00 & --- \\
GPT-5.2 & 10/10 & 176.8 & [169.9, 183.2] & 7.86 & --- \\
o3 & 10/10 & 164.5 & [157.9, 171.0] & 9.60 & --- \\
Grok 4.5 & 10/10 & 147.5 & [141.6, 152.8] & 6.40 & --- \\
Claude Opus 4.6 & 10/10 & 146.9 & [140.0, 153.5] & 19.71 & --- \\
Claude Sonnet 5 & 10/10 & 134.1 & [127.4, 140.5] & 8.83 & --- \\
Claude Opus 4.7 & 10/10 & 122.4 & [115.8, 128.7] & 19.08 & --- \\
\textbf{Claude Haiku 5.5} & 9/10 & 119.2 & [110.9, 127.1] & 0.23 & \checkmark \\
GPT-5.6 Terra & 10/10 & 114.3 & [107.8, 120.5] & 2.76 & --- \\
Grok 4.20 Reasoning & 10/10 & 106.4 & [100.3, 114.2] & 2.76 & --- \\
GPT-5.6 Luna & 10/10 & 104.1 & [96.4, 111.4] & 1.38 & --- \\
Claude Sonnet 4.6 & 10/10 & 100.3 & [93.3, 107.2] & 12.63 & --- \\
Claude Sonnet 4.5 & 10/10 & 94.9 & [87.6, 102.4] & 13.12 & --- \\
GPT-5 mini & 10/10 & 86.5 & [79.1, 94.4] & 1.93 & --- \\
GPT-5.4 mini & 10/10 & 81.4 & [74.2, 88.6] & 4.61 & --- \\
Grok 4.3 & 10/10 & 80.9 & [73.3, 88.2] & 2.14 & --- \\
\textbf{GPT-6 Luna} & 9/10 & 77.8 & [69.4, 86.0] & 0.13 & \checkmark \\
\bottomrule
\end{tabular}
\end{table}

\section{Game and Model Skill-Profile Details}
\label{app:skill-profile-radars}

\subsection{Skill-Profile Construction}
\label{app:skill-profile-method}

Let $\bar r_{g,d}$ be the mean expert rating of game~$g$ on construct
$d\in\{1,\ldots,6\}$. The construct-demand vector of game~$g$ is
\begin{equation}
    \mathbf{e}_g = \bigl(\bar{r}_{g,1},\; \bar{r}_{g,2},\; \dots,\; \bar{r}_{g,6}\bigr)
    \in \mathbb{R}^{6}.
    \label{eq:game-embedding}
\end{equation}

The per-game ratings $R_i^{(g)}$ that enter the profile are not the per-game
leaderboard ratings of \Cref{sec:per_game_bootstrap}: they come from a
separate, strongly regularized \bt{} fit. Let record~$m$ of the $N_g$ pairwise
records of game~$g$ (\Cref{sec:ulb-details}) be played between models $a_m$
and $b_m$, let $s_m\in\{0,0.5,1\}$ be the score of $a_m$, and let
$\pi_m=1/\bigl(1+10^{-(R_{a_m}^{(g)}-R_{b_m}^{(g)})/400}\bigr)$. The ratings are
obtained by running L-BFGS-B from a zero initialization on
\begin{equation}
    -\frac{1}{N_g}\sum_{m=1}^{N_g}\bigl[s_m\log\pi_m+(1-s_m)\log(1-\pi_m)\bigr]
    +\lambda\sum_{i\in\modelset_g}\bigl(R_i^{(g)}\bigr)^2,
    \qquad \lambda=10^{-5},
    \label{eq:profile-ratings}
\end{equation}
with default stopping tolerances (the optimizer stops within $0.001$ Elo of the
minimizer in nine games and after a single step in Bargaining, at $0.43$ of the
minimizer's scale), and are then centered over the models active in the game.
Because the loss is averaged over records while the penalty is not, the penalty
dominates: the ratings span only $0.6$--$9.9$ Elo per game, against
$317$--$1{,}648$ Elo for the leaderboard ratings, with which they have a
Spearman correlation of $0.93$--$0.98$ per game. For every model pair $(i,j)$,
the corresponding win probability is
\begin{equation}
    p_{ij}^{(g)}
    = \frac{1}{1+10^{-(R_i^{(g)}-R_j^{(g)})/400}}
    = \sigma\!\left((R_i^{(g)}-R_j^{(g)})\frac{\ln 10}{400}\right).
    \label{eq:pairwise-prob}
\end{equation}
Define the centered margin $m_{ij}^{(g)}:=p_{ij}^{(g)}-0.5$. With ratings this
small, $|m_{ij}^{(g)}|\le 0.015$ and the margin is linear in the rating
difference to within $0.03\%$,
$m_{ij}^{(g)}\approx(R_i^{(g)}-R_j^{(g)})\ln 10/1600$, so the profile is in
effect a demand-weighted average of each model's centered ratings rather than
of saturating win probabilities. The raw
skill-point vector for model~$i$ averages these margins over the games
$\games_i$ in which model~$i$ is active and, within each game, over the other
models in the set $\modelset_g$ of models active in game~$g$:
\begin{equation}
    \mathbf{s}_i
    = \frac{1}{|\games_i|}\sum_{g\in\games_i}\frac{1}{|\modelset_g|-1}
      \sum_{\substack{j\in\modelset_g\\j\ne i}}
      m_{ij}^{(g)}\,\mathbf{e}_g.
    \label{eq:skill-points}
\end{equation}
Finally, each coordinate is min--max rescaled across the evaluated models:
\begin{equation}
    \hat{s}_{i,d}
    = \frac{s_{i,d}-\min_{k\in\modelset}s_{k,d}}
           {\max_{k\in\modelset}s_{k,d}-\min_{k\in\modelset}s_{k,d}}.
    \label{eq:norm}
\end{equation}
This last step is a visualization convention: it forces every displayed
dimension to span $[0,1]$ within the evaluated model pool and does not produce
an absolute capability scale.

\begin{longtable}{p{0.24\textwidth} p{0.68\textwidth}}
\caption{Construct dimensions used for game selection and interpretation, with
endpoint anchors for the $0$--$5$ rating scale.}
\label{tab:constructs}\\
\toprule
\textbf{Dimension} & \textbf{Definition and scale anchors} \\
\midrule
\endfirsthead
\toprule
\textbf{Dimension} & \textbf{Definition and scale anchors} \\
\midrule
\endhead
\bottomrule
\endfoot
Search and planning &
Reasoning over future turns, possible replies, and long-term consequences
before committing to an action: look-ahead, sequencing, evaluating future
branches, and choosing actions based on downstream effects. Classic computer
chess and AlphaZero-style systems make this demand
explicit.\footnote{See \citet{shannon} and \citet{alphazero}.}
\anchors{a good move can be chosen without considering what happens next}
        {outcomes are decided by who searches deeper and plans further ahead} \\
\addlinespace
Tactical / constraint reasoning &
Local, rule-bound reasoning over the immediate position: legal and forced
moves, threats, traps, blocks, forks, valid clues, betting constraints, and
other combinatorial constraints whose consequences resolve within a turn or
two. Four in a row is a canonical example of a compact tactical game with solved
forced-line structure.\footnote{See \citet{allis}.}
\anchors{almost any action is legal and nothing resolves locally}
        {outcomes are decided by violating a constraint or missing an immediate
         consequence} \\
\addlinespace
Spatial / topological reasoning &
Understanding structured state geometry: board layout, adjacency,
connectivity, paths, territory, piece relations, and positional or spatial
pattern structure.
\anchors{the state has no meaningful geometry}
        {the core skill is reading board structure correctly} \\
\addlinespace
Uncertainty and belief reasoning &
Reasoning under hidden or incomplete information: inferring unseen states,
stochastic outcomes, expected value, and updating beliefs from
observations.\footnote{For the standard partially observable decision-process
framing, see \citet{pomdp}.}
\anchors{both players observe everything}
        {play is essentially an inference problem under hidden information} \\
\addlinespace
Social / mental-state reasoning &
Modeling other agents, opponents or partners, including their beliefs,
goals, likely actions, and likely interpretations; covers bluffing, deception,
cooperation, and adaptation. Poker and Codenames stress this dimension in
different ways.\footnote{For poker, see \citet{bowling} and \citet{libratus};
for Codenames, see \citet{codenames}.}
\anchors{the opponent is irrelevant beyond the position on the board}
        {success depends almost entirely on coordinating with or manipulating
         another player} \\
\addlinespace
Language and pragmatics &
Natural-language reasoning: semantic association and abstraction, clue
generation and interpretation, common ground, and speaker--listener
inference.\footnote{For the rational speech act framework, see \citet{rsa}.}
\anchors{play uses coordinates or notation only}
        {the game is played entirely in words} \\
\end{longtable}

\begin{table}[!htbp]
\centering
\small
\caption{Six-dimensional construct-demand profiles for the game suite. Scores are consensus ratings (0--5 scale) from expert survey.}
\label{tab:construct-demand}
\begin{tabular}{@{}lrrrrrr@{}}
\toprule
\textbf{Game} & \textbf{Planning} & \textbf{Tactical} & \textbf{Spatial} & \textbf{Belief} & \textbf{Social} & \textbf{Language} \\
\midrule
\textbf{Bargaining} & 2.14 & 2.00 & 0.29 & 4.14 & 4.14 & 1.14 \\
\textbf{Checkers} & 4.40 & 3.53 & 3.47 & 0.20 & 0.47 & 0.00 \\
\textbf{Chess} & 4.93 & 4.13 & 3.87 & 0.47 & 0.67 & 0.13 \\
\textbf{Coin Game} & 3.00 & 2.00 & 3.00 & 4.00 & 4.00 & 0.00 \\
\textbf{Dark Hex} & 4.25 & 3.25 & 3.50 & 3.25 & 1.75 & 0.00 \\
\textbf{Dots and Boxes} & 4.00 & 4.00 & 4.00 & 0.25 & 0.75 & 0.00 \\
\textbf{Four in a Row} & 4.50 & 3.79 & 3.36 & 0.21 & 0.36 & 0.00 \\
\textbf{Lines of Action} & 4.67 & 4.67 & 4.67 & 1.00 & 0.00 & 0.00 \\
\textbf{Poker (HUNL)} & 2.93 & 2.73 & 0.80 & 4.93 & 4.07 & 0.80 \\
\textbf{Word Association} & 1.22 & 1.56 & 0.44 & 3.11 & 3.89 & 4.56 \\
\bottomrule
\end{tabular}
\end{table}

\begin{table}[!htbp]
\centering
\small
\caption{Per-game expert rater coverage ($n_g$ of the $N=15$ AI research scientists, specializing in game research or game enthusiasts, who rated each game; the raters were recruited as volunteers from game-research and game-enthusiast groups, none is a paper author, and each skipped games they did not know), their self-reported familiarity ($0$--$5$ scale), and inter-rater reliability across canonical ordinal and squared-difference Krippendorff's $\alpha$ (overall ordinal $\alpha_{\mathrm{ord}} = 0.745$, $95\%$ rater-bootstrap CI $[0.725, 0.824]$; interval $\alpha = 0.755$; squared-difference $\alpha = 0.766$).}
\label{tab:appF-survey}
\resizebox{\textwidth}{!}{%
\begin{tabular}{@{}lrrrrr@{}}
\toprule
\textbf{Game} & \textbf{Active Raters ($n_g$)} & \textbf{Familiarity Mean} & \textbf{Familiarity SD} & \textbf{Canonical Ordinal $\alpha$} & \textbf{Squared-Diff.\ $\alpha$} \\
\midrule
Chess & $15$ & $4.33$ & $0.82$ & $0.81$ & $0.86$ \\
Checkers & $15$ & $4.07$ & $0.70$ & $0.81$ & $0.81$ \\
Four in a Row & $14$ & $3.92$ & $0.76$ & $0.82$ & $0.83$ \\
Dots and Boxes & $8$ & $3.62$ & $1.06$ & $0.77$ & $0.78$ \\
Lines of Action & $3$ & $2.67$ & $1.53$ & $0.82$ & $0.88$ \\
Poker (HUNL) & $15$ & $3.67$ & $1.11$ & $0.62$ & $0.61$ \\
Dark Hex & $4$ & $3.00$ & $1.41$ & $0.34$ & $0.43$ \\
Bargaining & $7$ & $3.14$ & $1.21$ & $0.70$ & $0.70$ \\
Word Association & $9$ & $3.67$ & $1.12$ & $0.67$ & $0.67$ \\
Coin Game & $3$ & $2.67$ & $1.53$ & $0.44$ & $0.55$ \\
\bottomrule
\end{tabular}}
\end{table}

\begin{figure}[p]
\centering
\includegraphics[width=\linewidth,height=0.84\textheight,keepaspectratio]{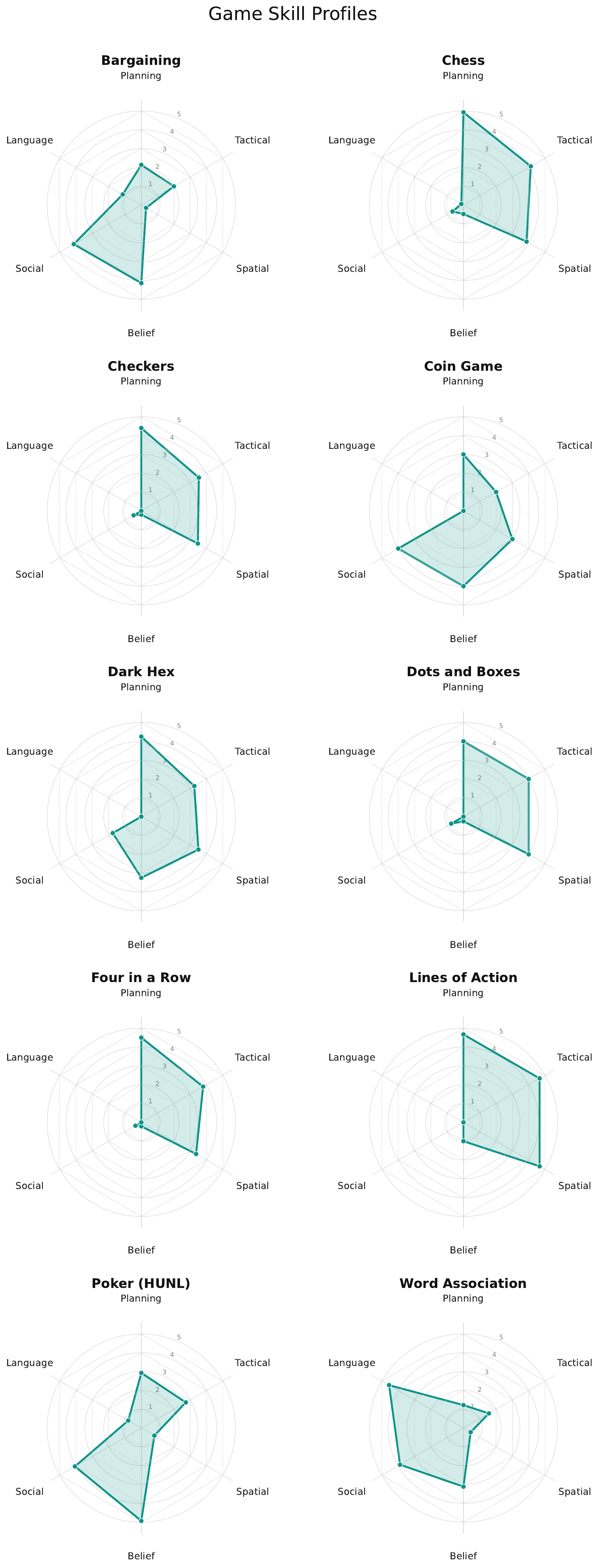}
\caption{Construct-demand profiles for the ten Game Arena benchmarks. The axes
correspond to the six dimensions defined in Section~\ref{sec:skills}; radial
values are the assigned game-level demand scores on the 0--5 scale.}
\label{fig:radar-games}
\end{figure}

\begin{figure}[p]
\centering
\includegraphics[width=\linewidth,height=0.86\textheight,keepaspectratio]{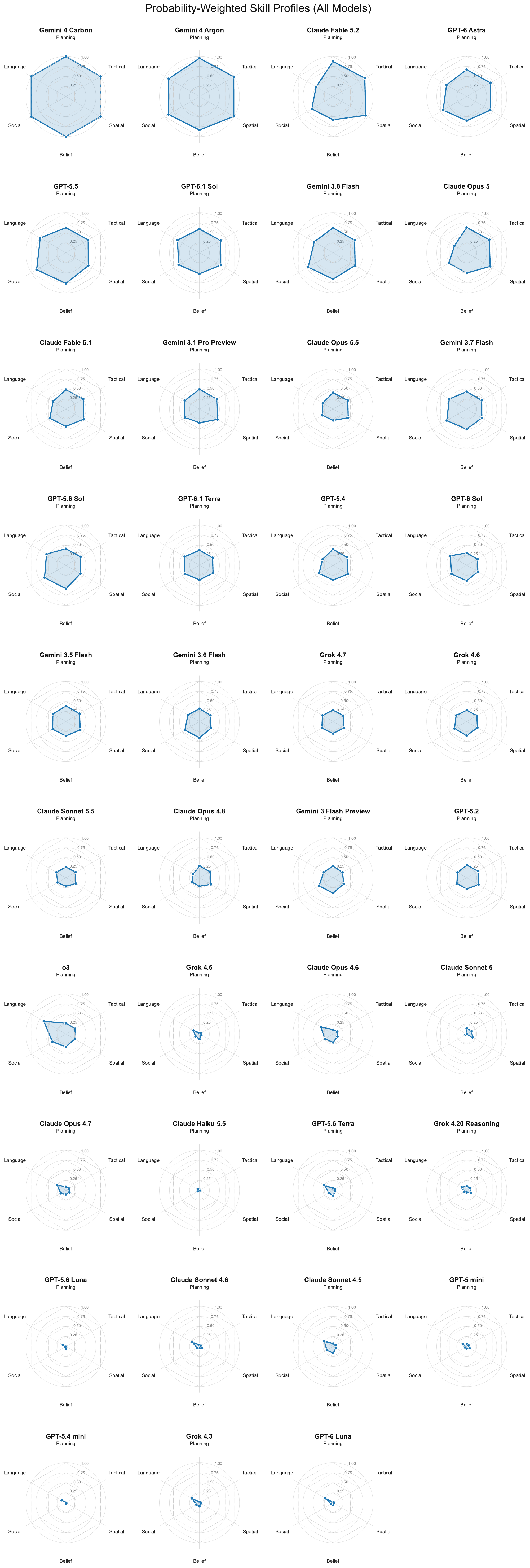}
\caption{Probability-weighted skill profiles for all 39 evaluated models. The axes correspond to the six construct-demand dimensions; values are normalized within each dimension according to Eq.~\ref{eq:norm}.}
\label{fig:radar-all-models}
\end{figure}

\section{Ground-Truth Tournament Simulation of the Adaptive Pairwise Scheduler}
\label{app:rq1-scheduler-validation}

Comparing the adaptive scheduler against a complete round-robin on real gameplay data requires observing the episodes that early stopping skips; due to API budget constraints (completing full round-robins across the ten-game suite would cost an additional \$2.369M; \Cref{tab:appG-dollars}), we use a complete real-game round-robin from an earlier Chess tournament (18 models, $6{,}120$ episodes; \Cref{app:chess-replay}), where replaying the scheduler saves $76.1\%$ of episodes while remaining highly correlated with the complete round-robin ranking ($\rho = 0.9965$, $\tau_K = 0.9780$). To verify across pool sizes $N \in \{10, 20, 30, 40, 50\}$ and diverse game regimes that the scheduler saves episodes while maintaining ranking accuracy, we run controlled ground-truth tournaments where underlying model strengths $E^* = (E_1^*, \dots, E_N^*)$ ($\theta_i^* = E_i^* \ln 10 / 400$) are known independently of the scheduler, supporting \Cref{sec:results-scheduler}.

\subsection{Generative Model and Scaling Across $N \in \{10, 20, 30, 40, 50\}$ Models}
\label{app:rq1-gt-sim}

For $N$ models with true centered Elo ratings $E^* \in \R^N$ ($\mathrm{SD} = 220$ Elo, per-pair cap $2T = 40$ episodes across $M = 200, 120, 80, 60, 50$ Monte Carlo trials for $N = 10, 20, 30, 40, 50$, respectively), each mirrored matchup on unordered pair $\{i, j\}$ simulates two games, $(i, j)$ with $i$ in Seat~0 and $(j, i)$ with $j$ in Seat~0, with optional Seat-0 advantage $\delta_{\mathrm{seat}}$ and Davidson draw parameter $\nu \ge 0$:
\[
  P(a \succ b \mid \Delta) = \frac{e^{\Delta/2}}{e^{\Delta/2} + e^{-\Delta/2} + \nu},
  \quad
  P(\text{draw} \mid \Delta) = \frac{\nu}{e^{\Delta/2} + e^{-\Delta/2} + \nu},
  \quad
  \Delta = \theta_a^* - \theta_b^* + \delta_{\mathrm{seat}}.
\]
Each trial compares two tournament runners: (1)~\textbf{Round-Robin} ($r = 1, \dots, T$ across all $\binom{N}{2}$ pairs, totaling $1{,}800 \to 49{,}000$ episodes at full cap, fitted via MM) and (2)~the \textbf{Adaptive Pairwise Scheduler} (Algorithm~\ref{alg:pairwise-scheduler}: online batch updates fitting penalized Newton--Raphson \bt{} with $\sigma_{\mathrm{prior}} = 1.0$, Hessian Wald confidence $q_{ij} = \Phi(|d_{ij}|/\sqrt{V_{ij}})$, and cubic target reduction $T_{ij}^{\mathrm{eff}} = \lfloor(1 - \mathrm{clip}(\frac{q_{ij}-0.95}{0.05}, 0, 1)^3)T\rfloor$, simulating games only while $c_{ij} < T_{ij}^{\mathrm{eff}}$).

\begin{figure}[!htbp]
\centering
\includegraphics[width=0.94\linewidth]{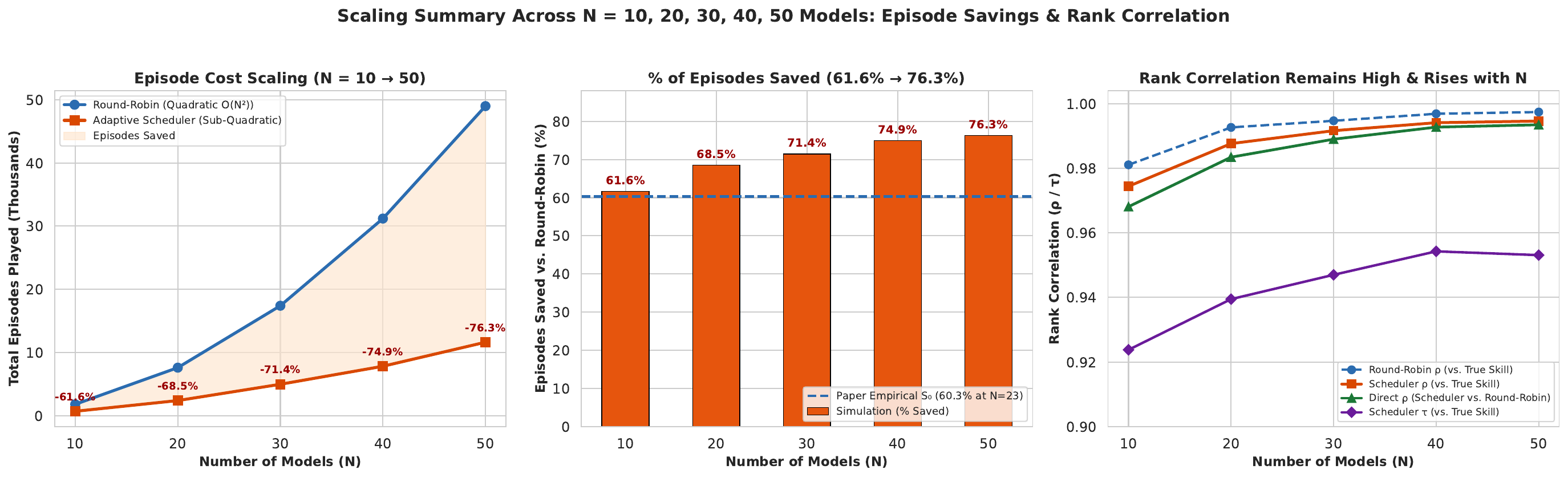}
\caption{Scaling laws of Adaptive Pairwise Scheduling across $N \in \{10, 20, 30, 40, 50\}$ models: sub-quadratic episode growth (\emph{left}), monotonic growth in episode savings from $61.6\%$ to $76.3\%$ (\emph{center}), and rising rank correlation against ground truth and Full Round-Robin (\emph{right}).}
\label{fig:scaling-laws-main}
\end{figure}

\begin{figure}[!htbp]
\centering
\includegraphics[width=0.94\linewidth]{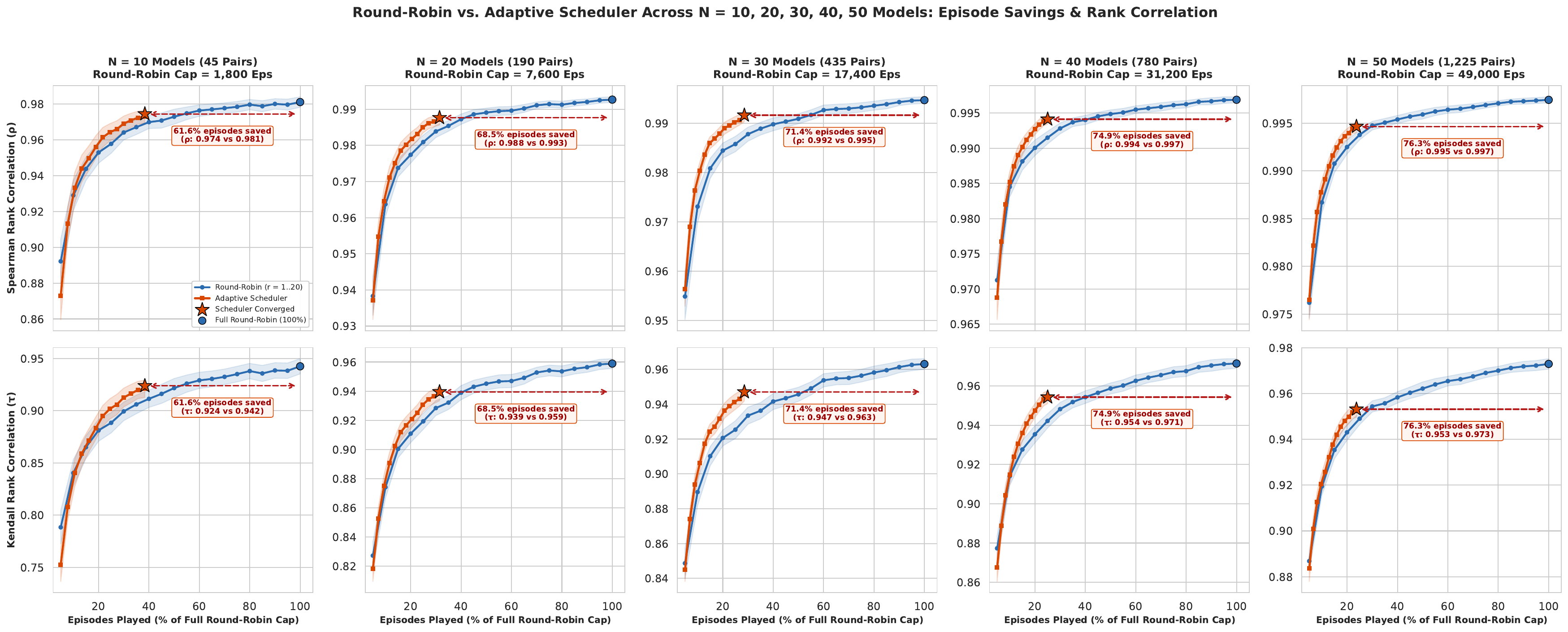}
\caption{Round-Robin ($r = 1,\dots,20$) vs.\ Adaptive Scheduler trajectories across $N \in \{10, 20, 30, 40, 50\}$ models: Spearman's $\rho$ (\emph{top}) and Kendall's $\tau_K$ (\emph{bottom}) as a function of episodes played (\% of Full Round-Robin cap $2T=40$). Beyond the initial warm-up rounds ($\ge 15\%$ of cap) through convergence, the Adaptive Scheduler sits above uniform Round-Robin at matched budgets across every pool size.}
\label{fig:appG-gt-frontier}
\end{figure}

\begin{figure}[!htbp]
\centering
\includegraphics[width=0.94\linewidth]{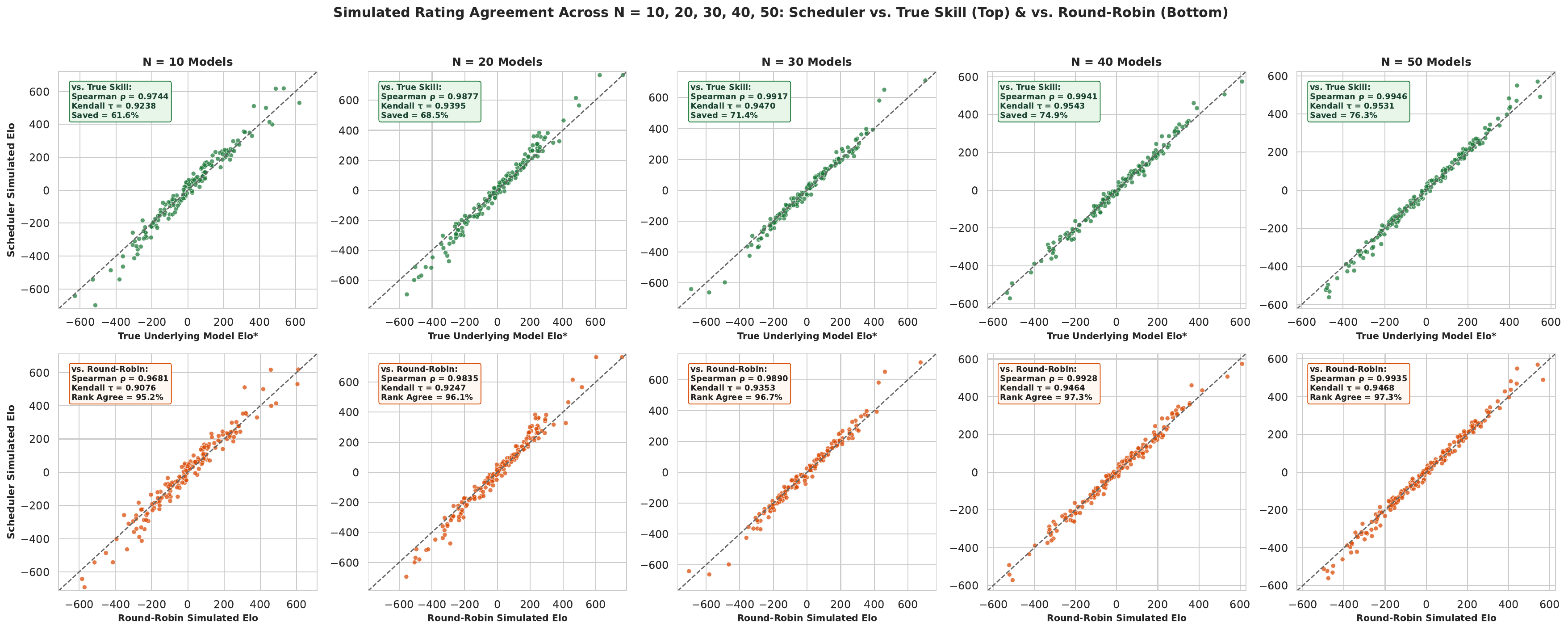}
\caption{Simulated Elo recovery across $N \in \{10, 20, 30, 40, 50\}$ models: Adaptive Scheduler vs.\ true underlying skill $E^*$ (\emph{top}) and directly vs.\ Full Round-Robin simulated Elo (\emph{bottom}, reaching $\rho = 0.9928$--$0.9935$ at $N=40$--$50$).}
\label{fig:appG-gt-scatter}
\end{figure}

\Cref{tab:gt-vs-true,tab:gt-direct-rr} and \Cref{fig:scaling-laws-main,fig:appG-gt-frontier,fig:appG-gt-scatter} report accuracy against ground truth and direct agreement between the two simulated leaderboards:
\begin{itemize}[leftmargin=*,itemsep=2pt,topsep=2pt]
  \item \textbf{Monotonic Sub-Quadratic Episode Savings:} Episode savings grow monotonically with pool size from 61.6\% ($691$ vs.\ $1{,}800$ episodes at $N=10$) to 68.5\% ($N=20$), 71.4\% ($N=30$), 74.9\% ($N=40$), and 76.3\% ($11{,}625$ vs.\ $49{,}000$ episodes at $N=50$), because shared comparisons across $N-2$ transitive opponents tighten $V_{ij} = C_{ii} + C_{jj} - 2C_{ij}$ faster in larger pools; the live releases saved $60.3\%$ at $N=23$ and $61.7\%$ at $N=39$ (\Cref{tab:scheduler-savings}).
  \item \textbf{Matched-Budget Superiority \& Full-Cap Rank Agreement:} Along the budget trajectories in \Cref{fig:appG-gt-frontier}, beyond the initial warm-up rounds ($\ge 15\%$ of cap) through convergence, the Adaptive Scheduler curve and converged operating point sit above uniform Round-Robin ($r = 1,\dots,20$) at matched episode budgets across every pool size $N=10\dots 50$ (e.g., at $N=30$ [$28.6\%$ cap], $\tau_K = 0.9470, \rho = 0.9917$ vs.\ equal-budget Round-Robin's $\tau_K = 0.9330, \rho = 0.9880$; at $N=50$ [$23.7\%$ cap], $\tau_K = 0.9531$ vs.\ $0.9441$). At convergence, Spearman rank correlation reaches $\rho = 0.9744 \to 0.9946$ ($96.19\%$--$97.71\%$ true pairwise ordering accuracy; $97.66\%$ at $N=50$) against $E^*$ (\Cref{tab:gt-vs-true}), and climbs from $\rho = 0.9681$ ($95.20\%$ pairwise rank agreement) at $N=10$ to $\rho = 0.9928$ ($97.26\%$) at $N=40$ and $\rho = 0.9935$ ($97.30\%$) at $N=50$ directly against Full Round-Robin (\Cref{tab:gt-direct-rr} and \Cref{fig:appG-gt-scatter}).
\end{itemize}

\begin{table}[!htbp]
\centering
\small
\caption{Ground-truth rank correlation and pairwise ordering accuracy ($\text{mean} \pm \text{SE}$) of Full Round-Robin ($2T=40$) vs.\ the Adaptive Pairwise Scheduler ($\tau_0=0.95$) across $N \in \{10, 20, 30, 40, 50\}$ models.}
\label{tab:gt-vs-true}
\resizebox{\textwidth}{!}{%
\begin{tabular}{@{}llrrrrrr@{}}
\toprule
\textbf{$N$} & \textbf{Tournament Policy} & \textbf{Mean Eps} & \textbf{\% Saved} & \textbf{Spearman $\rho$ (vs.\ True)} & \textbf{Kendall $\tau_K$ (vs.\ True)} & \textbf{Pearson $r$ (vs.\ True)} & \textbf{True Pairwise Acc.} \\
\midrule
$10$ & Round-Robin (Full $2T=40$) & 1{,}800 & $0.0\%$ & $0.9811 \pm 0.0015$ & $0.9425 \pm 0.0037$ & $0.9947$ & $97.07\%$ \\
$10$ & Adaptive Scheduler ($\tau_0=0.95$) & \textbf{691} & \textbf{61.6\%} & $\mathbf{0.9744 \pm 0.0018}$ & $\mathbf{0.9238 \pm 0.0042}$ & $0.9855$ & $96.19\%$ \\
\midrule
$20$ & Round-Robin (Full $2T=40$) & 7{,}600 & $0.0\%$ & $0.9927 \pm 0.0004$ & $0.9589 \pm 0.0017$ & $0.9974$ & $97.91\%$ \\
$20$ & Adaptive Scheduler ($\tau_0=0.95$) & \textbf{2{,}391} & \textbf{68.5\%} & $\mathbf{0.9877 \pm 0.0007}$ & $\mathbf{0.9395 \pm 0.0024}$ & $0.9904$ & $96.97\%$ \\
\midrule
$30$ & Round-Robin (Full $2T=40$) & 17{,}400 & $0.0\%$ & $0.9947 \pm 0.0003$ & $0.9631 \pm 0.0016$ & $0.9982$ & $98.13\%$ \\
$30$ & Adaptive Scheduler ($\tau_0=0.95$) & \textbf{4{,}969} & \textbf{71.4\%} & $\mathbf{0.9917 \pm 0.0005}$ & $\mathbf{0.9470 \pm 0.0019}$ & $0.9877$ & $97.35\%$ \\
\midrule
$40$ & Round-Robin (Full $2T=40$) & 31{,}200 & $0.0\%$ & $0.9969 \pm 0.0002$ & $0.9714 \pm 0.0013$ & $0.9987$ & $98.55\%$ \\
$40$ & Adaptive Scheduler ($\tau_0=0.95$) & \textbf{7{,}820} & \textbf{74.9\%} & $\mathbf{0.9941 \pm 0.0003}$ & $\mathbf{0.9543 \pm 0.0017}$ & $0.9916$ & $97.71\%$ \\
\midrule
$50$ & Round-Robin (Full $2T=40$) & 49{,}000 & $0.0\%$ & $0.9974 \pm 0.0002$ & $0.9729 \pm 0.0012$ & $0.9990$ & $98.63\%$ \\
$50$ & Adaptive Scheduler ($\tau_0=0.95$) & \textbf{11{,}625} & \textbf{76.3\%} & $\mathbf{0.9946 \pm 0.0003}$ & $\mathbf{0.9531 \pm 0.0014}$ & $0.9954$ & $97.66\%$ \\
\bottomrule
\end{tabular}}
\end{table}

\begin{table}[!htbp]
\centering
\small
\caption{Direct agreement ($\text{mean} \pm \text{SE}$) between the simulated Adaptive Scheduler leaderboard ($\hat{E}_{\mathrm{Scheduler}}$) and the simulated Full Round-Robin leaderboard ($\hat{E}_{\mathrm{Round\text{-}Robin}}$) across $N \in \{10, 20, 30, 40, 50\}$ models.}
\label{tab:gt-direct-rr}
\resizebox{\textwidth}{!}{%
\begin{tabular}{@{}lrrrrrrr@{}}
\toprule
\textbf{$N$ Models} & \textbf{RR Eps} & \textbf{Scheduler Eps} & \textbf{\% Saved} & \textbf{Direct Spearman $\rho$} & \textbf{Direct Kendall $\tau_K$} & \textbf{Direct Pearson $r$} & \textbf{Pairwise Rank Agreement} \\
\midrule
$N=10$ & 1{,}800 & \textbf{691} & \textbf{61.6\%} & $0.9681 \pm 0.0017$ & $0.9076 \pm 0.0041$ & $0.9801 \pm 0.0012$ & $95.20\%$ \\
$N=20$ & 7{,}600 & \textbf{2{,}391} & \textbf{68.5\%} & $0.9835 \pm 0.0008$ & $0.9247 \pm 0.0025$ & $0.9876 \pm 0.0015$ & $96.13\%$ \\
$N=30$ & 17{,}400 & \textbf{4{,}969} & \textbf{71.4\%} & $0.9890 \pm 0.0006$ & $0.9353 \pm 0.0020$ & $0.9861 \pm 0.0035$ & $96.70\%$ \\
$N=40$ & 31{,}200 & \textbf{7{,}820} & \textbf{74.9\%} & $0.9928 \pm 0.0004$ & $0.9464 \pm 0.0020$ & $0.9900 \pm 0.0044$ & $97.26\%$ \\
$N=50$ & 49{,}000 & \textbf{11{,}625} & \textbf{76.3\%} & $0.9935 \pm 0.0003$ & $0.9468 \pm 0.0016$ & $0.9942 \pm 0.0005$ & $97.30\%$ \\
\bottomrule
\end{tabular}}
\end{table}

\subsection{Robustness Across Skill Distributions, Game Regimes, and Wald Confidence Calibration}
\label{app:rq1-regimes-calib}

\begin{figure}[!htbp]
\centering
\includegraphics[width=0.94\linewidth]{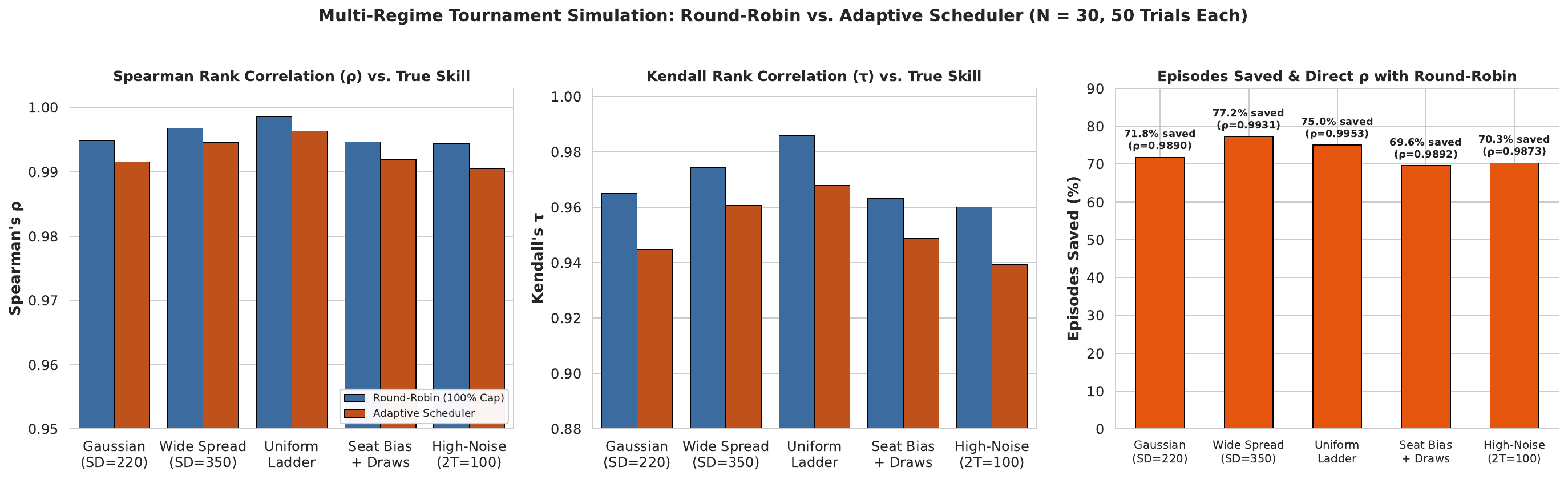}
\caption{Multi-regime ground-truth simulation ($N=30$, $50$ trials/regime): Spearman's $\rho$, Kendall's $\tau_K$, and percentage of episodes saved ($69.6\%$--$77.2\%$) alongside direct Round-Robin agreement.}
\label{fig:appG-gt-regimes}
\end{figure}

\begin{figure}[!htbp]
\centering
\includegraphics[width=0.94\linewidth]{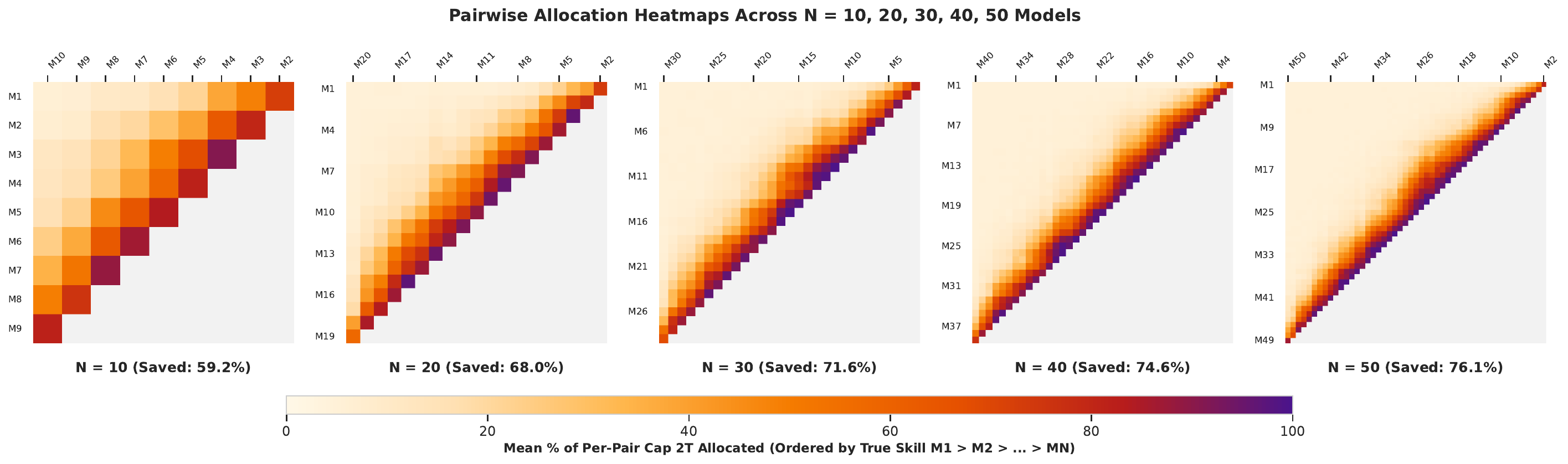}\\[3pt]
\includegraphics[width=0.88\linewidth]{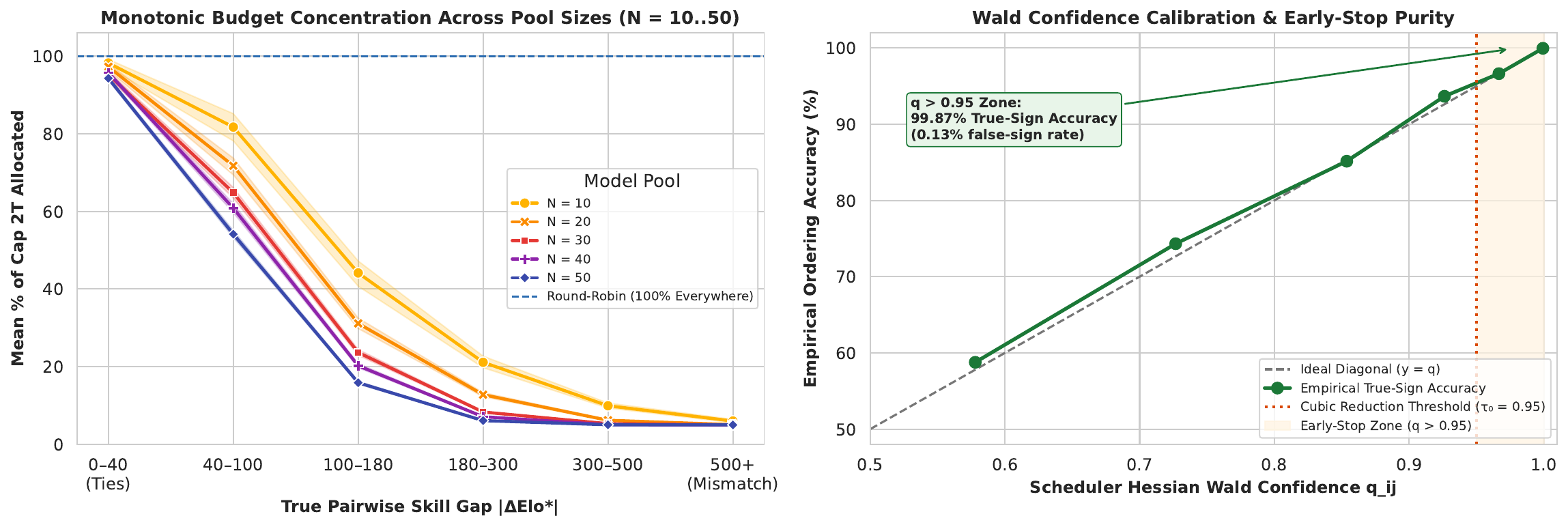}
\caption{\emph{Top:} Skill-ordered triangular allocation heatmaps across $N=10\dots 50$ showing sharpening diagonal concentration via transitive opponent information. \emph{Bottom:} Monotonic budget concentration across true $|\Delta\text{Elo}^*|$ and empirical $99.87\%$ true-sign accuracy in the $q_{ij} > 0.95$ early-stop zone. Both panels average 25 further simulated tournaments per pool size, drawn separately from the trials of \Cref{tab:gt-vs-true,tab:gt-direct-rr}, so the savings printed under the heatmaps ($59.2\%$--$76.1\%$) differ slightly from the tabulated $61.6\%$--$76.3\%$.}
\label{fig:appG-gt-calib}
\end{figure}

\Cref{tab:appG-gt-regimes} and \Cref{fig:appG-gt-regimes} repeat the tournament simulation across five skill and noise regimes at $N=30$ models ($50$ trials each): (1)~standard Gaussian ($\mathrm{SD}=220$, $2T=40$), (2)~wide spread matching Chess/Checkers ($\mathrm{SD}=350$, $2T=40$), (3)~uniform ladder ($[-400,+400]$, $2T=40$), (4)~strong first-player bias plus draws ($+60$ Elo Seat-0, $\nu=0.35$, $2T=40$), and (5)~compressed noisy skills matching Poker ($\mathrm{SD}=110$, $2T=100$), achieving 69.6\%--77.2\% episode savings and $\rho = 0.9873$--$0.9953$ direct agreement with Full Round-Robin across all five regimes. \Cref{fig:appG-gt-calib} shows the underlying allocation across $N=10\dots 50$: mismatched pairs stop after 1--2 mirrored rounds while closely matched pairs ($|\Delta\text{Elo}^*|\le 40$) retain near-full cap, and pairs entering the $q_{ij} > 0.95$ early-stop zone achieve 99.87\% empirical true-sign ordering accuracy ($0.13\%$ false-sign rate).

\begin{table}[!htbp]
\centering
\small
\caption{Multi-regime ground-truth tournament simulation ($N=30$ models, $50$ trials per regime) comparing Full Round-Robin ($100\%$ cap) and the Adaptive Pairwise Scheduler ($\tau_0=0.95$).}
\label{tab:appG-gt-regimes}
\resizebox{\textwidth}{!}{%
\begin{tabular}{@{}lrrrrrrr@{}}
\toprule
\textbf{Regime ($N=30$)} & \textbf{\% Saved} & \textbf{Sched $\rho$ (True)} & \textbf{RR $\rho$ (True)} & \textbf{Sched $\tau_K$ (True)} & \textbf{RR $\tau_K$ (True)} & \textbf{Direct $\rho$ (vs.\ RR)} & \textbf{Direct $\tau_K$ (vs.\ RR)} \\
\midrule
Gaussian Arena (SD=220, 2T=40) & \textbf{71.8\%} & \textbf{0.9916} & 0.9949 & \textbf{0.9446} & 0.9651 & \textbf{0.9890} & \textbf{0.9365} \\
Wide Skill Spread (SD=350, 2T=40) & \textbf{77.2\%} & \textbf{0.9945} & 0.9968 & \textbf{0.9607} & 0.9745 & \textbf{0.9931} & \textbf{0.9529} \\
Uniform Ladder ([-400,+400], 2T=40) & \textbf{75.0\%} & \textbf{0.9964} & 0.9986 & \textbf{0.9679} & 0.9859 & \textbf{0.9953} & \textbf{0.9614} \\
Seat Bias + Draws (+60 Elo, $\nu$=0.35) & \textbf{69.6\%} & \textbf{0.9919} & 0.9947 & \textbf{0.9487} & 0.9634 & \textbf{0.9892} & \textbf{0.9368} \\
High-Noise / Poker (SD=110, 2T=100) & \textbf{70.3\%} & \textbf{0.9905} & 0.9944 & \textbf{0.9394} & 0.9601 & \textbf{0.9873} & \textbf{0.9268} \\
\bottomrule
\end{tabular}}
\end{table}

\FloatBarrier

\section{Extended Continuous-Refresh Validation: Live 23-to-39 Model Refresh, Arrival Waves, and Hold-Out Replay}
\label{app:rq2-continuous-refresh}

This appendix documents the complete per-game accounting, rank stability, chronological wave history, leave-one-model-out insertion study, and full 39-model Unified Leaderboard supporting \Cref{sec:results-refresh}.

\paragraph{Retired models and the uniform baseline.}
Six of the 39 models (o3, GPT-5 mini, Grok 4.20 Reasoning, Claude Sonnet~4.5, Gemini 3.5 Flash, and Gemini 3.7 Flash) were retired from scheduling in early October 2026 as newer models arrived: none has an episode after October~9, whereas the other 33 models were still being scheduled in November. Their completed episodes remain in the $S_1$ set and in all ratings. Because of the retirement, 293 of the $6{,}972$ pair--game combinations among the models active in each game were never scheduled (the empty cells in \Cref{fig:scheduler-heatmaps}). Each pairs a retired model with a model that entered or resumed the game after the retired model's last episode there (in one Word Association pair the two overlap by ten minutes): Gemini 4 Carbon, Claude Fable~5.2, GPT-6.1 Terra, and Claude Haiku~5.5 in every game; Gemini 4 Argon in every game except four Bargaining pairs; Claude Opus~5.5 and Claude Sonnet~5.5 in Four in a Row and Word Association; and Claude Fable~5.1 in three Coin Game pairs. These combinations are unplayed for a reason unrelated to the scheduler, so they are excluded from every uniform baseline in this paper: the baselines in \Cref{tab:scheduler-savings,tab:appG-dollars,tab:appH-decomp} cover the $6{,}679$ combinations with at least one episode ($384{,}200$ episodes at the per-game caps), and the per-wave caps in \Cref{tab:appH-waves} are built from the same combinations. Counting the 293 combinations at their caps would add $14{,}440$ baseline episodes and raise the reported episode saving from $61.7\%$ to $63.1\%$ and the onboarding saving from $63.4\%$ to $65.6\%$.

\begin{figure}[!htbp]
\centering
\includegraphics[width=0.82\linewidth]{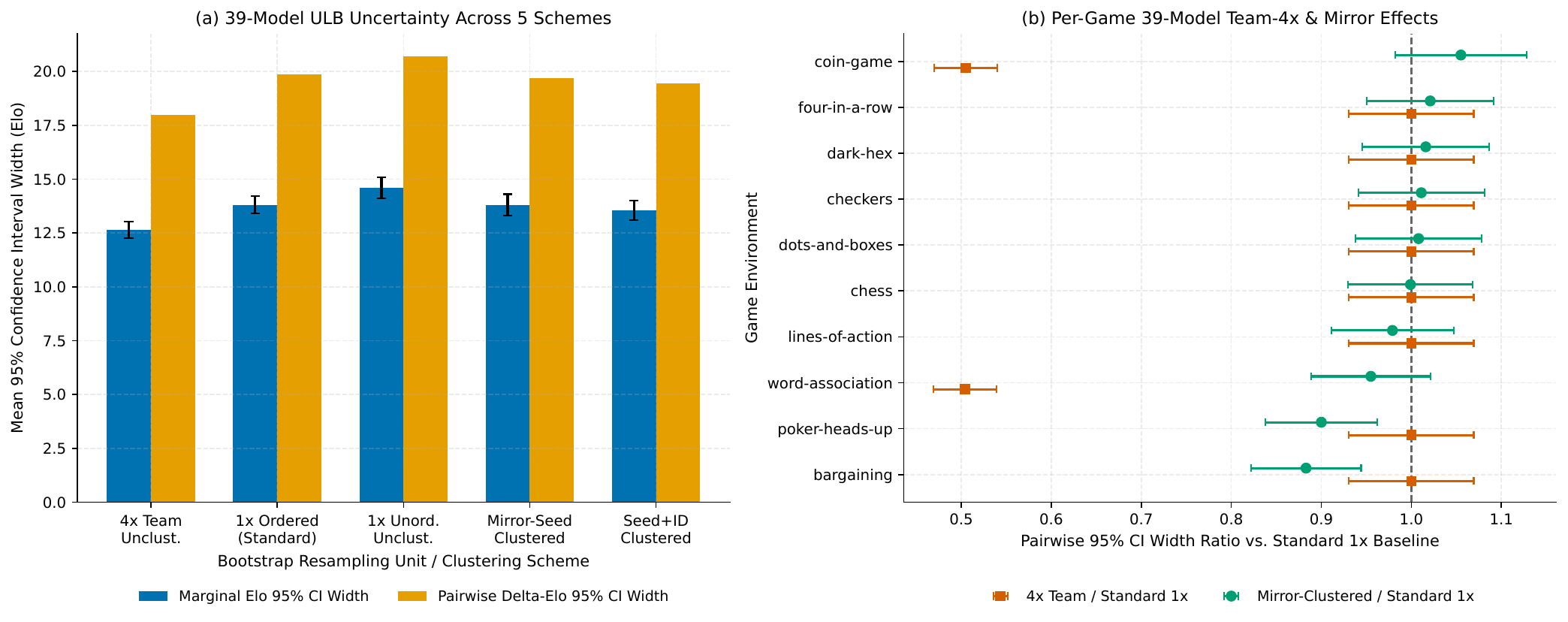}\\[3pt]
\includegraphics[width=0.82\linewidth]{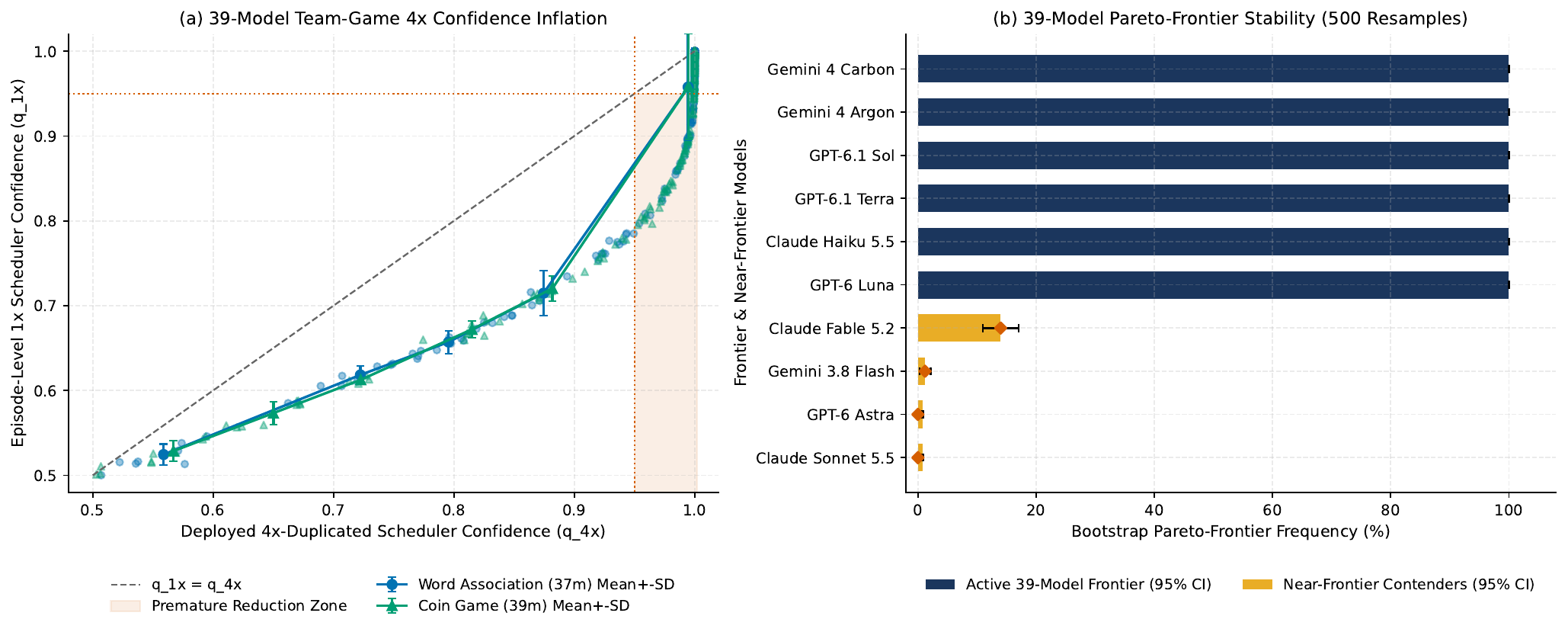}
\caption{\emph{Top:} 39-model ULB confidence interval widths across five bootstrap units and per-game mirror-seed variance ratios. \emph{Bottom:} $4\times \to 1\times$ team-episode scheduler confidence sensitivity in Word Association and Coin Game on the live logs ($38/624$ and $55/708$ scheduled pairs fall below $q_{ij} = 0.95$; in 25 simulated scheduler runs per game on the fitted ratings, savings fall from $79.6\%$ to $69.2\%$ and from $75.6\%$ to $62.8\%$) and bootstrap frequency on the 39-model Pareto frontier.}
\label{fig:appI-boot-team}
\end{figure}

\paragraph{Team-episode records and the live savings.}
Every live saving in this paper (\Cref{tab:scheduler-savings,tab:appG-dollars,tab:appH-decomp,tab:appH-waves}) counts episodes that the deployed scheduler played. In Word Association and Coin Game, that scheduler counted each $2\times 2$ team episode as four records (\Cref{app:scheduler-team-mirror}), which overstates $q_{ij}$ and stops pairs earlier than a one-record rule would (\Cref{fig:appI-boot-team}); the savings reported for these two games are therefore expected to exceed what a one-record scheduler would obtain. The eight two-player games use one record per episode and are unaffected. Over those eight games, the scheduler played $126{,}084$ episodes against $330{,}920$ at the caps ($61.9\%$ saved) for \$1.405M against \$3.726M ($62.3\%$, \$2.322M saved); onboarding the 16 new models saved $63.8\%$ of episodes ($68{,}556$ against $189{,}240$) and $64.8\%$ of API cost; and in the 23-model $S_0$ release the eight games saved $59.0\%$ of episodes ($58{,}082$ against $141{,}680$), below the ten-game $60.3\%$ because Coin Game saved $75.8\%$ there.

The two team games account for $13.9\%$ of the baseline episodes, $2.1\%$ of the baseline cost, and \$47.3k of the \$2.369M saved, which bounds their effect on the ten-game totals. If a one-record scheduler had played every scheduled team-game pair to its cap, the suite would still save $53.3\%$ of episodes and $61.0\%$ of API cost (\$2.322M), onboarding would save $54.1\%$ and $63.0\%$, and the $S_0$ release would save $51.6\%$ of episodes. The one-record savings lie between these floors and the deployed figures. As a point estimate, multiplying the live team-game episode counts by the increase in episodes played between the simulated four-record and one-record runs ($1.51\times$ and $1.52\times$; \Cref{fig:appI-boot-team}) gives $58.9\%$ of episodes and $61.8\%$ of API cost (\$2.352M). This is a projection rather than a measurement: under the deployed rule the simulated scheduler saves more than the live one ($79.6\%$ and $75.6\%$ against $61.5\%$ and $60.0\%$), and the two team games were not rerun with one record per episode.

\begin{figure}[!htbp]
\centering
\includegraphics[width=0.80\linewidth]{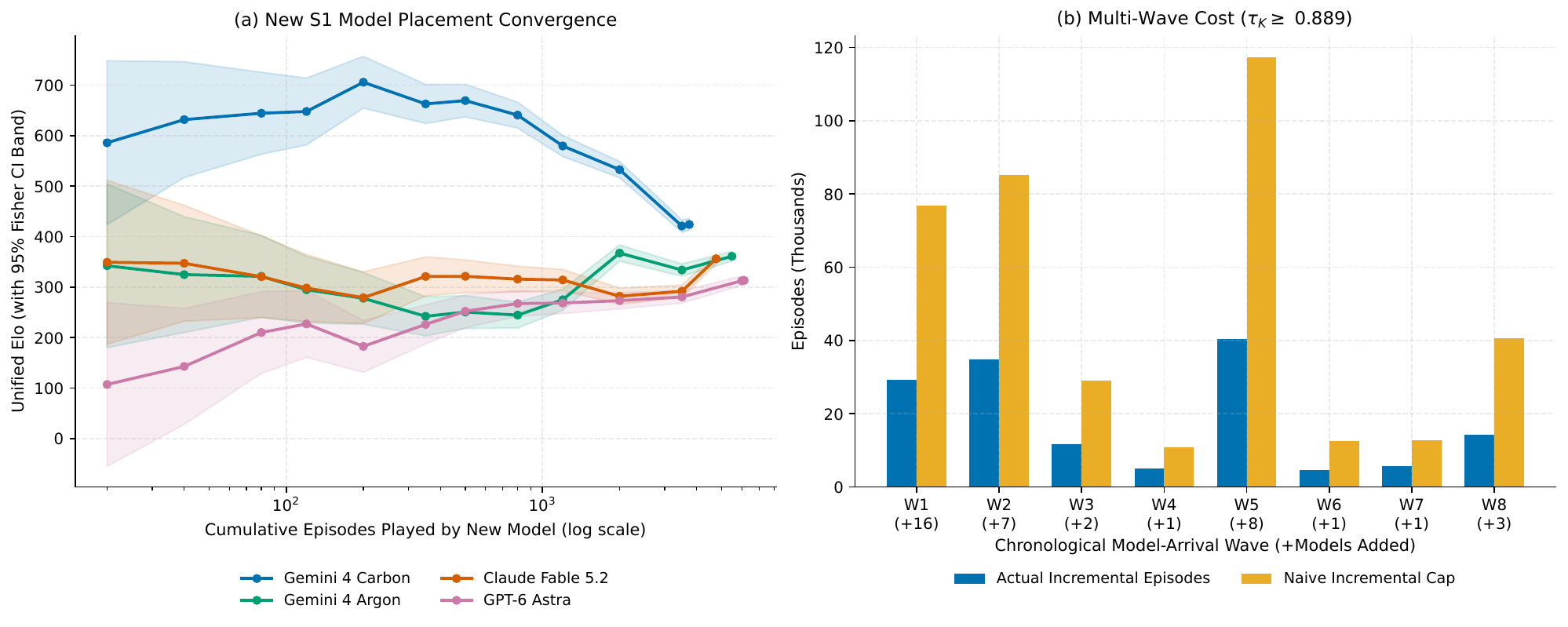}\\[3pt]
\includegraphics[width=0.80\linewidth]{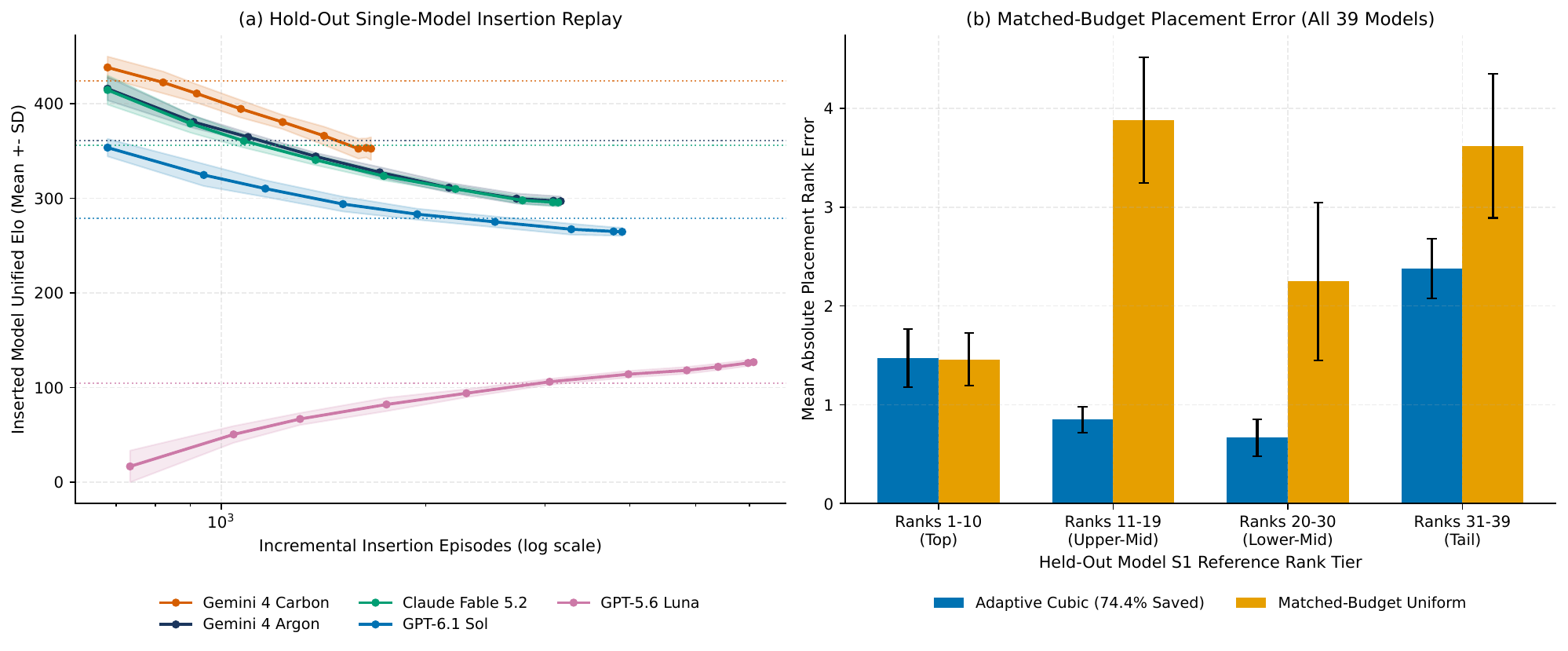}
\caption{\emph{Top:} New-model settling trajectories, and the episodes played on each wave's new pairs against their uniform cap across all 8 chronological waves (Waves~1--2 in $S_0$, Waves~3--8 post-$S_0$; \Cref{tab:appH-waves}). \emph{Bottom:} Simulated leave-one-model-out insertion across all 39 models: each model is removed and re-inserted against the remaining pool with outcomes drawn from the fitted per-game ratings, 12 runs per model ($74.4\%$ mean episode savings and a $2.09\times$ smaller mean distance from each model's full-data rank than equal-budget uniform allocation; GPT-5.6 Luna saves $72.7\%$ and ends $3.67$ vs.\ $6.00$ ranks from its full-data rank). The curves do not end on the dotted full-data ratings; the note at the start of this appendix explains why.}
\label{fig:appH-waves-holdout}
\end{figure}

\paragraph{Reading the simulated hold-out insertion.}
In \Cref{fig:appH-waves-holdout} (bottom left), the inserted model's unified rating does not converge to its full-data rating (dotted lines): the curves end 61--71 Elo below it for the three top-ranked models, 14 Elo below for GPT-6.1 Sol, and 22 Elo above for GPT-5.6 Luna. This is neither sampling noise nor an anchoring difference: given the live episodes, the simulation code returns the full-data rating exactly, and replacing the sampled outcomes with their expected scores while keeping each run's episode allocation moves the end point by at most 2.5 Elo for any of the 39 models. The offset comes from the allocation. One unified rating per model cannot match per-game strengths that differ across games (the per-game rating gap between a model and an opponent averages 235 Elo, against 92 Elo on the unified scale), so the unified rating depends on how a model's episodes are divided between lopsided and close matchups. The simulated scheduler is refit after every mirrored round and, for $98\%$ of the matchups with a per-game gap of at least 400 Elo, stops after 2 episodes in every run, whereas the live system played 6.9 episodes on such matchups on average; concentrating the inserted model's episodes on close opponents pulls its rating toward the middle of the pool (across the 39 models, the end point is approximately $0.76$ times the full-data rating plus 44 Elo). For the same reason, the rank distances reported for this simulation measure agreement with the live leaderboard under its own allocation rather than sampling error: with expected scores in place of sampled outcomes they are 1.29 ranks for the adaptive rule and 2.72 for matched-budget uniform allocation (1.31 and 2.74 with sampled outcomes). Across the 39 insertions, the scheduler requests $5{,}264$ episodes on average and saves $68.2\%$--$90.1\%$ of the inserted model's cap ($1{,}641$ of $16{,}640$ episodes for Gemini 4 Carbon, $3{,}136$ of $16{,}640$ for Claude Fable~5.2, and $6{,}006$ of $22{,}040$ for GPT-5.6 Luna). It ends closer to the full-data rank than matched-budget uniform allocation for 26 models and farther for 11, among them the top-ranked and the bottom-ranked model.

\begin{table}[!htbp]
\centering
\small
\caption{Full 39-model suite episode and token-level API dollar savings across all 10 games ($147{,}015$ clean episodes vs.\ a $384{,}200$-episode uniform cap over the $6{,}679$ pair--game combinations that were scheduled; cap dollars are summed over the same pairs). 100-hand HUNL Poker accounts for $72.5\%$ of total inference spend due to $257.6$ LLM calls/side and its $200$-episode pair cap.}
\label{tab:appG-dollars}
\resizebox{\textwidth}{!}{%
\begin{tabular}{@{}lrrrrrrrrr@{}}
\toprule
\textbf{Game} & \textbf{Active $M_g$} & \textbf{Cap $2T_g$} & \textbf{Active Cap Eps} & \textbf{Adaptive Eps} & \textbf{Ep.\ Saved} & \textbf{Turns/Side} & \textbf{Active Cap \$} & \textbf{Adaptive Actual \$} & \textbf{USD Saved vs.\ Cap} \\
\midrule
Chess & $39$ & $40$ & $28,440$ & $8,838$ & $68.92\%$ & $42.2$ & \$229,152 & \$72,977 & $68.15\%$ \\
Checkers & $39$ & $40$ & $28,440$ & $9,470$ & $66.70\%$ & $39.3$ & \$158,689 & \$57,587 & $63.71\%$ \\
Four in a Row & $39$ & $80$ & $55,920$ & $15,256$ & $72.72\%$ & $11.5$ & \$198,469 & \$55,204 & $72.19\%$ \\
Dots and Boxes & $39$ & $40$ & $28,440$ & $12,020$ & $57.74\%$ & $29.3$ & \$200,488 & \$85,327 & $57.44\%$ \\
Lines of Action & $39$ & $40$ & $28,440$ & $9,928$ & $65.09\%$ & $14.9$ & \$198,568 & \$65,785 & $66.87\%$ \\
Poker (HUNL) & $28$ & $200$ & $75,600$ & $29,932$ & $60.41\%$ & $257.6$ & \$2,684,516 & \$1,042,330 & $61.17\%$ \\
Dark Hex & $39$ & $40$ & $28,440$ & $12,314$ & $56.70\%$ & $8.5$ & \$23,353 & \$9,554 & $59.09\%$ \\
Bargaining & $39$ & $80$ & $57,200$ & $28,326$ & $50.48\%$ & $3.9$ & \$33,099 & \$15,892 & $51.99\%$ \\
Word Association & $37$ & $40$ & $24,960$ & $9,612$ & $61.49\%$ & $13.0$ & \$18,860 & \$7,049 & $62.62\%$ \\
Coin Game & $39$ & $40$ & $28,320$ & $11,319$ & $60.03\%$ & $20.0$ & \$61,372 & \$25,903 & $57.79\%$ \\
\midrule
\textbf{Total (10 Games)} & $39$ & $640$ & \textbf{384,200} & \textbf{147,015} & \textbf{61.73\%} & $66.0$ & \textbf{\$3,806,566} & \textbf{\$1,437,608} & \textbf{62.23\% (\$2.369M saved)} \\
\bottomrule
\end{tabular}}
\end{table}

\begin{table}[!htbp]
\centering
\small
\caption{Per-game live $S_0\ (23\text{m}) \to S_1\ (39\text{m}, \text{Nov 11, 2026})$ episode reuse and API dollar cost decomposition against exact Table~\ref{tab:scheduler-savings} caps ($28{,}573$ reused untouched $S_0$ episodes + $2{,}081$ existing-pair confidence re-checks + $34{,}725$ Sept-28 unified-harness upgrades + $81{,}636$ new-entrant onboarding episodes $= 147{,}015$ active $S_1$ episodes; of the $64{,}257$ archived $S_0$ export episodes [$64{,}251$ at the initial release cut], the remaining $64{,}257 - 28{,}573 = 35{,}684$ episodes are not in the active $S_1$ set: $35{,}522$ involved superseded pre-upgrade v1 submissions of the 8 harness-upgraded models, and $162$ were Word Association episodes among the 15 unchanged models, $148$ of them involving Claude Opus~5, which is not active in Word Association in $S_1$).}
\label{tab:appH-decomp}
\resizebox{\textwidth}{!}{%
\begin{tabular}{@{}lrrrrrrrrrrrrr@{}}
\toprule
\textbf{Game} & \textbf{Cap} & \textbf{$S_1$ Models} & \textbf{Reused $S_0$} & \textbf{Re-Check} & \textbf{Harness Upgr.} & \textbf{New Onboarding} & \textbf{Total $S_1$} & \textbf{Inc.\ Cap} & \textbf{Ep.\ Saved (New)} & \textbf{Full 39m Cap} & \textbf{New Actual \$} & \textbf{Inc.\ Cap \$} & \textbf{USD Saved (New)} \\
\midrule
Bargaining & $80$ & $39$ & $4,182$ & $152$ & $6,438$ & $17,554$ & $28,326$ & $36,960$ & $52.5\%$ & $57,200$ & \$10,101 & \$21,853 & $53.8\%$ \\
Checkers & $40$ & $39$ & $1,840$ & $34$ & $1,876$ & $5,720$ & $9,470$ & $18,320$ & $68.8\%$ & $28,440$ & \$31,957 & \$93,077 & $65.7\%$ \\
Chess & $40$ & $39$ & $1,888$ & $28$ & $1,302$ & $5,620$ & $8,838$ & $18,320$ & $69.3\%$ & $28,440$ & \$44,018 & \$136,258 & $67.7\%$ \\
Coin Game & $40$ & $39$ & $1,025$ & $585$ & $2,223$ & $7,486$ & $11,319$ & $18,200$ & $58.9\%$ & $28,320$ & \$16,255 & \$36,949 & $56.0\%$ \\
Dark Hex & $40$ & $39$ & $2,176$ & $86$ & $2,286$ & $7,766$ & $12,314$ & $18,320$ & $57.6\%$ & $28,440$ & \$5,991 & \$14,904 & $59.8\%$ \\
Dots \& Boxes & $40$ & $39$ & $2,802$ & $46$ & $2,200$ & $6,972$ & $12,020$ & $18,320$ & $61.9\%$ & $28,440$ & \$48,437 & \$123,945 & $60.9\%$ \\
Four in a Row & $80$ & $39$ & $2,698$ & $366$ & $3,144$ & $9,048$ & $15,256$ & $35,680$ & $74.6\%$ & $55,920$ & \$32,506 & \$123,258 & $73.6\%$ \\
Lines of Action & $40$ & $39$ & $1,828$ & $198$ & $2,000$ & $5,902$ & $9,928$ & $18,320$ & $67.8\%$ & $28,440$ & \$37,377 & \$118,863 & $68.6\%$ \\
Poker (HUNL) & $200$ & $28$ & $8,790$ & $290$ & $10,878$ & $9,974$ & $29,932$ & $25,000$ & $60.1\%$ & $75,600$ & \$408,465 & \$1,125,290 & $63.7\%$ \\
Word Assoc. & $40$ & $37$ & $1,344$ & $296$ & $2,378$ & $5,594$ & $9,612$ & $15,720$ & $64.4\%$ & $24,960$ & \$3,949 & \$11,495 & $65.6\%$ \\
\midrule
\textbf{Total} & $640$ & $39$ & \textbf{28,573} & \textbf{2,081} & \textbf{34,725} & \textbf{81,636} & \textbf{147,015} & \textbf{223,160} & \textbf{63.42\%} & \textbf{384,200} & \textbf{\$639,056} & \textbf{\$1,805,892} & \textbf{64.61\%} \\
\bottomrule
\end{tabular}}
\end{table}

\begin{table}[!htbp]
\centering
\small
\caption{Unified Leaderboard rank stability and ragged-coverage validation across $S_0 \to S_1$ (anchored to the frozen $S_0$ untouched cohort mean to eliminate weakest-model gauge shift).}
\label{tab:appH-stability}
\resizebox{\textwidth}{!}{%
\begin{tabular}{@{}lrrrrrrr@{}}
\toprule
\textbf{Cohort Comparison} & \textbf{Models ($n$)} & \textbf{Kendall's $\tau_K$} & \textbf{Spearman $\rho$} & \textbf{Pair Flips} & \textbf{Max Shift} & \textbf{Mean $|\Delta\text{Rank}|$} & \textbf{Top-5 Overlap} \\
\midrule
$S_0 \to S_1$ (Untouched $S_0$ Submissions) & $15$ & \textbf{0.9619} & \textbf{0.9929} & $2/105$ & $1$ & $0.27$ & $5/5$ \\
$S_0 \to S_1$ (All $23$ $S_0$ Names incl.\ $8$ Harness Upgrades) & $23$ & $0.7312$ & $0.8735$ & $34/253$ & $8$ & $2.70$ & $5/5$ \\
$S_1$ Full 39-Model Pool vs.\ $10/10$-Game Complete Cohort & $26$ & \textbf{0.9692} & \textbf{0.9959} & $5/325$ & $2$ & $0.31$ & $5/5$ \\
\bottomrule
\end{tabular}}
\end{table}

\begin{table}[!htbp]
\centering
\small
\caption{Reconstructed chronological model-arrival waves across May--November 2026. ``Window Eps'' and ``Window Cost'' count all episodes created between a wave's arrival and the next one; they include continued evaluation of earlier arrivals and, in Wave~5, the harness re-submissions. ``New-Pair Eps'' counts every episode played, in any window, on the pairs first formed by the wave (its new models against the existing pool and against each other), and ``Uniform Cap'' sums the per-game caps over the games in which those pairs were scheduled (Waves~1--2 on the $S_0$ archive, Waves~3--8 on $S_1$). Per-wave savings are $53.1\%$--$65.6\%$; Waves~1--2 add up to the archived $S_0$ total ($64{,}257$ of $161{,}920$; $64{,}251$ at the initial release cut) and Waves~3--8 to the onboarding total of \Cref{tab:appH-decomp} ($81{,}636$ of $223{,}160$). Existing-pool Kendall's $\tau_K = 0.9810$ in Wave~8, which added Gemini 4 Carbon, Claude Fable 5.2, and GPT-6.1 Terra.}
\label{tab:appH-waves}
\resizebox{\textwidth}{!}{%
\begin{tabular}{@{}lrrlrrrrrrr@{}}
\toprule
\textbf{Arrival Wave} & \textbf{Added} & \textbf{Pool} & \textbf{Models Onboarded} & \textbf{Window Eps} & \textbf{Window Cost (\$)} & \textbf{New-Pair Eps} & \textbf{Uniform Cap} & \textbf{Ep.\ Saved} & \textbf{Existing $\tau_K$} & \textbf{Max Shift} \\
\midrule
Wave 1 (Initial Launch) & $+16$ & $16$ & GPT-5 mini, Grok 4.20, GPT-5.2, o3 (+12) & $26{,}501$ & \$317,005 & $29{,}330$ & $76{,}800$ & $61.8\%$ & --- & --- \\
Wave 2 ($S_0$ Snapshot) & $+7$ & $23$ & Claude Sonnet 5, GPT-5.6 Sol/Terra/Luna (+3) & $37{,}756$ & \$489,003 & $34{,}927$ & $85{,}120$ & $59.0\%$ & $0.9333$ & $3$ \\
Wave 3 (Post-$S_0$ Refresh) & $+2$ & $25$ & Claude Fable 5.1, Grok 4.6 & $8{,}331$ & \$186,855 & $11{,}587$ & $29{,}000$ & $60.0\%$ & $1.0000$ & $0$ \\
Wave 4 (Post-$S_0$ Refresh) & $+1$ & $26$ & Grok 4.7 & $3{,}673$ & \$16,497 & $5{,}121$ & $10{,}920$ & $53.1\%$ & $0.9933$ & $1$ \\
Wave 5 (Gen-6 Batch + Harness) & $+8$ & $34$ & GPT-6 Astra/Sol, Opus 5.5, Gemini 3.7 (+4) & $78{,}094$ & \$809,072 & $40{,}349$ & $117{,}320$ & $65.6\%$ & $0.8892$ & $5$ \\
Wave 6 (Post-$S_0$ Refresh) & $+1$ & $35$ & Gemini 4 Argon & $1{,}284$ & \$773 & $4{,}686$ & $12{,}560$ & $62.7\%$ & $1.0000$ & $0$ \\
Wave 7 (Post-$S_0$ Refresh) & $+1$ & $36$ & Claude Haiku 5.5 & $9{,}481$ & \$29,196 & $5{,}606$ & $12{,}680$ & $55.8\%$ & $0.9765$ & $3$ \\
Wave 8 (Nov 11 Refresh) & $+3$ & $39$ & Gemini 4 Carbon, Claude Fable 5.2, GPT-6.1 Terra & $17{,}579$ & \$70,067 & $14{,}287$ & $40{,}680$ & \textbf{64.9\%} & \textbf{0.9810} & $2$ \\
\bottomrule
\end{tabular}}
\end{table}

\FloatBarrier
{\scriptsize\setlength{\tabcolsep}{4.0pt}
\begin{longtable}{@{}rllrrrrr@{}}
\caption{Longitudinal $S_0\ (23\text{m}) \to S_1\ (39\text{m}, \text{November 11, 2026})$ cohort comparison across all 39 active models (anchored to the frozen $S_0$ untouched-cohort mean; see \Cref{tab:ulb-results} for the main 39-model leaderboard and \Cref{tab:pareto-frontier} for Pareto cost coordinates).} \label{tab:appH-full39} \\
\toprule
\textbf{Rank} & \textbf{Model} & \textbf{Cohort} & \textbf{Games} & \textbf{Episodes} & \textbf{$S_0$ Elo [95\% CI]} & \textbf{$S_1$ Elo [95\% CI]} & \textbf{$\Delta$Elo} \\
\midrule
\endfirsthead
\toprule
\textbf{Rank} & \textbf{Model} & \textbf{Cohort} & \textbf{Games} & \textbf{Episodes} & \textbf{$S_0$ Elo [95\% CI]} & \textbf{$S_1$ Elo [95\% CI]} & \textbf{$\Delta$Elo} \\
\midrule
\endhead
\bottomrule
\endfoot
1 & Gemini 4 Carbon & Post-$S_0$ New & 9/10 & 3,739 & --- & 424.1 [416.2, 431.6] & --- \\
2 & Gemini 4 Argon & Post-$S_0$ New & 9/10 & 5,488 & --- & 361.1 [354.3, 368.1] & --- \\
3 & Claude Fable 5.2 & Post-$S_0$ New & 9/10 & 4,759 & --- & 356.1 [348.6, 363.5] & --- \\
4 & GPT-6 Astra & Post-$S_0$ New & 10/10 & 6,125 & --- & 313.3 [306.1, 320.8] & --- \\
5 & GPT-5.5 & $S_0$ Untouched & 10/10 & 7,410 & 310.6 [302.2, 319.3] & 283.6 [276.5, 290.1] & -27.0 \\
6 & GPT-6.1 Sol & Post-$S_0$ New & 9/10 & 5,823 & --- & 278.4 [271.6, 285.2] & --- \\
7 & Gemini 3.8 Flash & Post-$S_0$ New & 10/10 & 8,722 & --- & 266.1 [259.3, 272.8] & --- \\
8 & Claude Opus 5 & $S_0$ Untouched & 9/10 & 8,474 & 247.0 [238.2, 255.6] & 248.2 [242.2, 254.1] & +1.2 \\
9 & Claude Fable 5.1 & Post-$S_0$ New & 9/10 & 7,880 & --- & 231.7 [225.0, 237.9] & --- \\
10 & Gemini 3.1 Pro Preview & $S_0$ Untouched & 10/10 & 8,771 & 231.5 [223.0, 240.3] & 227.5 [221.5, 234.0] & -4.0 \\
11 & Claude Opus 5.5 & Post-$S_0$ New & 9/10 & 5,481 & --- & 226.3 [218.6, 234.6] & --- \\
12 & Gemini 3.7 Flash & Post-$S_0$ New & 10/10 & 7,678 & --- & 222.2 [215.2, 229.4] & --- \\
13 & GPT-5.6 Sol & $S_0$ Untouched & 10/10 & 9,028 & 243.3 [235.8, 251.3] & 221.8 [215.9, 228.4] & -21.5 \\
14 & GPT-6.1 Terra & Post-$S_0$ New & 9/10 & 6,195 & --- & 210.6 [203.2, 217.5] & --- \\
15 & GPT-5.4 & $S_0$ Untouched & 10/10 & 10,636 & 219.5 [212.6, 227.8] & 208.1 [201.4, 213.8] & -11.4 \\
16 & GPT-6 Sol & Post-$S_0$ New & 9/10 & 6,669 & --- & 201.7 [194.3, 208.1] & --- \\
17 & Gemini 3.5 Flash & $S_0$ Harness-v2 & 10/10 & 5,323 & 144.2 [136.0, 152.5] & 201.4 [192.8, 210.2] & +57.2 \\
18 & Gemini 3.6 Flash & $S_0$ Harness-v2 & 10/10 & 9,037 & 161.0 [152.3, 169.0] & 201.4 [195.0, 206.5] & +40.4 \\
19 & Grok 4.7 & Post-$S_0$ New & 9/10 & 7,501 & --- & 195.3 [189.0, 201.9] & --- \\
20 & Grok 4.6 & Post-$S_0$ New & 10/10 & 10,178 & --- & 191.9 [185.8, 198.5] & --- \\
21 & Claude Sonnet 5.5 & Post-$S_0$ New & 9/10 & 5,796 & --- & 183.1 [175.5, 190.1] & --- \\
22 & Claude Opus 4.8 & $S_0$ Harness-v2 & 10/10 & 9,695 & 96.9 [89.1, 105.8] & 178.2 [172.0, 184.8] & +81.3 \\
23 & Gemini 3 Flash Preview & $S_0$ Untouched & 10/10 & 9,134 & 173.2 [165.1, 180.6] & 177.6 [171.4, 183.9] & +4.4 \\
24 & GPT-5.2 & $S_0$ Untouched & 10/10 & 9,840 & 164.1 [155.9, 171.9] & 176.8 [169.9, 183.2] & +12.7 \\
25 & o3 & $S_0$ Untouched & 10/10 & 9,210 & 151.8 [143.7, 159.0] & 164.5 [157.9, 171.0] & +12.7 \\
26 & Grok 4.5 & $S_0$ Untouched & 10/10 & 10,774 & 154.7 [145.6, 162.5] & 147.5 [141.6, 152.8] & -7.2 \\
27 & Claude Opus 4.6 & $S_0$ Harness-v2 & 10/10 & 9,026 & 45.6 [36.2, 55.5] & 146.9 [140.0, 153.5] & +101.3 \\
28 & Claude Sonnet 5 & $S_0$ Untouched & 10/10 & 8,901 & 126.4 [117.2, 134.3] & 134.1 [127.4, 140.5] & +7.7 \\
29 & Claude Opus 4.7 & $S_0$ Harness-v2 & 10/10 & 8,041 & 68.8 [60.4, 77.8] & 122.4 [115.8, 128.7] & +53.6 \\
30 & Claude Haiku 5.5 & Post-$S_0$ New & 9/10 & 5,872 & --- & 119.2 [110.9, 127.1] & --- \\
31 & GPT-5.6 Terra & $S_0$ Untouched & 10/10 & 8,501 & 111.3 [103.0, 119.9] & 114.3 [107.8, 120.5] & +3.0 \\
32 & Grok 4.20 Reasoning & $S_0$ Untouched & 10/10 & 6,738 & 106.4 [97.8, 115.7] & 106.4 [100.3, 114.2] & 0.0 \\
33 & GPT-5.6 Luna & $S_0$ Untouched & 10/10 & 9,019 & 92.9 [84.6, 100.8] & 104.1 [96.4, 111.4] & +11.2 \\
34 & Claude Sonnet 4.6 & $S_0$ Harness-v2 & 10/10 & 7,070 & 23.9 [13.7, 33.4] & 100.3 [93.3, 107.2] & +76.4 \\
35 & Claude Sonnet 4.5 & $S_0$ Harness-v2 & 10/10 & 7,458 & 0.0 [-10.6, 10.6] & 94.9 [87.6, 102.4] & +94.9 \\
36 & GPT-5 mini & $S_0$ Untouched & 10/10 & 5,838 & 87.3 [78.1, 95.7] & 86.5 [79.1, 94.4] & -0.8 \\
37 & GPT-5.4 mini & $S_0$ Untouched & 10/10 & 6,999 & 62.2 [53.3, 70.9] & 81.4 [74.2, 88.6] & +19.2 \\
38 & Grok 4.3 & $S_0$ Harness-v2 & 10/10 & 5,883 & 60.3 [50.4, 69.8] & 80.9 [73.3, 88.2] & +20.6 \\
39 & GPT-6 Luna & Post-$S_0$ New & 9/10 & 5,318 & --- & 77.8 [69.4, 86.0] & --- \\
\end{longtable}
}

\begin{figure}[!htbp]
\centering
\includegraphics[width=0.74\linewidth]{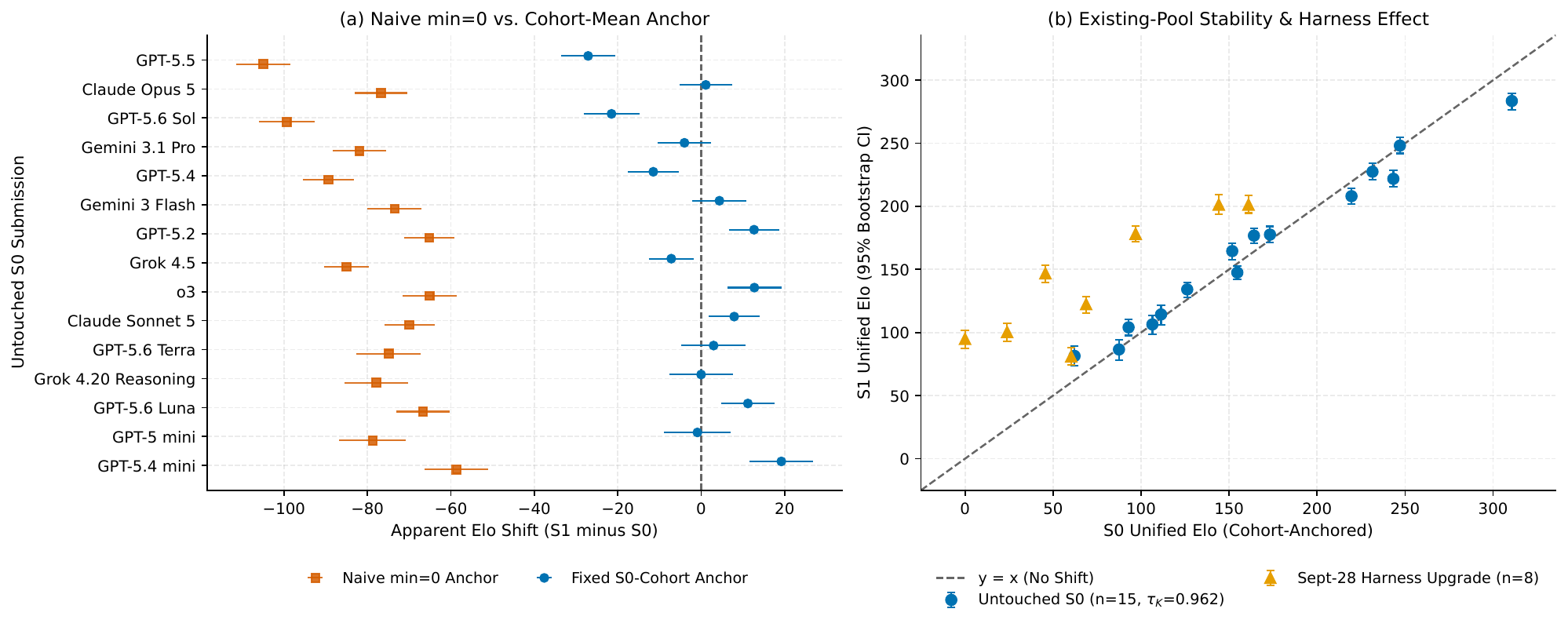}\\[3pt]
\includegraphics[width=0.37\linewidth]{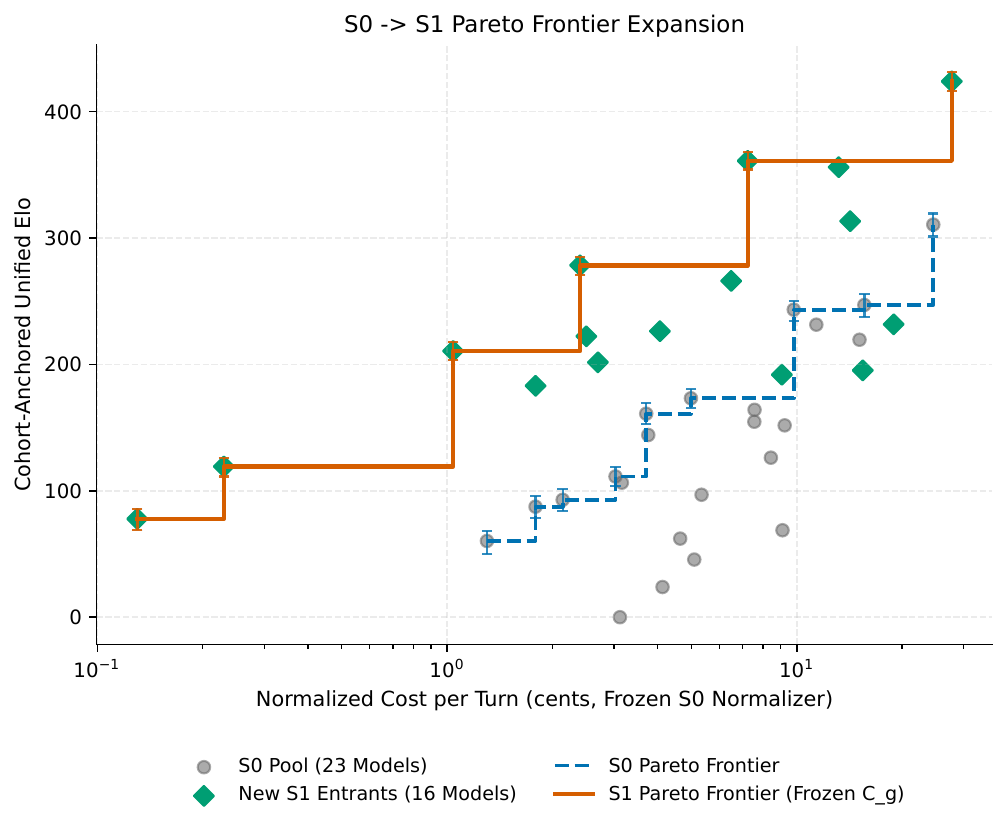}
\caption{\emph{Top:} Cohort-mean anchoring preserves $\tau_K = 0.9619$ ($\rho = 0.9929$, max shift $= 1$) across unchanged $S_0$ models during the $23 \to 39$ refresh. \emph{Bottom:} Complete turnover of the 39-model Pareto frontier to 6 post-$S_0$ entrants (each is on the frontier in $100\%$ of bootstrap resamples; \Cref{fig:appI-boot-team}).}
\label{fig:refresh-main}
\end{figure}

\begin{figure}[!htbp]
\centering
\includegraphics[width=0.73\linewidth]{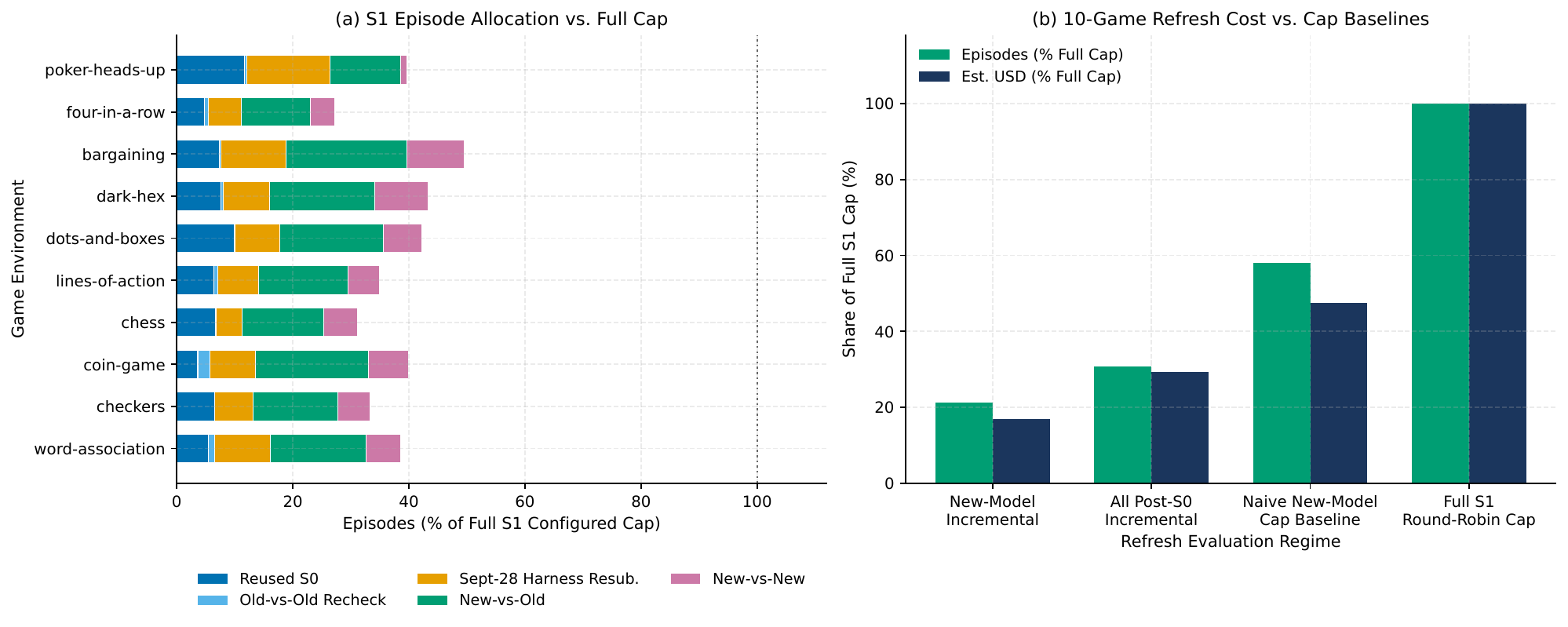}\\[3pt]
\includegraphics[width=0.73\linewidth]{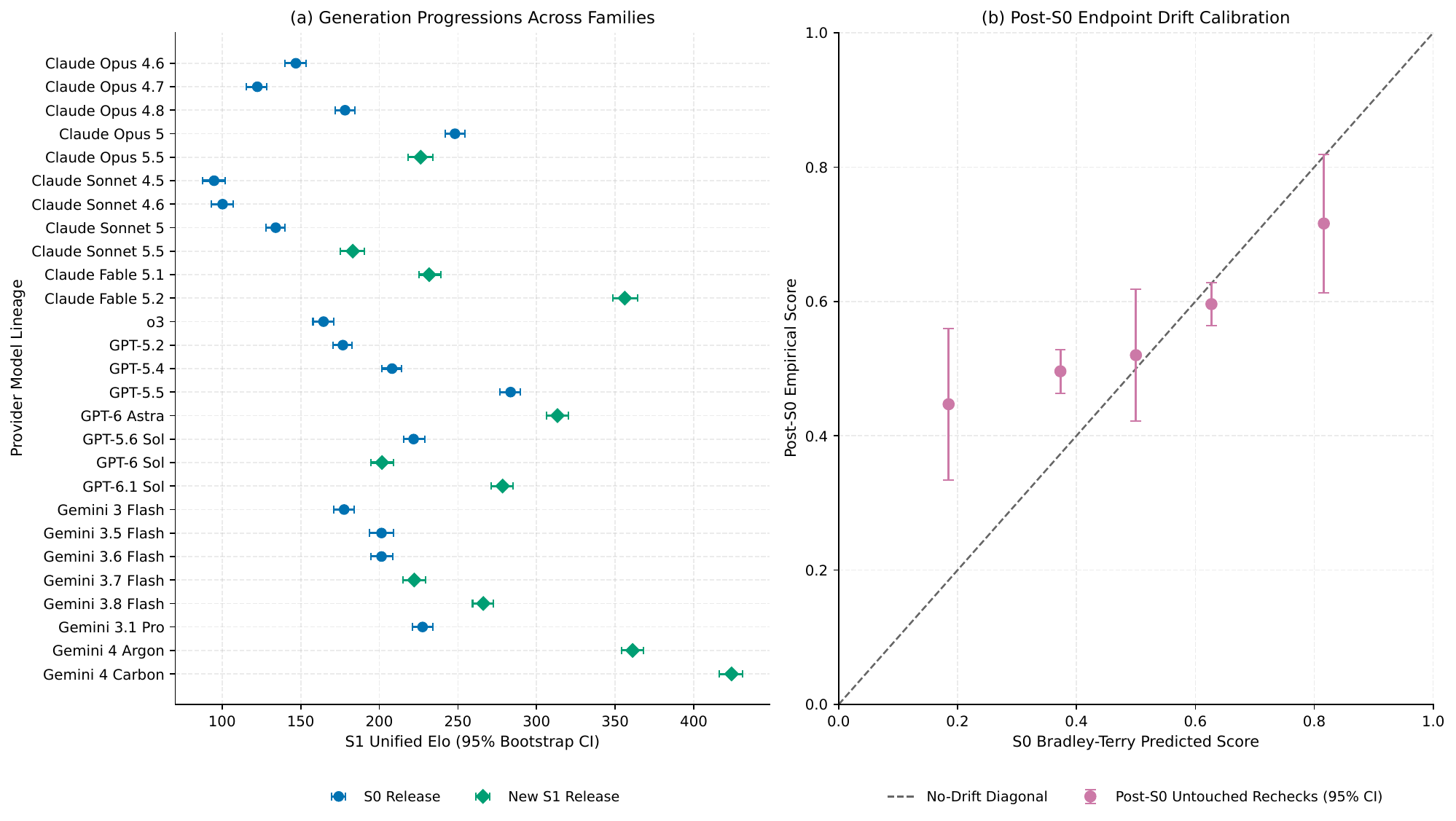}
\caption{\emph{Top:} Per-game episode and API dollar decomposition across $S_0 \to S_1$ ($63.4\%$ new-model onboarding savings; all caps are taken over scheduled pairs). \emph{Bottom:} Multi-generation family Elo progressions, and outcomes of the $2{,}081$ post-$S_0$ re-check episodes on existing pairs against their $S_0$-predicted scores (the scheduler requests further episodes for a pair only while its confidence is not high, and these outcomes lie closer to even than predicted).}
\label{fig:appH-cost-drift}
\end{figure}

\FloatBarrier
\section{Leaderboard Uncertainty Units and Cross-Game Aggregation Sensitivity}
\label{app:rq3-robustness}

This appendix reports the bootstrap resampling-unit comparison, $1\times$ vs.\ $4\times$ team-episode scheduler sensitivity, cross-game aggregation comparison, and leave-one-game-out stability supporting \Cref{sec:results-ulb,sec:ulb-details}.

\begin{table}[!htbp]
\centering
\small
\caption{39-model Unified Leaderboard uncertainty across five bootstrap resampling units ($B=500$ resamples, $147{,}015$ episodes, $741$ pairs). Grouping mirrored environment seeds yields tighter pairwise $\Delta\text{Elo}$ intervals ($693/741$ separated pairs) due to variance reduction across paired Bargaining ($0.883\times$ width) and Poker ($0.900\times$ width) seeds. Each row is a separate run of $500$ resamples. The intervals in \Cref{tab:ulb-results} come from the leaderboard's own run, which uses the unit of the second row; the two runs agree in mean width ($13.83$ and $13.82$ Elo), and individual widths, and with them the median, differ by Monte Carlo error.}
\label{tab:appI-boot}
\resizebox{\textwidth}{!}{%
\begin{tabular}{@{}lrrrrrrr@{}}
\toprule
\textbf{Bootstrap Resampling Unit (39 Models)} & \textbf{Samples} & \textbf{Mean CI Width} & \textbf{Median CI Width} & \textbf{Mean Pairwise $\Delta$ CI} & \textbf{Adjacent $\Delta$ CI} & \textbf{Separated Pairs} & \textbf{Separated Adj.} \\
\midrule
$4\times$ Team Cross-Product (Unclustered) & $500$ & $12.65$ & $12.64$ & $17.99$ & $17.76$ & $695/741$ & $11/38$ \\
$1\times$ Whole Episode (Standard Unit) & $500$ & $13.82$ & $14.16$ & $19.86$ & $19.87$ & $691/741$ & $11/38$ \\
$1\times$ Unordered Pair Cell (Unclustered) & $500$ & $14.60$ & $14.68$ & $20.72$ & $20.56$ & $691/741$ & $11/38$ \\
$1\times$ Mirror-Seed Clustered & $500$ & $13.82$ & $13.32$ & $19.69$ & $19.36$ & $693/741$ & $11/38$ \\
$1\times$ Mirror-Seed $+$ Match-ID Clustered & $500$ & $13.55$ & $13.54$ & $19.45$ & $19.13$ & $692/741$ & $11/38$ \\
\bottomrule
\end{tabular}}
\end{table}

\begin{table}[!htbp]
\centering
\small
\caption{Cross-game aggregation sensitivity across all 39 models ($741$ pairs, $147{,}015$ episodes) and the 23-model $S_0$ snapshot relative to Game-Balanced ULB. The top four models (Gemini 4 Carbon, Gemini 4 Argon, Claude Fable~5.2, GPT-6 Astra) remain the top four under six of the seven rules (equal-category weighting places GPT-5.5 fourth and GPT-6 Astra fifth), while fifth- and sixth-ranked GPT-5.5 ($283.6$) and GPT-6.1 Sol ($278.4$) sit within $5.2$ Elo.}
\label{tab:appI-agg}
\resizebox{\textwidth}{!}{%
\begin{tabular}{@{}lrrrrrrr@{}}
\toprule
\textbf{Aggregation Method} & \textbf{39m $\tau_K$} & \textbf{39m $\rho$} & \textbf{39m Flips} & \textbf{39m Max Shift} & \textbf{39m Top-5} & \textbf{$S_0$ (23m) $\tau_K$} & \textbf{$S_0$ Top-5} \\
\midrule
1. Game-Balanced BT (ULB Baseline) & 1.0000 & 1.0000 & 0/741 & 0 & 5/5 & 1.0000 & 5/5 \\
2. Equal-Pair Weighted BT & 0.9406 & 0.9909 & 22/741 & 4 & 4/5 & 0.9526 & 5/5 \\
3. Equal-Category Weighted BT & 0.9190 & 0.9883 & 30/741 & 4 & 5/5 & 0.9368 & 5/5 \\
4. Equal-Game BT Win-Prob Mean & 0.9406 & 0.9925 & 22/741 & 3 & 4/5 & 0.9684 & 5/5 \\
5. Equal-Game Empirical Win-Rate Mean & 0.9163 & 0.9862 & 31/741 & 4 & 5/5 & 0.9526 & 5/5 \\
6. Mean Per-Game Rank (Borda) & 0.9379 & 0.9917 & 23/741 & 4 & 5/5 & 0.8893 & 5/5 \\
7. Unweighted Pooled BT & 0.9460 & 0.9921 & 20/741 & 5 & 5/5 & 0.9209 & 5/5 \\
\bottomrule
\end{tabular}}
\end{table}

\begin{figure}[!htbp]
\centering
\includegraphics[width=0.84\linewidth]{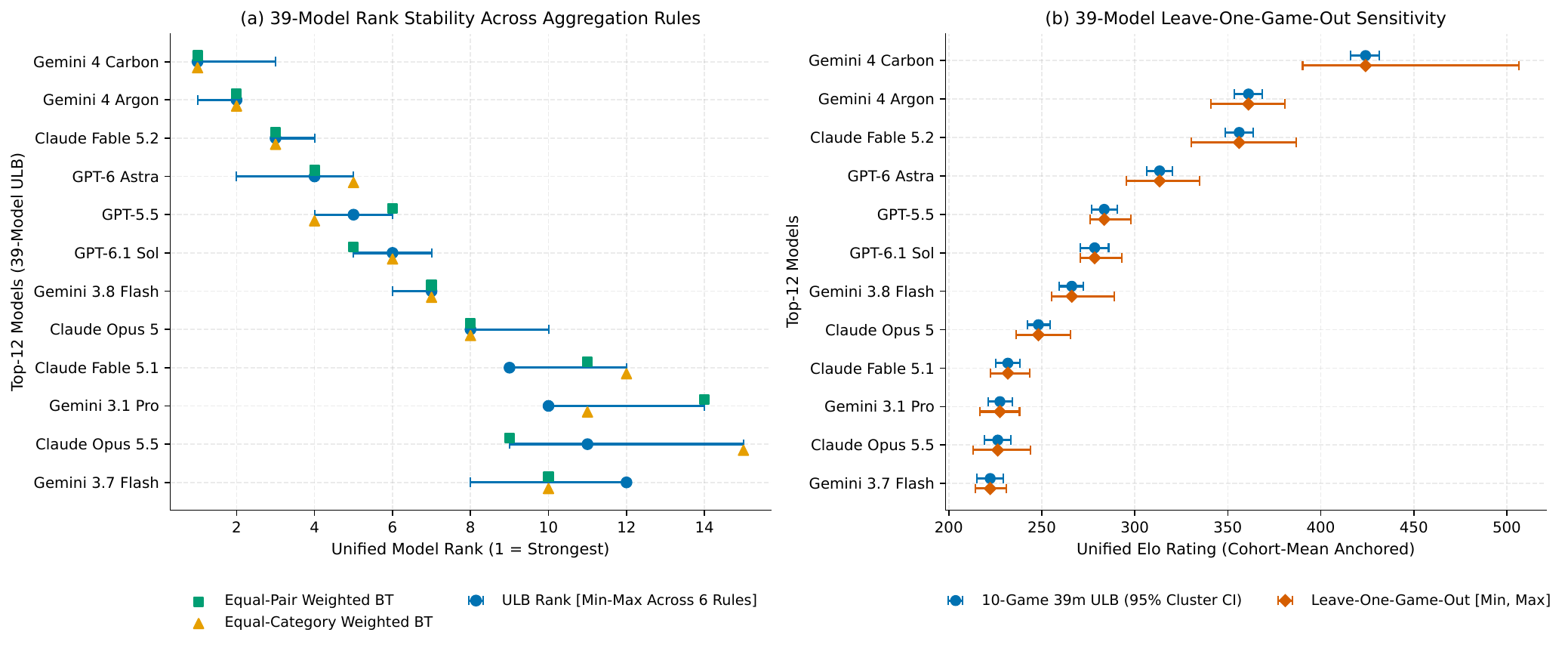}
\caption{39-model rank stability across cross-game aggregation rules (\emph{left}) and leave-one-game-out sensitivity (\emph{right}; Kendall's $\tau_K \in [0.9298, 0.9730]$ across all 10 single-game drops, under each of which the four top-ranked models keep the top four places and at least four of the top five are retained; each refit is shifted to the mean rating of the full 39-model fit).}
\label{fig:appI-agg-logo}
\end{figure}

\FloatBarrier
\section{Skill-Profile Specialization Case Study and Gameplay Logs}
\label{app:rq4-skill-specialization}

\begin{figure}[!htbp]
\centering
\includegraphics[width=0.95\linewidth]{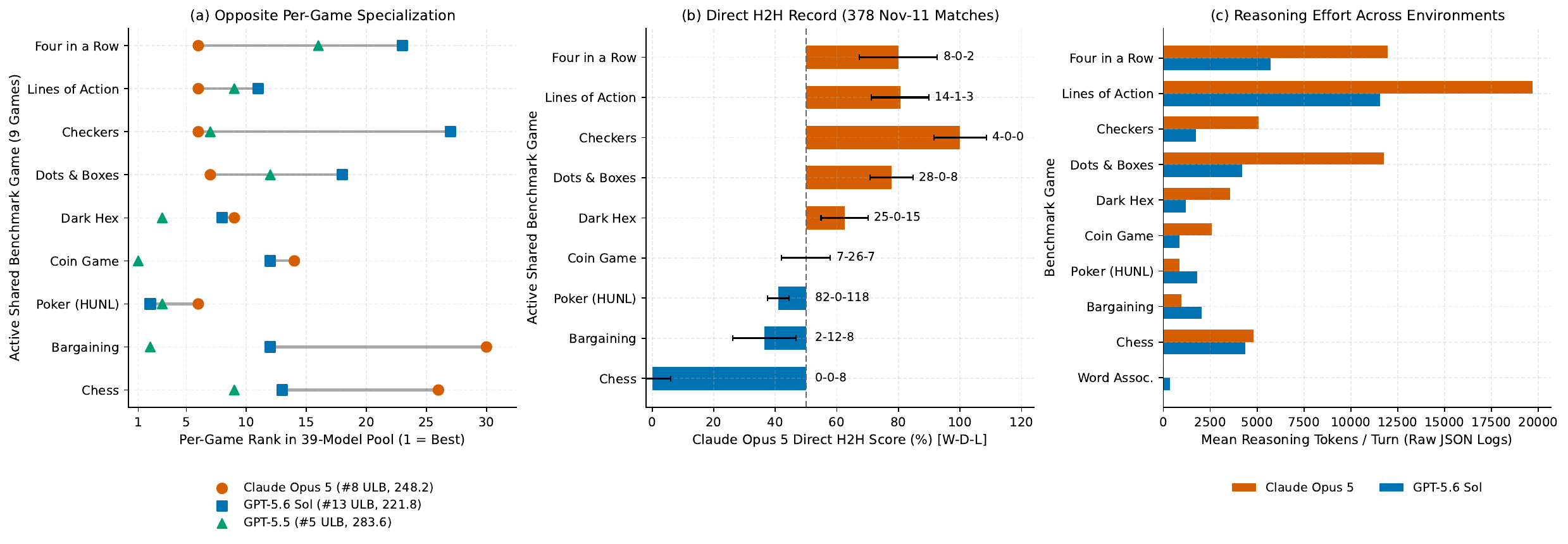}
\caption{Per-game specialization (\emph{left}; GPT-5.5, ranked fifth, is shown for reference), direct head-to-head outcomes across 378 matches (\emph{center}), and mean reasoning tokens per turn across 210 episode transcripts (\emph{right}) for Claude Opus~5 and GPT-5.6 Sol.}
\label{fig:appJ-gameplay}
\end{figure}

To illustrate how the construct-demand skill profiles distinguish similarly ranked models (\Cref{sec:results-skill-profiles}), \Cref{tab:appJ-opussol} and \Cref{fig:appJ-gameplay} compare Claude Opus~5 ($247.0$ Elo in $S_0$; $248.2$ in $S_1$) and GPT-5.6 Sol ($243.3$ Elo in $S_0$; $221.8$ in $S_1$) across all 378 direct head-to-head matches and 210 episode logs ($2{,}000$ Poker hands). Consistent with their construct-demand profiles, Claude Opus~5 outperforms GPT-5.6 Sol across combinatorial board games requiring multi-step look-ahead and spatial-tactical calculation ($54\text{--}1\text{--}13$ across Checkers, Four in a Row, Dots and Boxes, and Lines of Action, expending $1.7\times\text{--}2.9\times$ more reasoning tokens per turn; Chess is the exception, $0\text{--}0\text{--}8$), whereas GPT-5.6 Sol outperforms Claude Opus~5 in hidden-information strategic interaction and negotiation ($118\text{--}82$ in Poker and $8\text{--}12\text{--}2$ in Bargaining, expending $2.11\times\text{--}2.15\times$ more reasoning tokens per turn). In the 20 Poker transcripts ($2{,}000$ hands, $11{,}661$ actions), GPT-5.6 Sol checks or calls in $56.2\%$ of its actions and bets or raises in $29.4\%$, against $38.6\%$ and $49.6\%$ for Claude Opus~5. Claude Opus~5 is not active in Word Association in $S_1$, so that game is not part of this comparison.

\begin{table}[!htbp]
\centering
\small
\caption{Per-game rankings, head-to-head outcomes ($378$ matches), and mean reasoning tokens per turn ($210$ episode transcripts, including $2{,}000$ Poker hands; $198$ are head-to-head, and the $12$ Word Association transcripts are GPT-5.6 Sol's episodes against other opponents) for Claude Opus~5 and GPT-5.6 Sol.}
\label{tab:appJ-opussol}
\resizebox{\textwidth}{!}{%
\begin{tabular}{@{}lrrrrrrrrrr@{}}
\toprule
\textbf{Game} & \textbf{Active $M_g$} & \textbf{Opus 5 Rank} & \textbf{Opus 5 Elo} & \textbf{Sol Rank} & \textbf{Sol Elo} & \textbf{H2H (Opus--Draw--Sol)} & \textbf{Logs} & \textbf{Opus 5 Reas.\ Tok} & \textbf{Sol Reas.\ Tok} & \textbf{Token Ratio} \\
\midrule
Checkers & $39$ & $\#6$ & $1162.2$ & $\#27$ & $877.8$ & $4\text{--}0\text{--}0$ & $4$ & $5{,}089$ & $1{,}741$ & $2.92\times$ Opus \\
Four in a Row & $39$ & $\#6$ & $677.9$ & $\#23$ & $344.8$ & $8\text{--}0\text{--}2$ & $10$ & $11{,}968$ & $5{,}731$ & $2.09\times$ Opus \\
Dots and Boxes & $39$ & $\#7$ & $579.2$ & $\#18$ & $436.3$ & $28\text{--}0\text{--}8$ & $36$ & $11{,}752$ & $4{,}195$ & $2.80\times$ Opus \\
Lines of Action & $39$ & $\#6$ & $821.5$ & $\#11$ & $522.1$ & $14\text{--}1\text{--}3$ & $18$ & $19{,}707$ & $11{,}547$ & $1.71\times$ Opus \\
Dark Hex & $39$ & $\#9$ & $314.5$ & $\#8$ & $316.3$ & $25\text{--}0\text{--}15$ & $40$ & $3{,}553$ & $1{,}180$ & $3.01\times$ Opus \\
Coin Game & $39$ & $\#14$ & $405.7$ & $\#12$ & $416.1$ & $7\text{--}26\text{--}7$ & $40$ & $2{,}591$ & $855$ & $3.03\times$ Opus \\
Chess & $39$ & $\#26$ & $692.6$ & $\#13$ & $1134.2$ & $0\text{--}0\text{--}8$ & $8$ & $4{,}802$ & $4{,}361$ & $1.10\times$ Opus \\
\midrule
Poker (HUNL) & $28$ & $\#6$ & $375.4$ & $\#2$ & $417.3$ & $82\text{--}0\text{--}118$ & $20$ & $844$ & $1{,}819$ & $2.15\times$ Sol \\
Bargaining & $39$ & $\#30$ & $152.3$ & $\#12$ & $217.7$ & $2\text{--}12\text{--}8$ & $22$ & $968$ & $2{,}043$ & $2.11\times$ Sol \\
Word Association & $37$ & --- & --- & $\#8$ & $322.3$ & --- ($S_0$: $1\text{--}0\text{--}3$) & $12$ & --- & $358$ & --- \\
\bottomrule
\end{tabular}}
\end{table}

\FloatBarrier

\section{Scheduler Replay on a Recorded Chess Round-Robin Tournament}
\label{app:chess-replay}

The simulations in \Cref{app:rq1-scheduler-validation} draw outcomes from a \bt{} model, because the episodes that the scheduler skipped in the live suite were never played; playing them would cost a further \$2.369M at the per-game caps (\Cref{tab:appG-dollars}). An earlier Game Arena Chess tournament provides a check on recorded games. It was run as a complete round-robin among 18 models, 15 of which are among the 39 models of $S_1$: every pair played the 40-episode cap used for Chess in this paper, 20 episodes in each seat order, for $6{,}120$ episodes between December 2025 and May 2026. These episodes come from earlier submissions and share none with $S_0$ or $S_1$.

\paragraph{Replay.}
We run the scheduler code of the simulation ($\tau_0 = 0.95$, cubic target reduction, $\sigma_{\mathrm{prior}} = 1.0$, confidence refreshed after every 25 requested rounds) on these records: when it requests a mirrored round for a pair, it receives that pair's next recorded round (one episode in each seat order) instead of simulated games. Given the same rounds, the replay reproduces the allocation of the simulation code exactly. The order in which a pair's 20 recorded rounds are served is drawn at random, 200 times. For each ordering, the MM leaderboard fitted to the requested episodes is compared with the leaderboard of the complete tournament and with uniform allocation of exactly the same number of episodes, in which every pair receives the same number of recorded rounds (a round-robin with a smaller per-pair target).

\begin{figure}[!t]
\centering
\includegraphics[width=0.9\linewidth]{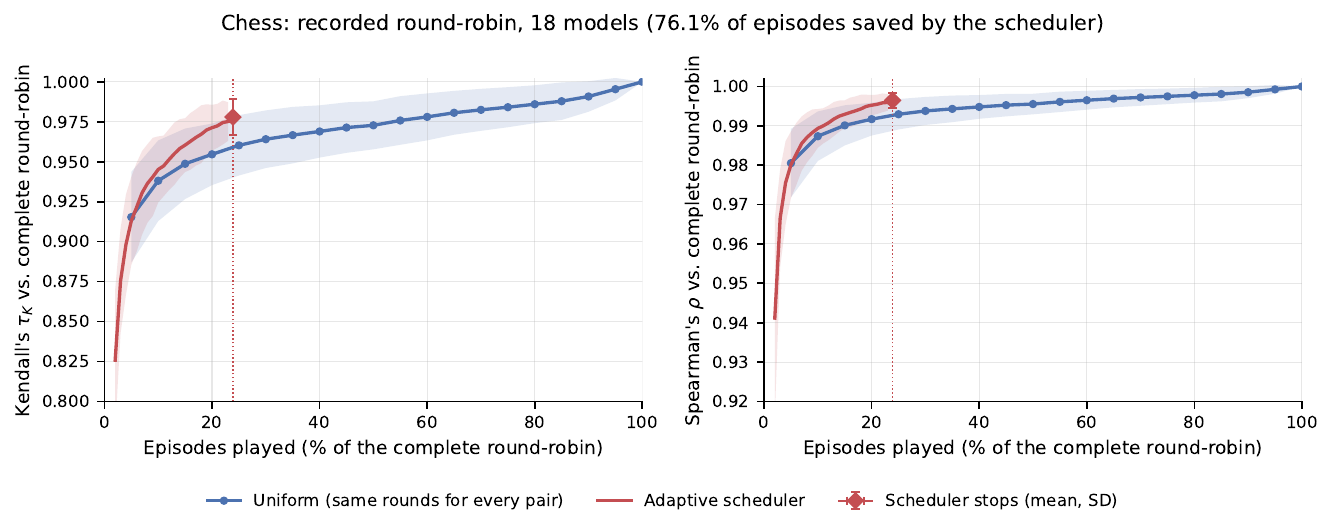}
\caption{Recorded Chess round-robin tournament: Kendall's $\tau_K$ (\emph{left}) and Spearman's $\rho$ (\emph{right}) against the leaderboard of the complete tournament, as a function of episodes played. Blue: uniform allocation after $r = 1, \dots, 20$ rounds per pair. Red: the adaptive scheduler while it runs and the point at which it stops (mean and standard deviation over 200 orderings; bands show one standard deviation).}
\label{fig:appL-chess-replay}
\end{figure}

\paragraph{Results.}
The scheduler requests $1{,}462$ of the $6{,}120$ episodes on average ($76.1\%$ saved; \Cref{tab:appL-chess-replay}). Its leaderboard agrees with that of the complete tournament at $\tau_K = 0.9780$ ($\rho = 0.9965$), against $0.9585$ ($\rho = 0.9926$) for uniform allocation of the same episodes; its $\tau_K$ is higher in $71.5\%$ of the orderings, equal in $18.5\%$, and lower in $10.0\%$. Matched-budget uniform allocation only knows to stop at $23.9\%$ of the cap in hindsight from the scheduler's own stopping point: a fixed smaller-$r$ round-robin offers no stopping criterion across games of differing noise or entrants of differing strength, whereas the adaptive rule halts automatically once pairwise orderings reach confidence ($q_{ij}>0.95$), nearly halves the remaining pairwise ordering error ($1.10\%$ vs.\ $2.08\%$), and requires $2.5\times$ fewer episodes than uniform allocation to reach the same $\tau_K$ ($23.9\%$ vs.\ $60\%$ of the complete tournament; \Cref{fig:appL-chess-replay}). The gain is on close pairs: the six pairs within 40 Elo are played to the cap and ordered as in the complete tournament in $75.2\%$ of cases, against $61.1\%$ under uniform allocation, whereas pairs more than 400 Elo apart stop after $2.4$ episodes on average, and no pair stopped before the cap has a fitted sign that differs from the complete tournament. The five highest-ranked models are the same five in every ordering ($54.5\%$ of the orderings under uniform allocation). With the confidence fit refreshed less often, as in the live system, where episodes are created before earlier ones complete, the scheduler saves less and stays ahead of uniform allocation (once per three passes over all pairs: $68.4\%$ saved, $\tau_K = 0.9777$ against $0.9637$).

Two quantities do not improve. First, the scheduler's ratings are not closer to those of the complete tournament in Elo terms (RMSE $36.2$ against $34.1$ Elo). Second, the scheduler does not recover the first place more often: the complete tournament separates its two highest-ranked models by only $6.7$ Elo ($95\%$ bootstrap interval $[-62, +67]$), and the scheduler places the second of them first in $77.5\%$ of the orderings, against $47.0\%$ under uniform allocation. The replay covers one game and 18 models, and its reference is itself a finite sample.

\begin{table}[!htbp]
\centering
\small
\caption{Replay of the adaptive scheduler on the recorded Chess round-robin tournament (18 models, 153 pairs, $6{,}120$ episodes): agreement with the leaderboard of the complete tournament for the scheduler and for uniform allocation of the same number of episodes. Values are means over 200 random orderings of each pair's recorded rounds; $\pm$ gives the standard deviation over orderings.}
\label{tab:appL-chess-replay}
\begin{tabular}{@{}lrr@{}}
\toprule
 & \textbf{Adaptive scheduler} & \textbf{Uniform, same episodes} \\
\midrule
Episodes played (of $6{,}120$) & $1{,}462$ ($23.9\%$) & $1{,}462$ ($23.9\%$) \\
Kendall's $\tau_K$ & $0.9780 \pm 0.0114$ & $0.9585 \pm 0.0188$ \\
Spearman's $\rho$ & $0.9965$ & $0.9926$ \\
Pairs ordered as in the complete tournament & $98.90\%$ & $97.92\%$ \\
\quad Pairs within 40 Elo (6 pairs) & $75.2\%$ & $61.1\%$ \\
\quad Adjacent-rank pairs (17 pairs) & $90.2\%$ & $83.6\%$ \\
Top five identical (share of orderings) & $100.0\%$ & $54.5\%$ \\
Largest rank change of a model & $0.950$ & $1.345$ \\
Rating RMSE (Elo) & $36.2$ & $34.1$ \\
\bottomrule
\end{tabular}
\end{table}

\clearpage
\bibliography{references}

\end{document}